\documentclass[lettersize,journal]{IEEEtran}
\usepackage{amsmath,amsfonts}
\usepackage{algorithmic}
\usepackage{array}
\usepackage{textcomp}
\usepackage{stfloats}
\usepackage{url}
\usepackage{verbatim}
\usepackage{graphicx}
\usepackage{float}  
\usepackage{subfigure}  
\usepackage{pifont}
\usepackage{utfsym}
\usepackage{booktabs}
\usepackage{multirow}
\usepackage{orcidlink} 
\hypersetup{hidelinks} 
\usepackage[normalem]{ulem}
\useunder{\uline}{\ul}{}

\def\BibTeX{{\rm B\kern-.05em{\sc i\kern-.025em b}\kern-.08em
    T\kern-.1667em\lower.7ex\hbox{E}\kern-.125emX}}
\usepackage{balance}

\begin{document}
\title{Passive Non-Line-of-Sight Imaging with Light Transport Modulation}
\author{
Jiarui Zhang$^{\orcidlink{0009-0007-5955-9537}}$,~\IEEEmembership{Graduate Student Member,~IEEE,}
Ruixu Geng$^{\orcidlink{0000-0002-5794-9802}}$,~\IEEEmembership{Graduate Student Member,~IEEE,}
\\
Xiaolong Du$^{\orcidlink{0009-0003-7152-2033}}$,~\IEEEmembership{Graduate Student Member,~IEEE,}
Yan Chen$^{\orcidlink{0000-0002-3227-4562}}$,~\IEEEmembership{Senior Member,~IEEE,}
\\
Houqiang Li$^{\orcidlink{0000-0003-2188-3028}}$,~\IEEEmembership{Fellow,~IEEE,}
and Yang Hu$^{\orcidlink{0000-0003-0379-1525}}$,~\IEEEmembership{Member,~IEEE}
\thanks{Manuscript received September 12, 2023; revised July 12, 2024 and
November 5, 2024; accepted December 7, 2024. This work was supported by the National Natural Science Foundation of China under Grant 62172381. The associate editor coordinating the review of this manuscript and approving it for publication was Dr. Robert Bregovic. (Corresponding authors: Yang Hu.)}
\thanks{Jiarui Zhang, Xiaolong Du, Houqiang Li and Yang Hu are with the School of Information Science and Technology, University of Science and Technology of China, Hefei 230026, China (e-mail: jrzhang@mail.ustc.edu.cn, duxiaolong@mail.ustc.edu.cn, lihq@ustc.edu.cn, eeyhu@ustc.edu.cn).}
\thanks{Ruixu Geng, Yan Chen are with the School of Cyber Science and Technology, University of Science and Technology of China, Hefei 230026, China (e-mail: gengruixu@mail.ustc.edu.cn, eecyan@ustc.edu.cn).}
\thanks{Digital Object Identifier 10.1109/TIP.2024.3518097}
}
\markboth{IEEE TRANSACTIONS ON IMAGE PROCESSING}%
{Passive Non-Line-of-Sight Imaging with Light Transport Modulation}

\maketitle

\begin{abstract}
Passive non-line-of-sight (NLOS) imaging has witnessed rapid development in recent years, due to its ability to image objects that are out of sight. The light transport condition plays an important role in this task since changing the conditions will lead to different imaging models. Existing learning-based NLOS methods usually train independent models for different light transport conditions, which is computationally inefficient and impairs the practicality of the models. In this work, we propose NLOS-LTM, a novel passive NLOS imaging method that effectively handles multiple light transport conditions with a single network. We achieve this by inferring a latent light transport representation from the projection image and using this representation to modulate the network that reconstructs the hidden image from the projection image. We train a light transport encoder together with a vector quantizer to obtain the light transport representation. To further regulate this representation, we jointly learn both the reconstruction network and the reprojection network during training. A set of light transport modulation blocks is used to modulate the two jointly trained networks in a multi-scale way. Extensive experiments on a large-scale passive NLOS dataset demonstrate the superiority of the proposed method. The code is available at https://github.com/JerryOctopus/NLOS-LTM.
\end{abstract}


\section{Introduction}
\IEEEPARstart{N}{on-line-of-sight (NLOS)} imaging makes it possible to image objects that are occluded by obstacles via analyzing the reflective scattering at a relay surface. 
\begin{figure}[htbp]
    \centering
    \includegraphics[width=\linewidth]{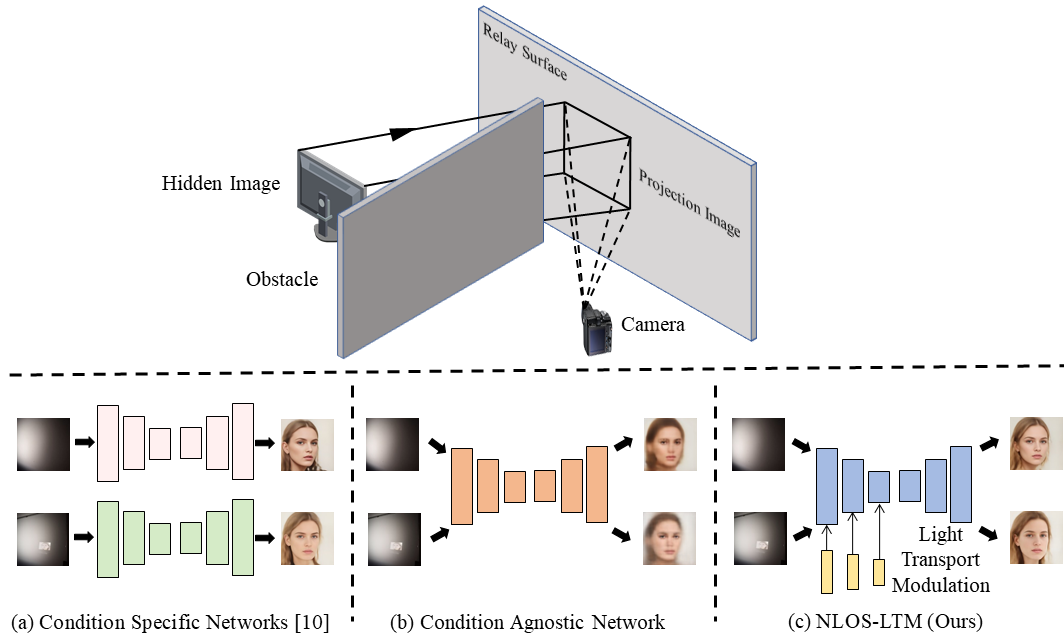}
    \caption{Top row: Passive non-line-of-sight imaging setting. Light emitted from the hidden image projects on the relay surface and is then captured by a camera. Bottom row: Comparison of three different solutions for NLOS imaging under multiple light transport conditions. (a) Train a condition specific network for each light transport condition; (b) Train a condition agnostic network without explicitly considering the light transport conditions; (c) Train a unified network for different light transport conditions, which uses a learned light transport representation to modulate the reconstruction network.
}
    \label{setup-all-in-one}
\end{figure}
With the ability to see hidden objects that are out of sight, this technology effectively extends our perception range, which is beyond the reach of traditional line-of-sight (LOS) imaging methods. NLOS imaging has promising applications in various fields such as autonomous vehicles, accident rescue, medical imaging and privacy protection, and has thus gained increasing research attention in recent years. 

NLOS imaging consists of active and passive approaches. Active NLOS imaging uses controllable light and extracts information from multiple reflections via ultra-high speed light systems \cite{kirmani2011looking, velten2012recovering}. The technology relies heavily on costly detectors and involves a time-consuming, labor-intensive calibration process \cite{cao2022computational,cao2022high, heide2019non,li2023nlost, liu2023non, chen2024hyper}. Recent work \cite{liu2019non} has explored reconstruction in bright environments, but its ability to recovery complex scenes remains limited. In contrast, passive NLOS imaging \cite{he2022non,tanaka2020polarized,yedidia2019using,zhou2020non,saunders2021fast,czajkowski2024two} employs an ordinary camera to capture the intensity of scattered light as it reaches the relay surface, eliminating the need for controllable illumination and complex detectors. Current techniques for passive NLOS imaging encompass traditional physics-based methods and deep learning-based methods. While the former is restricted to handling simple scenes, the advent of deep learning has substantially enhanced reconstruction resolution and made it possible to reconstruct complex hidden scenes. In this work, we focus on deep learning-based passive NLOS imaging methods.

Passive NLOS imaging is a highly ill-posed problem. With some mild assumptions, the discrete forward model for passive NLOS imaging can be formulated as \cite{geng2021passive}:
\begin{equation}
    \mathbf{y}=\mathbf{A x}+\mathbf{n},
    \label{y}
\end{equation} where $\mathbf{x} \in \mathbb{R}^{H_xW_x}$ is the vectorized hidden image and $H_x \times W_x$ is the resolution of it. $\mathbf{y} \in \mathbb{R}^{H_yW_y}$ is the observed projection image in vector form, and $H_y \times W_y$ is the corresponding resolution. $\mathbf{A} \in \mathbb{R}^{H_y W_y \times H_x W_x}$ denotes the light transport matrix. $\mathbf{n} \in \mathbb{R}^{H_yW_y}$ represents the noise. The close contributions between pixels caused by isotropic diffuse reflection lead to a very large condition number of the light transport matrix. This makes it difficult to obtain good reconstructions from the observations \cite{geng2021passive}. To alleviate this problem, existing strategies involve placing a partial occluder \cite{yedidia2019using}, using polarizers \cite{tanaka2020polarized} and applying deep learning \cite{zhou2020non,geng2021passive,aittala2019computational}. Among them, deep learning-based passive NLOS imaging techniques can substantially enhance the reconstruction quality \cite{geng2021passive}, \cite{geng2021recent}. 

The light transport condition plays an important role in NLOS imaging, which refers to the set of environmental and geometric parameters that affect how light is transported from the hidden object to the relay surface and then to the camera. These parameters include the distance between the hidden object and the relay surface, the angle of the camera, the type of ambient illumination, and the material of the relay surface. Variations in these conditions result in different imaging models, posing challenges to the generalization ability of learning-based reconstruction models. Existing methods, such as \cite{zhou2020non,geng2021passive}, typically train different models for different light transport conditions, which not only results in high computational loads but also fails to exploit the relations between different conditions. Besides, during testing, one needs to know under which light transport condition a projection image is taken so as to apply the appropriate model for it, which limits the practical usage of the models. One potential solution is to mix the datasets collected under different conditions and train a single condition agnostic model. A potential risk of this strategy is, the projection images that correspond to the same hidden image but captured under different light transport conditions will have different appearances, by getting through the same network they will be reconstructed differently, although we
expect to obtain the same reconstruction for them because they correspond to the same hidden image.

In this paper, we aim to develop a unified reconstruction model for passive NLOS imaging that can effectively handle multiple light transport conditions at the same time. The main idea is to infer from the projection image a latent representation of the corresponding light transport condition and use this representation to ``modulate'' the process that reconstructs the hidden image from the projection image. Here, ``modulate'' refers to the process of adjusting the features extracted by the encoder using the light transport representation. With this design, we are able to disentangle the information about the hidden scene and the information related to the light transport process in the projection image, which allows for a single model for multiple light transport conditions and better reconstruction quality. 

Inferring the light transport condition and using it to guide the reconstruction is very challenging, which requires the exploration of the light transport matrix that is very high-dimensional and lacks structures that facilitate learning. There is no explicit label information about the light transport representation available for training. And how to effectively modulate the reconstruction network using the light transport representation has never been studied before. Besides, the projection images are very fuzzy with very limited information. We propose three core designs to address the challenges.

Firstly, we present a joint learning strategy that simultaneously learns the reconstruction process from the projection image to the hidden image and the projection process from the hidden image to the projection image. The motivation for this is, a good estimation of the light transport representation should not only help the reconstruction process but also mimic the NLOS projection process to enable regenerating the projection image from the hidden image. This added network branch is an effective constraint to ensure the quality of the estimated light transport representation.

Secondly, we learn a light transport encoder that extracts a latent representation for the light transport condition. Moreover, by noting that projection images taken under the same condition should share the same light transport representation, we apply vector quantization (VQ) \cite{van2017neural} to the representation extracted by the light transport encoder. This not only lets the learned representation meet the above requirement, but also helps to reduce the noise when using the latent representation to modulate the reconstruction and the reprojection networks.

Thirdly, we use the light transport modulation blocks to inject the light transport representation and modulate the features of the projection image in the reconstruction network and the features of the hidden image in the reprojection network. The injections are carried out in a multi-scale way to expand the influence of the light transport representation and strengthen the connections between the light transport representation and the two generation networks. These designs enable our network to handle various light transport conditions effectively.
 
The main contributions of this work are summarized below:
\begin{itemize}
    \item We propose to learn a unified model for passive NLOS imaging that can effectively handle multiple light transport conditions. This is achieved by inferring a latent light transport representation from the projection image and using it to modulate the reconstruction network.
    \item We provide three critical designs to tackle the challenges in the problem, including a joint learning of both the reconstruction network and the reprojection network, a light transport encoder together with VQ to get the latent light transport representation, and the light transport modulation blocks that modulate the two generation networks in a multi-scale way. 
    \item Extensive experiments are conducted on a large-scale passive NLOS dataset. The results show that the proposed method outperforms existing passive NLOS models and several state-of-the-art image restoration methods.         
\end{itemize}  

\section{Related work}
\subsection{Passive Non-Line-of-Sight Imaging.}
Our work focuses on the 2D reconstruction problem in passive NLOS imaging. Existing methods for passive NLOS imaging usually include physics-based methods and deep learning-based methods. Physics-based methods employ strategies such as placing a partial occluder \cite{yedidia2019using, saunders2021fast, saunders2019computational}, using polarizers \cite{tanaka2020polarized,hassan2019polarization}, or exploiting the optical memory effect \cite{katz2014non} to image through opaque layers or reconstruct concealed objects. However, they can only achieve rough reconstructions of simple scenes. In recent years, deep learning-based methods have shown promising results in NLOS imaging. Based on the network design principles, deep learning methods can be divided into end-to-end networks \cite{he2022non,grau2022occlusion} and physics-based networks \cite{chen2020learned,metzler2020deep}. These methods can learn appropriate feature representations and leverage scene priors, which results in improved resolution and quality of the reconstruction image. Notably, Tancik \textit{et al.} \cite{tancik2018flash} used a Variational Auto-Encoder (VAE) \cite{kingma2013auto} for NLOS imaging. However, the model is limited to reconstructing a single specific object. Geng \textit{et al.} \cite{geng2021passive} developed NLOS-OT, a novel framework that used manifold embedding and optimal transport to map projection images to hidden images in the latent space. Besides, they built NLOS-Passive, the first public large-scale passive NLOS dataset, which facilitates the research in this field.

\subsection{Image Restoration.} 
Image restoration is a crucial task that aims to restore high-quality images by eliminating degradations from the low-quality inputs. Extensive research has been done for degradation patterns such as noise \cite{cheng2021nbnet,kawar2022denoising,nair2023ddpm,zamir2021multi}, blur \cite{vitoria2022event,zhang2022unifying,zhao2022crnet}, and adverse weather conditions \cite{song2023tusr,yang2022self,cheng2022snow,zhang2021deep,yan2022feature}. Broadly speaking, the passive NLOS imaging problem can be seen as a special image restoration problem if we take the projection process from the hidden image to the projection image as a degradation operation. However, this problem is dramatically different from traditional image restoration problems. The hidden images are distorted more severely in passive NLOS imaging than in other problems. Nevertheless, methods proposed for image restoration problems provide useful references for solving passive NLOS imaging. We therefore review some of them in the following.

The degradation representation has been taken as a crucial component in image restoration tasks. For example, learning the degradation representations has recently been explored by some learning-based deblurring methods. SelfDeblur \cite{ren2020neural} introduced the Deep Image Prior (DIP) \cite{ulyanov2018deep} to model clean images and kernels separately. It assumed the blur kernels were linear and uniform, which is impractical in real-world scenarios. To overcome this limitation, \cite{tran2021explore} introduced explicit representation for the blur kernels and operators, and used DIP to reparameterize the degradation and the sharp images. In \cite{li2022learning}, Li \textit{et al.} proposed a framework to learn spatially adaptive degradation representations of blurry images. A joint image reblurring and deblurring learning process was presented to improve the learning of the degradation representation.

For the restoration problem of removing adverse weather conditions like rain, fog and snow from images, some recent works were interested in developing a single neural network that could recover images from different types of corruptions. Li \textit{et al.} \cite{li2020all} proposed an All-in-One image restoration network for bad weather removal. It used multiple task-specific  encoders for different weather types and a generic decoder, which still resulted in a large number of parameters. To address this limitation, Valanarasu \textit{et al.} \cite{valanarasu2022transweather} proposed Transweather, a transformer-based network that has a single encoder and a decoder to restore an image degraded by any weather condition.

The motivations of some of these works are to some extent similar to this work. However, due to the distinct difference between classic image restoration problems and the passive NLOS imaging problem, directly applying image restoration models to NLOS imaging cannot achieve satisfactory results, as shown by our experiments. 

\subsection{Lensless Imaging.} 
Lensless imaging is a technique that forms images without using traditional lenses. In lensless imaging systems, the scene is either captured directly on the sensor \cite{kim2017lensless} or after being modulated by a mask element. The projection images obtained on the relay surface in NLOS imaging can seemingly be regarded as a lensless imaging process, and the projection images in NLOS also have some visual similarities to those in lensless imaging. However, despite these superficial similarities, the imaging principles and solutions of the two methods are different. NLOS imaging forms images of objects occluded by obstacles by analyzing the reflected and scattered light from
relay surfaces. Lensless imaging, on the other hand, uses optical encoders instead of traditional lenses and combines appropriate computational reconstruction algorithms to recover the scene from captured sensor measurements \cite{boominathan2022recent}. Moreover, NLOS imaging aims to reconstruct hidden images that are occluded by obstacles, addressing a highly ill-posed inverse problem. In contrast, lensless imaging typically works with a known forward model, focusing on computationally recovering the scene. These differences mean that algorithms developed for lensless imaging are generally not applicable to NLOS imaging.
\begin{figure*}[!t]
    \centering
    \includegraphics[width=\linewidth]{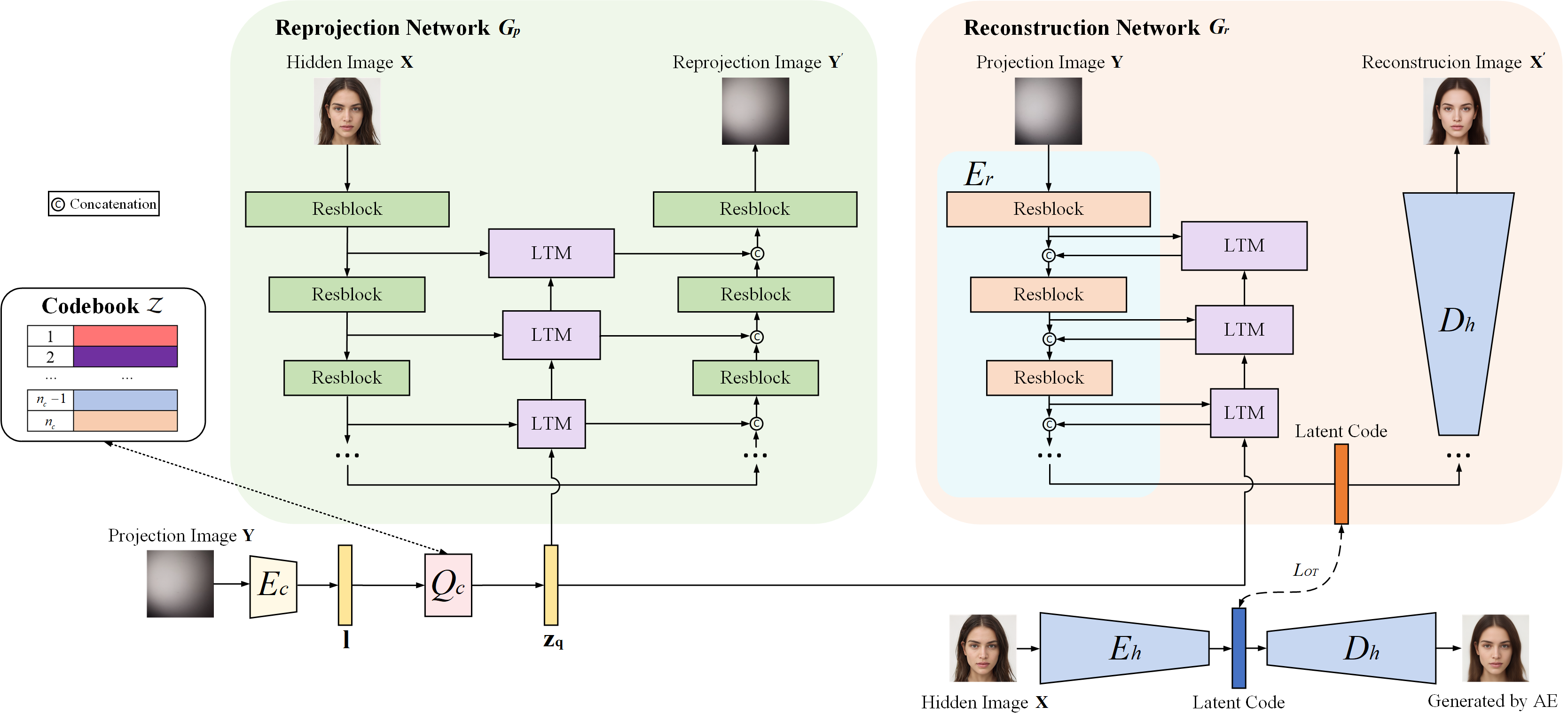}
    \caption{Architecture of the proposed NLOS-LTM method for passive NLOS imaging. A reprojection network $G_p$ that reprojects the hidden image to the projection image, and a reconstruction network $G_r$ that reconstructs the hidden image from the projection image are jointly learned during training. We use an encoder $E_c$ and a vector quantizer $Q_c$ to obtain a latent representation of the light transport condition associated with the projection image, which is then used to modulate the feature maps of the reprojection and the reconstruction networks through a set of light transport modulation (LTM) blocks. We also pretrain an auto-encoder, which consists of $E_h$ and $D_h$, with the hidden images. The decoder $D_h$ of the auto-encoder is taken as the decoder part of $G_r$. During testing, only $E_c$, $Q_c$ and $G_r$ are needed for NLOS reconstruction. 
    }
    \label{NLOS-LTM}
\end{figure*}  
\section{Methodology}
In this section, we first present the overall pipeline of the proposed NLOS-LTM for passive NLOS imaging. The details of the major components in the network are provided in the next subsections.
\subsection{Overall Pipeline}
Fig. \ref{NLOS-LTM} illustrates the architecture of the proposed NLOS-LTM network. Given a projection image $\mathbf{Y} \in \mathbb{R}^{H \times W \times 3}$, NLOS-LTM first uses an encoder $E_c$ to obtain a latent representation $E_c(\mathbf{Y})$ of the light transport condition applied to generate $\mathbf{Y}$. The projection images captured under the same light condition should share the same latent representation. To impose this important and useful constraint, we use a vector quantizer $Q_c$ to quantize the latent representation $E_c(\mathbf{Y})$ to one of the codes in a codebook, whose entries correspond to representations of different light conditions. The quantized light transport representation is then used to modulate two parallel networks $G_p$ and $G_r$. For a pair of projection image $\mathbf{Y}$ and its corresponding hidden image $\mathbf{X} \in \mathbb{R}^{H \times W \times 3}$, the reprojection network $G_p$ generates the projection image $\mathbf{Y}$ from the hidden image $\mathbf{X}$, and the reconstruction network $G_r$ reconstructs the hidden image $\mathbf{X}$ from the projection image $\mathbf{Y}$. Both networks employ an encoder-decoder structure and the light transport representation modifies the feature maps in $G_p$ and $G_r$ in a multi-scale way. The decoder of the reconstruction network $G_r$ is obtained from an auto-encoder pretrained on the hidden images. This simplifies the high-dimensional image-to-image mapping problem of $G_r$ to be mapping the projection image $\mathbf{Y}$ to the latent code of $\mathbf{X}$, which to some extent mitigates the ill-condition of the reconstruction problem.

\subsection{Learning Light Transport Representations}
We use a latent representation to characterize the light transport condition associated with a projection image. We introduce an encoder $E_c$ to encode the projection image $\mathbf{Y}$ into the light transport representation $\mathbf{l}=E_c(\mathbf{Y})$. It is evident that projection images captured under the same condition should share the same light transport representation. To regulate the learned representation with this observation, we propose a discrete latent light transport representation method. Specifically, let $\mathcal{Z}=\left\{\mathbf{z_i}\right\}_{i=1}^{n_c} $, $\mathbf{z_i} \in \mathbb{R}^{n_d}$ be a codebook where $n_c$ is the size of the codebook and corresponds to the number of different conditions considered by the model, $n_d$ is the dimensionality of each code. The discrete light transport representation for $\mathbf{Y}$ is obtained by quantizing $\mathbf{l}$ to one of the codes in $\mathcal{Z}$ as shown in Eq. \eqref{z_q}. 
\begin{equation}
    \mathbf{z_{q}}=Q_c(\mathbf{l})=\left\{\begin{array}{cc}\mathbf{z_{+}}, & \text {Training},  \\ arg \min _{\mathbf{z}_{\mathbf{i}} \in \mathcal{Z}}\left\|\mathbf{l}-\mathbf{z}_{\mathbf{i}}\right\|_2, & \text {Testing}.\end{array}\right.
    \label{z_q}
\end{equation}
The quantization operation during training and testing is slightly different. During training, we use the label information about the light transport condition. $\mathbf{z_{+}}$ is the code in the codebook that corresponds to the light transport condition $\mathbf{Y}$ is taken under. During testing, we do not need to know which condition a projection image belongs to. We just perform nearest neighbor look-up in $\mathcal{Z}$ to get the corresponding latent condition representation. The discrete representation $\mathbf{z_{q}}$ is then used to modulate the reprojection network $G_p$ and reconstruction network $G_r$.

The loss function for learning is given in Eq. \eqref{L_VQ}. It has three components for different purposes. The first term is the InfoNCE loss \cite{oord2018representation}, which aims to classify the encoder output $\mathbf{l}$ into the corresponding codebook entry. $\mathbf{l}$ and $\mathbf{z_{+}}$ form a positive pair whereas $\mathbf{l}$ and all other codes in $\mathcal{Z}$ form negative pairs. $\tau$ is a temperature hyper-parameter \cite{wu2018unsupervised}. The second term is used for updating the codebook $\mathcal{Z}$, $\operatorname{sg}[\cdot]$ denotes the stop-gradient operation \cite{esser2021taming}. The third term is the commitment loss \cite{van2017neural} that makes sure the encoder $E_c$ commits to a codebook entry. $\alpha$ and $\beta$ are the weighting factors.
\begin{equation}
    \begin{aligned}
        & L_{VQ}=-\log \frac{\exp \left(\mathbf{l} \cdot \mathbf{z_{+}} / \tau\right)}{\sum_{i=1}^{n_c} \exp (\mathbf{l} \cdot \mathbf{z_i} / \tau)} \\
        & \ \ \ \ \qquad+\alpha\left\|\operatorname{sg}[\mathbf{l}]-\mathbf{z_+}\right\|_2^2+\beta\left\|\operatorname{sg}\left[\mathbf{z_+}\right]-\mathbf{l}\right\|_2^2.
    \end{aligned}
    \label{L_VQ}
\end{equation}

By introducing vector quantization, we ensure that hidden images captured under the same light transport condition have the same latent representation for the condition. It also makes the model concentrate on important features of the light condition as opposed to focusing on noise and imperceptible details. The adding of the contrastive loss in Eq. \eqref{L_VQ} actually extends standard VQ which is usually an unsupervised process to a supervised version. Although VQ is not a new tool, applying it to the extraction of light transport representation is a brand new usage that has not been exploited before.  
\begin{figure}[!t]
    \centering
    \includegraphics[width=0.7\linewidth]{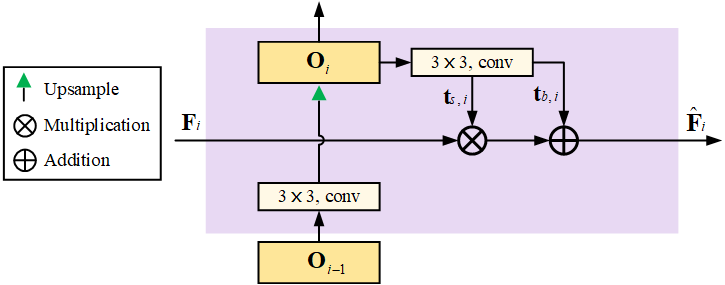}
    \caption{Detailed structure of the light transport modulation (LTM) block. $\mathbf{O}_{i}$ is the upsampled light transport representation of different scales. $\mathbf{F}_{i}$ is the input feature map. $\mathbf{t}_{s,i}$ and $\mathbf{t}_{b,i}$ are the scaling and translation parameters respectively. The LTM block modulates the input feature map with a normalization style operation.}
    \label{LTM}
\end{figure}
\subsection{Image Reconstruction Network}
The reconstruction network $G_r$ predicts the hidden image $\mathbf{X}$ from the projection image $\mathbf{Y}$ with the help of the latent light transport representation estimated from $\mathbf{Y}$. This can be formulated as: 
\begin{equation}
    \mathbf{X}^{\prime}=G_r(\mathbf{Y} , {Q_c}(E_c(\mathbf{Y}))).
\end{equation} The integration of light transport representation enables $G_r$ to achieve reconstruction for multiple light transport conditions and also enhances the overall quality of the reconstructed image.

\textit{1) Light Transport Modulation:} 
The reconstruction network $G_r$ takes an encoder-decoder structure. The aim of the encoder in $G_r$ is to extract semantic information from the projection image that is irrelevant to the light transport conditions. To achieve this, we use the light transport representation to modulate the encoder. To make full use of the light transport representation $\mathbf{z_q}$, we upsample it at multiple scales to act on the features of the projection image across different resolutions. Let $\mathbf{F}_{i} \in \mathbb{R}^{H_i \times W_i \times C_i}$ be the feature map of the projection image $\mathbf{Y}$ extracted by the encoder of $G_r$ at scale $i$ and $\mathbf{O}_i$ be the upsampled light transport representation at the same scale, which is obtained by convolution-based upsampling the representation $\mathbf{O}_{i-1} \in \mathbb{R}^{H_i/2 \times W_i/2 \times C_{i-1}}$ of the previous scale. A block called light transport modulation (LTM) is designed to modulate $\mathbf{F}_{i}$ through a normalization style operation \cite{huang2017arbitrary}, \cite{karras2019style}. As shown in Fig. \ref{LTM}, we use a convolution layer to extract the modulating parameters $\mathbf{t}_i=\left(\mathbf{t}_{i, s}, \mathbf{t}_{i, b}\right)$ from $\mathbf{O}_i$, which modulate $\mathbf{F}_{i}$ with scaling and additions as:
\begin{equation}
    \hat{\mathbf{F}_i}=\mathbf{t}_{i, s} \frac{\mathbf{F}_i-\mu\left(\mathbf{F}_i\right)}{\sigma\left(\mathbf{F}_i\right)}+\mathbf{t}_{i, b},
\end{equation}
where $\operatorname{\mu}(\cdot)$ and $\operatorname{\sigma}(\cdot)$ are the mean and the standard deviation respectively. 

The modulated feature $\hat{\mathbf{F}_i}$ is then concatenated with $\mathbf{F}_i$ and the concatenated representation is taken as the input of the next convolution block of the image encoder. With the help of the LTM blocks, the encoder is encouraged to disentangle the information related to the light transport process in the projection image and only keep the semantic information related to the hidden image. Injection at multiple scales strengthens the connections between the latent light transport representation and the reconstruction network. 

The output of the encoder of $G_r$ is a latent representation of the projection image. Instead of directly learning a decoder to map this latent representation to the estimated hidden image, we follow NLOS-OT \cite{geng2021passive} to get the decoder in an indirect way. The idea is to first train an auto-encoder through self-supervised learning for the hidden images. The encoder part of the auto-encoder $E_h$ maps a hidden image into a latent code and the decoder part $D_h$ restores the input hidden image from the latent code. Then the problem of reconstructing the hidden image from the projection image can be decomposed into first mapping the projection image to the latent code of the target hidden image and then using the decoder $D_h$ of the auto-encoder to get the reconstructed hidden image. This reduces the original ill-conditioned high-dimensional to high-dimensional mapping problem to a high-dimensional to low-dimensional mapping problem and results in better reconstruction quality. Note that the NLOS-OT method did not explicitly model and use the light transport condition in its encoder for the projection image, so it needed to train different models for different light conditions. Our designs enable training an effective unified model for different conditions.

\textit{2) Loss Function:}
The loss function for image reconstruction is a combination of the $L_1$ loss and the optimal transport (OT) loss \cite{geng2021passive}:
\begin{equation}
    L_r=L_1+\lambda_1 L_{O T}=\|\mathbf{X}-\mathbf{X}^{\prime}\|_1+\lambda_1 \|E_h(\mathbf{X})-E_r(\mathbf{Y})\|_1.
    \label{L_r}
\end{equation}
The first term measures the pixel-wise distance between the reconstructed image $\mathbf{X}^{\prime}$ and the target hidden image $\mathbf{X}$ using the $L_1$-norm. The second term is the OT loss based on an $L_1$-norm \cite{geng2021passive}, measuring distance in latent space. $E_h(\mathbf{X})$ is the encoder of the auto-encoder that extracts the latent code of the hidden image $\mathbf{X}$. $E_r(\mathbf{Y})$ is the encoder of $G_r$ that gets the latent representation of the projection image $\mathbf{Y}$. This loss drives $E_r(\mathbf{Y})$ to be similar to $E_h(\mathbf{X})$, i.e., to remove the part in $\mathbf{Y}$ that is related to the NLOS light transport process with the help of the estimated light transport representation and the LTM blocks.

\subsection{Image Reprojection Network}
The image reprojection network $G_p$ models the projecting process by learning to generate the projection image from the hidden image and the corresponding light transport representation. The whole process is formulated as:
\begin{equation}
    \mathbf{Y}^{\prime}=G_p(\mathbf{X} , Q_c(E_c(\mathbf{Y}))).
\end{equation}
where $\mathbf{Y}^{\prime}$ is the generated reprojection image conditioned on the
hidden image $\mathbf{X}$ and the latent light condition representation $Q_c(E_c(\mathbf{Y}))$. $G_p$ works as an additional constraint to ensure the validity of the light transport representation learned.

\textit{1) Light Transport Modulation:} 
The reprojection network also takes an encoder-decoder structure. We also use the light transport representation estimated from the projection image to modulate the features of the hidden image. Again, the latent representation $\mathbf{z_q}$ is upsampled recurrently and another set of LTM blocks is used to transform the feature maps of the encoder of $G_p$. Unlike the LTM blocks in $G_r$ that work within its encoder part, the LTM blocks here work between the encoder and the decoder of $G_p$. Specifically, the modulated feature $\hat{\mathbf{F}_i}$ is passed to the decoder through a skip connection and concatenated with the decoder feature at the same scale. Although the LTM blocks in $G_p$ and $G_r$ have similar structures, their functionalities are different. The LTM blocks in $G_r$ aim to remove the influence of the light transportation on the features whereas the blocks in $G_p$ aim to exert the influence. Injecting the effect of the light transport representation at multiple scales can also be seen as a way to decompose the high-dimensional and complex light transport matrix $\mathbf{A}$ in Eq. \eqref{y}. 

\textit{2) Loss Function:} 
For image reprojection, we follow the loss function in \cite{li2022learning}, incorporating the perceptual loss \cite{johnson2016perceptual} and adversarial training \cite{goodfellow2020generative}:
\begin{equation}
    \begin{aligned}
        L_G= & -E_{\mathbf{X} \sim p_{\text {data }}} D(\mathbf{X}, \mathbf{Y}^{\prime})+\lambda_2 L_{\text {perceptual }}(\mathbf{Y}, \mathbf{Y}^{\prime}), \\
        L_D= & -E_{(\mathbf{X}, \mathbf{Y}) \sim p_{\text {data }}}[\min (0,-1+D(\mathbf{X}, \mathbf{Y}))] \\
        & -E_{\mathbf{X} \sim p_{\text {data }}}[\min (0,-1-D(\mathbf{X}, \mathbf{Y}^{\prime}))],
    \end{aligned}
\end{equation} where the discriminator
\begin{math}
    D
\end{math} is a multi-scale conditional discriminator conditioning on the hidden image
\begin{math}
    \mathbf{X}
\end{math}, assessing whether the generated reprojection image 
\begin{math}
    \mathbf{Y}^{\prime}
\end{math} has the same contents with the corresponding hidden image
\begin{math}
    \mathbf{X}
\end{math}. The weighting factor
\begin{math}
    \lambda_2
\end{math} balances the discriminator loss and the perceptual loss between the predicted projection image $\mathbf{Y}^{\prime}$ and the real $\mathbf{Y}$. The hinge loss is taken as the adversarial loss for training the discriminator.

The projection network $G_p$ is mainly used to add additional regulation to the latent light transport representation during training. It is not used at testing time which only reconstructs the hidden images from the projection images. However, as a byproduct of training, it is possible to use $G_p$ to generate new synthetic projection images, which is meaningful considering the difficulty of collecting large-scale training samples for the NLOS imaging problem.
\section{Experiments}
\subsection{Experimental Setup}
\textit{1) Datasets:}
We evaluate the performance of NLOS-LTM with the large-scale passive NLOS dataset NLOS-Passive \cite{geng2021passive}. It contains projection images captured under multiple light transport conditions by changing the distance $D$ between the hidden image and the relay surface, the angle $\angle \alpha$ of the camera, the ambient illumination and the material of the relay surface. NLOS-Passive is collected using four different types of images as the hidden images: the MNIST \cite{lecun1998gradient} dataset, the supermodel face dataset generated by Style-GAN \cite{karras2019style}, the anime face dataset DANBOORU2019 \cite{danbooru2019} and STL-10 \cite{coates2011analysis}. The multi-condition datasets we used for evaluation are created as follows. The MNIST and supermodel datasets are projected with eight different light transport conditions, termed as All-MNIST and All-Supermodel. The projections of the anime face dataset are mixed under four and eight light transport conditions, referred to as Four-Anime and All-Anime, respectively. Following \cite{geng2021passive}, we use the STL-10 dataset to analyze the generalization ability of the model. 

\textit{2) Baselines:}
We compare NLOS-LTM with the state-of-the-art passive NLOS imaging method NLOS-OT \cite{geng2021passive}. Since this reconstruction problem can be seen as a special image restoration problem, we also compare NLOS-LTM with five recent image restoration methods, including KBNet \cite{zhang2023kbnet}, NAFNet \cite{chen2022simple}, MSDINet \cite{li2022learning}, Uformer \cite{wang2022uformer} and SwinIR \cite{liang2021swinir}. 
The widely used Peak Signal-to-Noise Ratio (PSNR) and Structure Similarity (SSIM) are used to assess the reconstruction performance.
\begin{table*}[!t]
    \fontsize{8.7pt}{8.7pt}\selectfont
    \caption{Details of the datasets used with mixed light transport conditions. In the table, a light transport condition is characterized by four variables which are separated by semicolons. 70/100 is the distance between the hidden images and relay surface, measured in centimeters. Two different angles between the camera and the relay surface are used, which are \\denoted by 1/2. The ambient illumination is denoted by A/L, where A represents dark light and L means day \\light. The material of the relay surface is denoted by Wall/Wb, indicating ordinary indoor wall or \\whiteboard. The symbol ``\protect\usym{1F5F8}'' denotes the presence of the light transport condition in the mixed \\datasets, while ``\protect\usym{2715}'' denotes the absence of the condition. ``-'' indicates that the NLOS-\\Passive dataset lacks data for the light transport condition. ``Train/Test'' indicates the \\number of training and testing samples of each dataset.} 
    \begin{tabular}{ccccccccccccc}
    \toprule
    \textbf{}      & \multicolumn{12}{c}{\textbf{Light transport conditions}}                                                                                                                                                                                                                                                                                                                                                                                                                                                                                                                                                                                   \\
                   & \begin{tabular}[c]{@{}c@{}}70;1;\\ A;Wall\end{tabular} & \begin{tabular}[c]{@{}c@{}}100;1;\\ A;Wall\end{tabular} & \begin{tabular}[c]{@{}c@{}}70;2;\\ A;Wall\end{tabular} & \begin{tabular}[c]{@{}c@{}}100;2;\\ A; Wall\end{tabular} & \begin{tabular}[c]{@{}c@{}}70;1;\\ L;Wall\end{tabular} & \begin{tabular}[c]{@{}c@{}}70;2;\\ L;Wall\end{tabular} & \begin{tabular}[c]{@{}c@{}}70;1;\\ A;Wb\end{tabular} & \begin{tabular}[c]{@{}c@{}}100;1;\\ A;Wb\end{tabular} & \begin{tabular}[c]{@{}c@{}}70;2;\\ A;Wb\end{tabular} & \begin{tabular}[c]{@{}c@{}}100;2;\\ A;Wb\end{tabular} & \begin{tabular}[c]{@{}c@{}}70;1;\\ L;Wb\end{tabular} & Train/Test \\ \midrule
    All-MNIST      & \usym{1F5F8}                                                    & \usym{1F5F8}                                                      & \usym{1F5F8}                                                     & \usym{1F5F8}                                                     & -                                                     & -                                                     & \usym{1F5F8}                                                   &\usym{1F5F8}                                                    & \usym{1F5F8}                                                  & \usym{1F5F8}                                                    & -                                                   & 60,000/6,664 \\
    All-Supermodel & \usym{2715}                                                    & \usym{1F5F8}                                                     & \usym{1F5F8}                                                     & \usym{1F5F8}                                                     & -                                                     & \usym{1F5F8}                                                     & \usym{1F5F8}                                                   & \usym{2715}                                                    & \usym{1F5F8}                                                   & \usym{1F5F8}                                                    & \usym{1F5F8}                                                   & 17,776/5,328 \\
    All-Anime      & -                                                     & \usym{1F5F8}                                                      & \usym{1F5F8}                                                    & \usym{1F5F8}                                                      & \usym{1F5F8}                                                     & \usym{2715}                                                     & \usym{1F5F8}                                                   & -                                                    & \usym{1F5F8}                                                  & \usym{1F5F8}                                                   & \usym{1F5F8}                                                  & 99,336/5,768 \\
    Four-Anime     & -                                                     & \usym{1F5F8}                                                   & \usym{1F5F8}                                                    & \usym{2715}                                                      & -                                                     & \usym{2715}                                                     & \usym{1F5F8}                                                  & -                                                    & \usym{1F5F8}                                                   & \usym{2715}                                                    & \usym{2715}                                                   & 99,336/2,884 \\ \bottomrule
    \end{tabular}
    \label{Table-datasets}
\end{table*}
\begin{table*}[!t]
    \fontsize{6pt}{6pt}\selectfont
    \caption{Quantitative comparison on All-MNIST, All-Supermodel and All-Anime datasets. The best and the second best scores of the multi-condition methods are \textbf{highlighted} and \underline{underlined} respectively. }
\begin{tabular}{lccccccccccccccccc}
\toprule
\multicolumn{2}{c}{\textbf{All-MNIST}}                                         & \multicolumn{2}{c}{\textbf{\begin{tabular}[c]{@{}c@{}}70;1;\\ A;Wall\end{tabular}}}  & \multicolumn{2}{c}{\textbf{\begin{tabular}[c]{@{}c@{}}100;1;\\ A;Wall\end{tabular}}} & \multicolumn{2}{c}{\textbf{\begin{tabular}[c]{@{}c@{}}70;2;\\ A;Wall\end{tabular}}}  & \multicolumn{2}{c}{\textbf{\begin{tabular}[c]{@{}c@{}}100;2;\\ A;Wall\end{tabular}}} & \multicolumn{2}{c}{\textbf{\begin{tabular}[c]{@{}c@{}}70;1;\\ A;Wb\end{tabular}}} & \multicolumn{2}{c}{\textbf{\begin{tabular}[c]{@{}c@{}}100;1;\\ A;Wb\end{tabular}}} & \multicolumn{2}{c}{\textbf{\begin{tabular}[c]{@{}c@{}}70;2;\\ A;Wb\end{tabular}}}  & \multicolumn{2}{c}{\textbf{\begin{tabular}[c]{@{}c@{}}100;2;\\ A;Wb\end{tabular}}} \\
\multicolumn{1}{c}{\textbf{Type}}                    & \textbf{Method}         & PSNR                                      & SSIM                                    & PSNR                                     & SSIM                                     & PSNR                                      & SSIM                                    & PSNR                                     & SSIM                                     & PSNR                                    & SSIM                                   & PSNR                                    & SSIM                                    & PSNR                                     & SSIM                                   & PSNR                                    & SSIM                                    \\ \midrule
\multicolumn{1}{c}{Single Condition}                 & NLOS-OT-S               & 15.71                                     & 0.838                                   & 14.61                                    & 0.814                                    & 14.13                                     & 0.804                                   & 13.32                                    & 0.789                                    & 24.27                                   & 0.941                                  & 22.46                                   & 0.927                                   & 23.22                                    & 0.933                                  & 21.31                                   & 0.914                                   \\ \midrule
\multirow{7}{*}{Multi-Condition}                     & KBNet                   & 14.42                                     & 0.817                                   & 13.46                                    & 0.796                                    & 14.11                                     & 0.811                                   & 13.80                                    & 0.804                                    & 24.90                                   & {\ul 0.945}                            & 23.45                                   & {\ul 0.935}                             & 24.05                                    & {\ul 0.939}                            & 22.69                                   & {\ul 0.927}                             \\
                                                     & NAFNet                  & 15.29                                     & 0.580                                   & 14.27                                    & 0.573                                    & {\ul 15.47}                               & 0.587                                   & 14.18                                    & 0.658                                    & {\ul 26.79}                             & 0.780                                  & {\ul 24.14}                             & 0.870                                   & {\ul 25.62}                              & 0.802                                  & {\ul 23.30}                             & 0.863                                   \\
                                                     & MSDINet                 & 11.84                                     & 0.749                                   & 11.51                                    & 0.742                                    & 11.70                                     & 0.740                                   & 11.62                                    & 0.742                                    & 17.44                                   & 0.860                                  & 15.99                                   & 0.838                                   & 17.40                                    & 0.858                                  & 14.83                                   & 0.815                                   \\
                                                     & Uformer                 & 13.88                                     & 0.727                                   & 13.18                                    & 0.715                                    & 13.81                                     & 0.725                                   & 13.42                                    & 0.721                                    & 21.69                                   & 0.833                                  & 20.64                                   & 0.825                                   & 21.19                                    & 0.831                                  & 20.46                                   & 0.825                                   \\
                                                     & SwinIR                  & 15.55                                     & 0.687                                   & {\ul 14.47}                              & 0.654                                    & 14.74                                     & 0.650                                   & {\ul 14.67}                              & 0.648                                    & 23.10                                   & 0.873                                  & 22.20                                   & 0.863                                   & 22.61                                    & 0.868                                  & 21.58                                   & 0.851                                   \\
                                                     & NLOS-OT                 & {\ul 15.68}                               & {\ul 0.840}                             & 13.72                                    & {\ul 0.801}                              & 15.38                                     & {\ul 0.833}                             & 13.97                                    & {\ul 0.805}                              & 22.73                                   & 0.928                                  & 21.99                                   & 0.922                                   & 22.28                                    & 0.924                                  & 21.65                                   & 0.918                                   \\
                                                     & \textbf{NLOS-LTM}       & \textbf{16.27}                            & \textbf{0.844}                          & \textbf{14.74}                           & \textbf{0.813}                           & \textbf{16.17}                            & \textbf{0.841}                          & \textbf{15.03}                           & \textbf{0.820}                           & \textbf{27.29}                          & \textbf{0.957}                         & \textbf{26.39}                          & \textbf{0.953}                          & \textbf{25.90}                           & \textbf{0.949}                         & \textbf{25.43}                          & \textbf{0.944}                          \\ \hline \hline
\multicolumn{2}{c}{\textbf{All-Supermodel}}                                         & \multicolumn{2}{c}{\textbf{\begin{tabular}[c]{@{}c@{}}100;1;\\ A;Wall\end{tabular}}} & \multicolumn{2}{c}{\textbf{\begin{tabular}[c]{@{}c@{}}70;2;\\ A;Wall\end{tabular}}}  & \multicolumn{2}{c}{\textbf{\begin{tabular}[c]{@{}c@{}}100;2;\\ A;Wall\end{tabular}}} & \multicolumn{2}{c}{\textbf{\begin{tabular}[c]{@{}c@{}}70;2;\\ L;Wall\end{tabular}}}  & \multicolumn{2}{c}{\textbf{\begin{tabular}[c]{@{}c@{}}70;1;\\ A;Wb\end{tabular}}} & \multicolumn{2}{c}{\textbf{\begin{tabular}[c]{@{}c@{}}70;2;\\ A;Wb\end{tabular}}}  & \multicolumn{2}{c}{\textbf{\begin{tabular}[c]{@{}c@{}}100;2;\\ A;Wb\end{tabular}}} & \multicolumn{2}{c}{\textbf{\begin{tabular}[c]{@{}c@{}}70;1;\\ L;Wb\end{tabular}}}  \\
\multicolumn{1}{c}{\textbf{Type}}                    & \textbf{Method}         & PSNR                                      & SSIM                                    & PSNR                                     & SSIM                                     & PSNR                                      & SSIM                                    & PSNR                                     & SSIM                                     & PSNR                                    & SSIM                                   & PSNR                                    & SSIM                                    & PSNR                                     & SSIM                                   & PSNR                                    & SSIM                                    \\ \midrule
Single Condition                                     & NLOS-OT-S               & 15.50                                     & 0.589                                   & 15.68                                    & 0.589                                    & 15.42                                     & 0.586                                   & 14.45                                    & 0.554                                    & 19.47                                   & 0.683                                  & 19.40                                   & 0.679                                   & 19.83                                    & 0.690                                  & 18.99                                   & 0.670                                   \\ \midrule
\multirow{7}{*}{Multi-Condition}                     & KBNet                   & {\ul 14.96}                               & {\ul 0.630}                             & {\ul 15.85}                              & {\ul 0.670}                              & {\ul 15.25}                               & {\ul 0.652}                             & 14.70                                    & {\ul 0.651}                              & 19.03                                   & {\ul 0.740}                            & 19.01                                   & {\ul 0.746}                             & 18.94                                    & {\ul 0.741}                            & 18.85                                   & \textbf{0.738}                          \\
                                                     & NAFNet                  & {\ul 14.96}                               & 0.595                                   & 15.71                                    & 0.638                                    & 15.00                                     & 0.615                                   & 14.89                                    & 0.632                                    & {\ul 19.21}                             & 0.706                                  & {\ul 19.16}                             & 0.709                                   & {\ul 19.19}                              & 0.702                                  & {\ul 19.05}                             & 0.696                                   \\
                                                     & MSDINet                 & 14.62                                     & 0.626                                   & 15.52                                    & 0.650                                    & 14.94                                     & 0.632                                   & 14.14                                    & 0.638                                    & 17.45                                   & 0.692                                  & 17.48                                   & 0.702                                   & 17.33                                    & 0.692                                  & 16.63                                   & 0.668                                   \\
                                                     & Uformer                 & 14.74                                     & 0.559                                   & 15.67                                    & 0.581                                    & 15.23                                     & 0.573                                   & 14.69                                    & 0.553                                    & 18.91                                   & 0.634                                  & 18.80                                   & 0.635                                   & 18.23                                    & 0.626                                  & 18.65                                   & 0.628                                   \\
                                                     & SwinIR                  & 14.61                                     & 0.539                                   & 15.34                                    & 0.553                                    & 14.80                                     & 0.544                                   & 14.29                                    & 0.534                                    & 17.20                                   & 0.582                                  & 17.27                                   & 0.582                                   & 17.02                                    & 0.577                                  & 16.73                                   & 0.571                                   \\
                                                     & NLOS-OT                 & 14.86                                     & 0.541                                   & 15.42                                    & 0.555                                    & 15.00                                     & 0.542                                   & \textbf{15.07}                           & 0.544                                    & 18.35                                   & 0.658                                  & 18.24                                   & 0.654                                   & 18.40                                    & 0.658                                  & 18.15                                   & 0.653                                   \\
                                                     & \textbf{NLOS-LTM}       & \textbf{15.18}                            & \textbf{0.641}                          & \textbf{15.99}                           & \textbf{0.672}                           & \textbf{15.42}                            & \textbf{0.653}                          & {\ul 15.01}                              & \textbf{0.656}                           & \textbf{19.58}                          & \textbf{0.744}                         & \textbf{19.58}                          & \textbf{0.749}                          & \textbf{19.56}                           & \textbf{0.745}                         & \textbf{19.43}                          & {\ul 0.732}                             \\ \hline \hline
\multicolumn{2}{c}{\textbf{All-Anime}}                                         & \multicolumn{2}{c}{\textbf{\begin{tabular}[c]{@{}c@{}}100;1;\\ A;Wall\end{tabular}}} & \multicolumn{2}{c}{\textbf{\begin{tabular}[c]{@{}c@{}}70;2;\\ A;Wall\end{tabular}}}  & \multicolumn{2}{c}{\textbf{\begin{tabular}[c]{@{}c@{}}100;2;\\ A;Wall\end{tabular}}} & \multicolumn{2}{c}{\textbf{\begin{tabular}[c]{@{}c@{}}70;1;\\ L;Wall\end{tabular}}}  & \multicolumn{2}{c}{\textbf{\begin{tabular}[c]{@{}c@{}}70;1;\\ A;Wb\end{tabular}}} & \multicolumn{2}{c}{\textbf{\begin{tabular}[c]{@{}c@{}}70;2;\\ A;Wb\end{tabular}}}  & \multicolumn{2}{c}{\textbf{\begin{tabular}[c]{@{}c@{}}100;2;\\ A;Wb\end{tabular}}} & \multicolumn{2}{c}{\textbf{\begin{tabular}[c]{@{}c@{}}70;1;\\ L;Wb\end{tabular}}}  \\
\multicolumn{1}{c}{\textbf{Type}}                    & \textbf{Method}         & PSNR                                      & SSIM                                    & PSNR                                     & SSIM                                     & PSNR                                      & SSIM                                    & PSNR                                     & SSIM                                     & PSNR                                    & SSIM                                   & PSNR                                    & SSIM                                    & PSNR                                     & SSIM                                   & PSNR                                    & SSIM                                    \\ \midrule
\multicolumn{1}{c}{Single Condition}                 & NLOS-OT-S               & 13.33                                     & 0.414                                   & 13.59                                    & 0.428                                    & 13.35                                     & 0.415                                   & 12.76                                    & 0.401                                    & 16.12                                   & 0.500                                  & 16.09                                   & 0.498                                   & 15.45                                    & 0.475                                  & 16.16                                   & 0.497                                   \\ \midrule
\multicolumn{1}{c}{\multirow{7}{*}{Multi-Condition}} & KBNet                   & 13.13                                     & {\ul 0.501}                             & 13.42                                    & {\ul 0.529}                              & 13.24                                     & {\ul0.512}                          & 12.31                                    & 0.422                                    & 15.61                                   & {\ul 0.601}                            & 15.62                                   & {\ul 0.611}                             & 14.88                                    & {\ul 0.573}                            & 15.52                                   & {\ul 0.602}                             \\
\multicolumn{1}{c}{}                                 & NAFNet                  & {\ul 13.50}                               & 0.497                                   & 13.55                                    & 0.502                                    & 13.46                                     & 0.494                                   & {\ul 12.81}                              & {\ul 0.447}                              & 16.22                                   & 0.587                                  & 15.82                                   & 0.567                                   & 15.62                                    & 0.556                                  & 16.00                                   & 0.575                                   \\
\multicolumn{1}{c}{}                                 & MSDINet                 & 12.92                                     & 0.491                                   & 13.27                                    & 0.513                                    & 13.16                                     & 0.502                                   & 12.11                                    & 0.430                                    & 14.52                                   & 0.557                                  & 14.56                                   & 0.551                                   & 14.03                                    & 0.542                                  & 14.51                                   & 0.554                                   \\
\multicolumn{1}{c}{}                                 & Uformer                 & 13.37                                     & 0.382                                   & 13.61                                    & 0.388                                    & 13.40                                     & 0.387                                   & 12.80                                    & 0.371                                    & {\ul 16.74}                             & 0.456                                  & {\ul 16.70}                             & 0.460                                   & {\ul 15.65}                              & 0.434                                  & \textbf{16.74}                          & 0.459                                   \\
\multicolumn{1}{c}{}                                 & SwinIR                  & \textbf{13.70}                            & 0.440                                   & \textbf{13.96}                           & 0.441                                    & \textbf{13.77}                            & 0.439                                   & \textbf{13.02}                           & 0.434                                    & 15.94                                   & 0.469                                  & 15.96                                   & 0.469                                   & 15.20                                    & 0.455                                  & 15.81                                   & 0.467                                   \\
\multicolumn{1}{c}{}                                 & NLOS-OT                 & 13.14                                     & 0.405                                   & 13.55                                    & 0.414                                    & 13.09                                     & 0.400                                   & {\ul 12.81}                              & 0.397                                    & 15.10                                   & 0.466                                  & 15.24                                   & 0.466                                   & 14.79                                    & 0.454                                  & 15.08                                   & 0.466                                   \\
\multicolumn{1}{c}{}                                 & \textbf{NLOS-LTM}       & 13.39                                     & \textbf{0.503}                          & {\ul 13.89}                              & \textbf{0.536}                           & {\ul 13.48}                               & \textbf{0.513}                          & 12.65                                    & \textbf{0.464}                           & \textbf{16.81}                          & \textbf{0.631}                         & \textbf{16.76}                          & \textbf{0.631}                          & \textbf{15.75}                           & \textbf{0.589}                         & {\ul 16.69}                             & \textbf{0.628}                          \\ \bottomrule
\end{tabular}
\label{Table-All}
\end{table*}
\begin{figure*}[!t]
    \centering
    \subfigure{
        \rotatebox{90}{\footnotesize{\ \ 70;1;A;Wall}}
        \begin{minipage}[t]{0.1\linewidth} 
            \centering
            \includegraphics[width=1\linewidth]{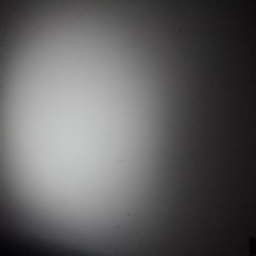}
        \end{minipage}
    }
    \hspace{-4mm}
    \subfigure{
        \begin{minipage}[t]{0.1\linewidth}
            \centering
            \includegraphics[width=1\linewidth]{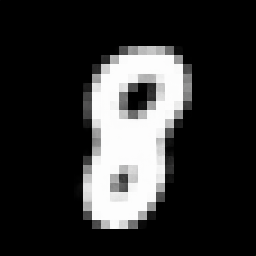}
        \end{minipage}
    }
    \hspace{-4mm}
    \subfigure{
        \begin{minipage}[t]{0.1\linewidth}
            \centering
            \includegraphics[width=1\linewidth]{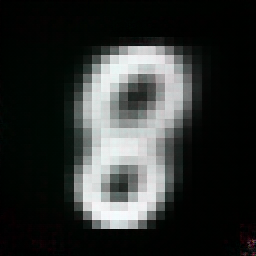}
        \end{minipage}
    }
    \hspace{-4mm}
    \subfigure{
        \begin{minipage}[t]{0.1\linewidth}
            \centering
            \includegraphics[width=1\linewidth]{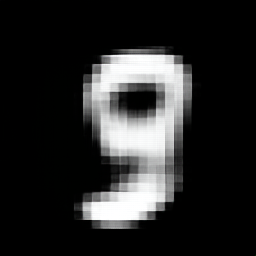}
        \end{minipage}
    }
    \hspace{-4mm}
    \subfigure{
        \begin{minipage}[t]{0.1\linewidth}
            \centering
            \includegraphics[width=1\linewidth]{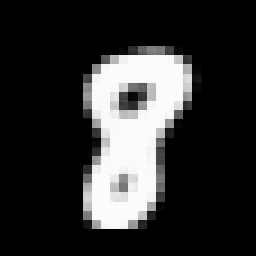}
        \end{minipage}
    }
    \hspace{-4mm}
    \subfigure{
        \begin{minipage}[t]{0.1\linewidth}
            \centering
            \includegraphics[width=1\linewidth]{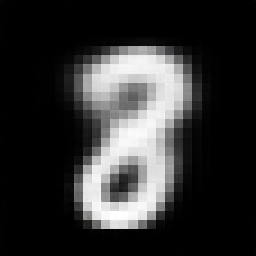}
        \end{minipage}
    }
    \hspace{-4mm}
    \subfigure{
        \begin{minipage}[t]{0.1\linewidth}
            \centering
            \includegraphics[width=1\linewidth]{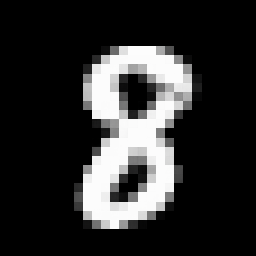}
        \end{minipage}
    }
    \hspace{-4mm}
    \subfigure{
        \begin{minipage}[t]{0.1\linewidth}
            \centering
            \includegraphics[width=1\linewidth]{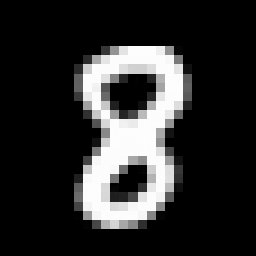}
        \end{minipage}
    }
    \hspace{-4mm}
    \subfigure{
        \begin{minipage}[t]{0.1\linewidth}
            \centering
            \includegraphics[width=1\linewidth]{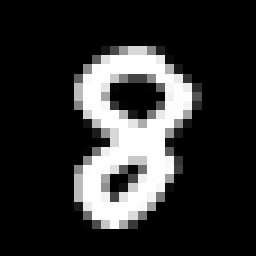}
        \end{minipage}
    }	
    
    \vspace{-2.8mm}
    \setcounter{subfigure}{0}
    
    \subfigure{
        \rotatebox{90}{\footnotesize{\ \ 100;1;A;Wall}}
        \begin{minipage}[t]{0.1\linewidth} 
            \centering
            \includegraphics[width=1\linewidth]{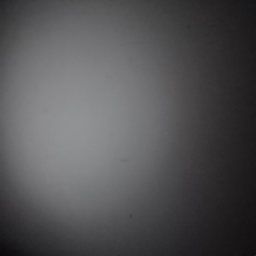}
        \end{minipage}
    }
    \hspace{-4mm}
    \subfigure{
        \begin{minipage}[t]{0.1\linewidth}
            \centering
            \includegraphics[width=1\linewidth]{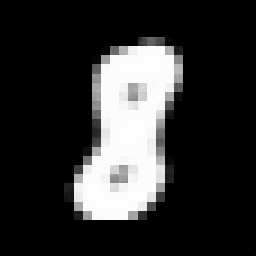}
        \end{minipage}
    }
    \hspace{-4mm}
    \subfigure{
        \begin{minipage}[t]{0.1\linewidth}
            \centering
            \includegraphics[width=1\linewidth]{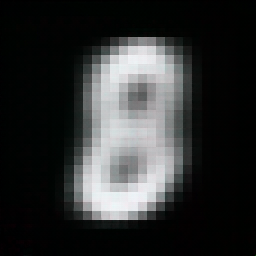}
        \end{minipage}
    }
    \hspace{-4mm}
    \subfigure{
        \begin{minipage}[t]{0.1\linewidth}
            \centering
            \includegraphics[width=1\linewidth]{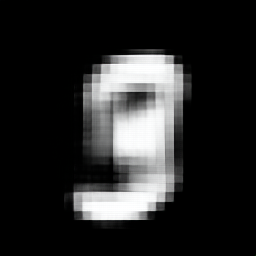}
        \end{minipage}
    }
    \hspace{-4mm}
    \subfigure{
        \begin{minipage}[t]{0.1\linewidth}
            \centering
            \includegraphics[width=1\linewidth]{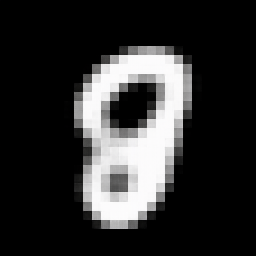}
        \end{minipage}
    }
    \hspace{-4mm}
    \subfigure{
        \begin{minipage}[t]{0.1\linewidth}
            \centering
            \includegraphics[width=1\linewidth]{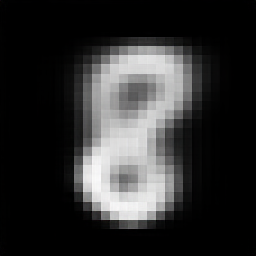}
        \end{minipage}
    }
    \hspace{-4mm}
    \subfigure{
        \begin{minipage}[t]{0.1\linewidth}
            \centering
            \includegraphics[width=1\linewidth]{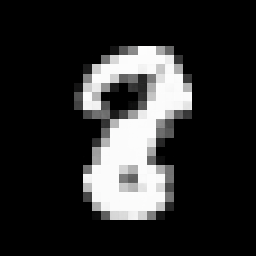}
        \end{minipage}
    }
    \hspace{-4mm}
    \subfigure{
        \begin{minipage}[t]{0.1\linewidth}
            \centering
            \includegraphics[width=1\linewidth]{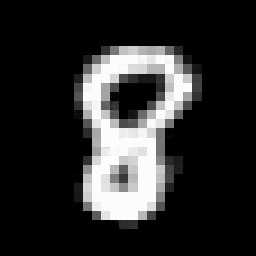}
        \end{minipage}
    }
    \hspace{-4mm}
    \subfigure{
        \begin{minipage}[t]{0.1\linewidth}
            \centering
            \includegraphics[width=1\linewidth]{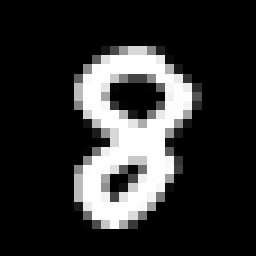}
        \end{minipage}
    }
    
    \vspace{-2.8mm}
    \setcounter{subfigure}{0}
    
    \subfigure{
        \rotatebox{90}{\footnotesize{\ \ 70;2;A;Wall}}
        \begin{minipage}[t]{0.1\linewidth} 
            \centering
            \includegraphics[width=1\linewidth]{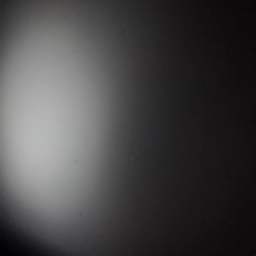}
        \end{minipage}
    }
    \hspace{-4mm}
    \subfigure{
        \begin{minipage}[t]{0.1\linewidth}
            \centering
            \includegraphics[width=1\linewidth]{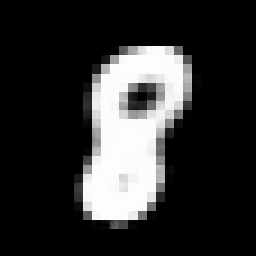}
        \end{minipage}
    }
    \hspace{-4mm}
    \subfigure{
        \begin{minipage}[t]{0.1\linewidth}
            \centering
            \includegraphics[width=1\linewidth]{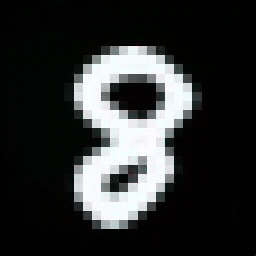}
        \end{minipage}
    }
    \hspace{-4mm}
    \subfigure{
        \begin{minipage}[t]{0.1\linewidth}
            \centering
            \includegraphics[width=1\linewidth]{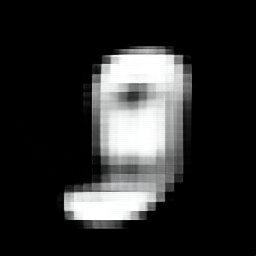}
        \end{minipage}
    }
    \hspace{-4mm}
    \subfigure{
        \begin{minipage}[t]{0.1\linewidth}
            \centering
            \includegraphics[width=1\linewidth]{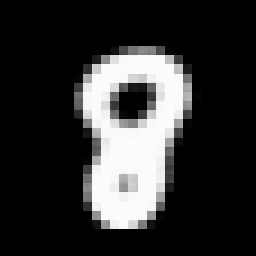}
        \end{minipage}
    }
    \hspace{-4mm}
    \subfigure{
        \begin{minipage}[t]{0.1\linewidth}
            \centering
            \includegraphics[width=1\linewidth]{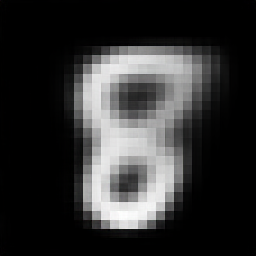}
        \end{minipage}
    }
    \hspace{-4mm}
    \subfigure{
        \begin{minipage}[t]{0.1\linewidth}
            \centering
            \includegraphics[width=1\linewidth]{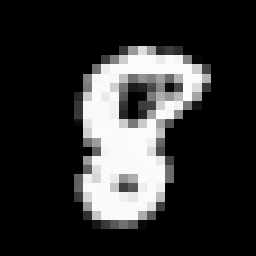}
        \end{minipage}
    }
    \hspace{-4mm}
    \subfigure{
        \begin{minipage}[t]{0.1\linewidth}
            \centering
            \includegraphics[width=1\linewidth]{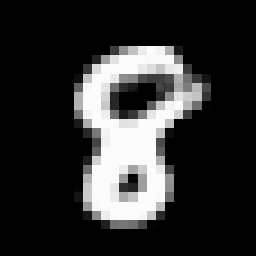}
        \end{minipage}
    }
    \hspace{-4mm}
    \subfigure{
        \begin{minipage}[t]{0.1\linewidth}
            \centering
            \includegraphics[width=1\linewidth]{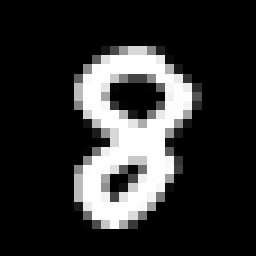}
        \end{minipage}
    }
    
    \vspace{-2.8mm}
    \setcounter{subfigure}{0}
    
    \subfigure{
        \rotatebox{90}{\footnotesize{\ \ 100;2;A;Wall}}
        \begin{minipage}[t]{0.1\linewidth} 
            \centering
            \includegraphics[width=1\linewidth]{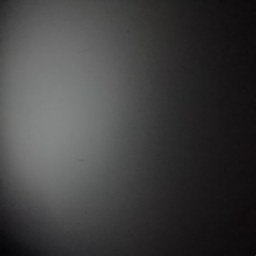}
        \end{minipage}
    }
    \hspace{-4mm}
    \subfigure{
        \begin{minipage}[t]{0.1\linewidth}
            \centering
            \includegraphics[width=1\linewidth]{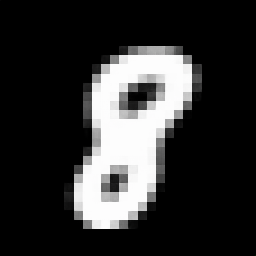}
        \end{minipage}
    }
    \hspace{-4mm}
    \subfigure{
        \begin{minipage}[t]{0.1\linewidth}
            \centering
            \includegraphics[width=1\linewidth]{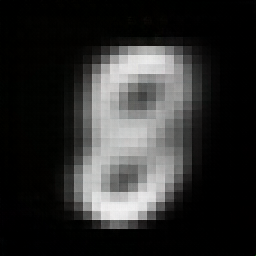}
        \end{minipage}
    }
    \hspace{-4mm}
    \subfigure{
        \begin{minipage}[t]{0.1\linewidth}
            \centering
            \includegraphics[width=1\linewidth]{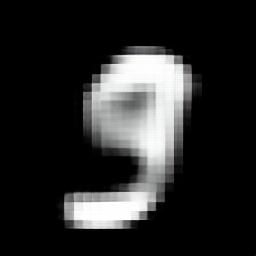}
        \end{minipage}
    }
    \hspace{-4mm}
    \subfigure{
        \begin{minipage}[t]{0.1\linewidth}
            \centering
            \includegraphics[width=1\linewidth]{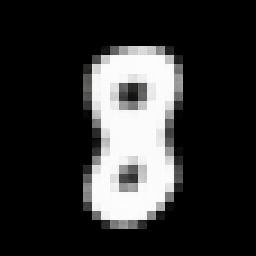}
        \end{minipage}
    }
    \hspace{-4mm}
    \subfigure{
        \begin{minipage}[t]{0.1\linewidth}
            \centering
            \includegraphics[width=1\linewidth]{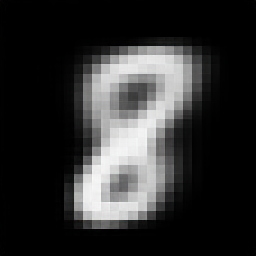}
        \end{minipage}
    }
    \hspace{-4mm}
    \subfigure{
        \begin{minipage}[t]{0.1\linewidth}
            \centering
            \includegraphics[width=1\linewidth]{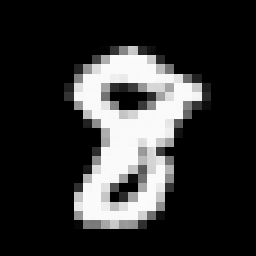}
        \end{minipage}
    }
    \hspace{-4mm}
    \subfigure{
        \begin{minipage}[t]{0.1\linewidth}
            \centering
            \includegraphics[width=1\linewidth]{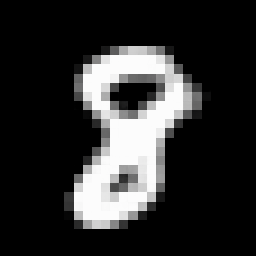}
        \end{minipage}
    }
    \hspace{-4mm}
    \subfigure{
        \begin{minipage}[t]{0.1\linewidth}
            \centering
            \includegraphics[width=1\linewidth]{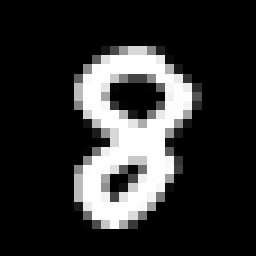}
        \end{minipage}
    }
    
    \vspace{-2.8mm}
    \setcounter{subfigure}{0}
    
    \subfigure{
        \rotatebox{90}{\footnotesize{\ \ 70;1;A;Wb}}
        \begin{minipage}[t]{0.1\linewidth} 
            \centering
            \includegraphics[width=1\linewidth]{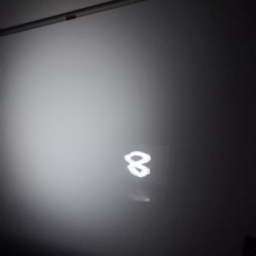}
        \end{minipage}
    }
    \hspace{-4mm}
    \subfigure{
        \begin{minipage}[t]{0.1\linewidth}
            \centering
            \includegraphics[width=1\linewidth]{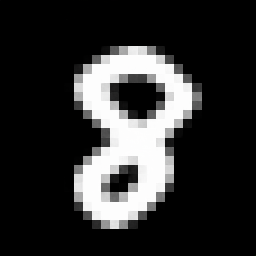}
        \end{minipage}
    }
    \hspace{-4mm}
    \subfigure{
        \begin{minipage}[t]{0.1\linewidth}
            \centering
            \includegraphics[width=1\linewidth]{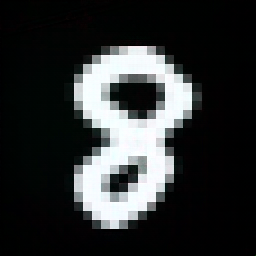}
        \end{minipage}
    }
    \hspace{-4mm}
    \subfigure{
        \begin{minipage}[t]{0.1\linewidth}
            \centering
            \includegraphics[width=1\linewidth]{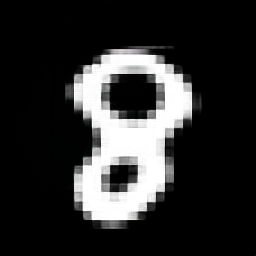}
        \end{minipage}
    }
    \hspace{-4mm}
    \subfigure{
        \begin{minipage}[t]{0.1\linewidth}
            \centering
            \includegraphics[width=1\linewidth]{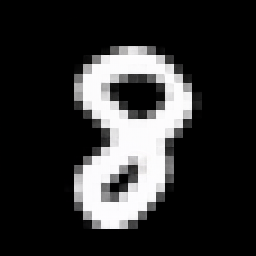}
        \end{minipage}
    }
    \hspace{-4mm}
    \subfigure{
        \begin{minipage}[t]{0.1\linewidth}
            \centering
            \includegraphics[width=1\linewidth]{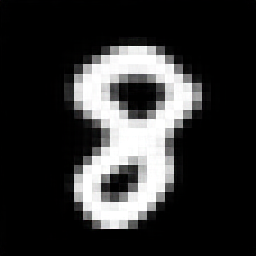}
        \end{minipage}
    }
    \hspace{-4mm}
    \subfigure{
        \begin{minipage}[t]{0.1\linewidth}
            \centering
            \includegraphics[width=1\linewidth]{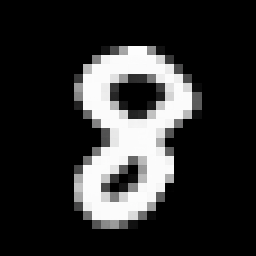}
        \end{minipage}
    }
    \hspace{-4mm}
    \subfigure{
        \begin{minipage}[t]{0.1\linewidth}
            \centering
            \includegraphics[width=1\linewidth]{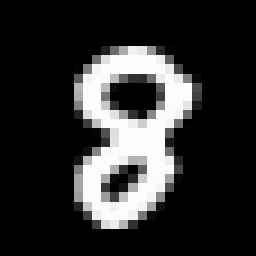}
        \end{minipage}
    }
    \hspace{-4mm}
    \subfigure{
        \begin{minipage}[t]{0.1\linewidth}
            \centering
            \includegraphics[width=1\linewidth]{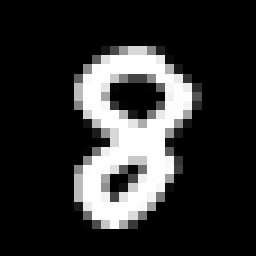}
        \end{minipage}
    }
    
    \vspace{-2.8mm}
    \setcounter{subfigure}{0}
    
    \subfigure{
        \rotatebox{90}{\footnotesize{\ \ 100;1;A;Wb}}
        \begin{minipage}[t]{0.1\linewidth} 
            \centering
            \includegraphics[width=1\linewidth]{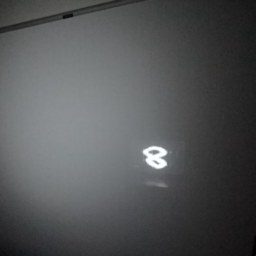}
        \end{minipage}
    }
    \hspace{-4mm}
    \subfigure{
        \begin{minipage}[t]{0.1\linewidth}
            \centering
            \includegraphics[width=1\linewidth]{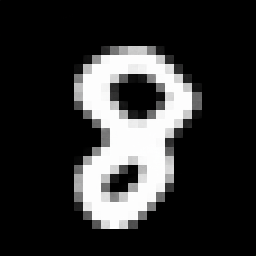}
        \end{minipage}
    }
    \hspace{-4mm}
    \subfigure{
        \begin{minipage}[t]{0.1\linewidth}
            \centering
            \includegraphics[width=1\linewidth]{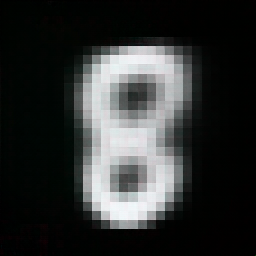}
        \end{minipage}
    }
    \hspace{-4mm}
    \subfigure{
        \begin{minipage}[t]{0.1\linewidth}
            \centering
            \includegraphics[width=1\linewidth]{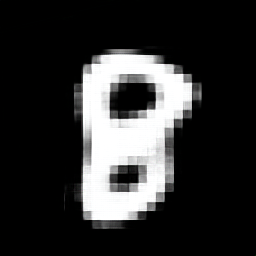}
        \end{minipage}
    }
    \hspace{-4mm}
    \subfigure{
        \begin{minipage}[t]{0.1\linewidth}
            \centering
            \includegraphics[width=1\linewidth]{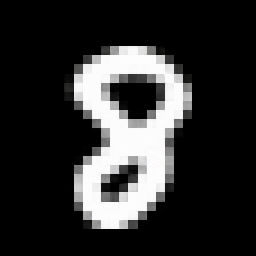}
        \end{minipage}
    }
    \hspace{-4mm}
    \subfigure{
        \begin{minipage}[t]{0.1\linewidth}
            \centering
            \includegraphics[width=1\linewidth]{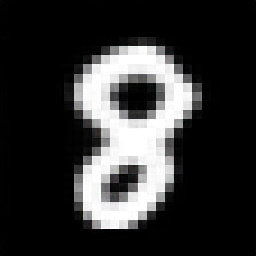}
        \end{minipage}
    }
    \hspace{-4mm}
    \subfigure{
        \begin{minipage}[t]{0.1\linewidth}
            \centering
            \includegraphics[width=1\linewidth]{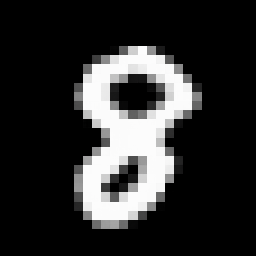}
        \end{minipage}
    }
    \hspace{-4mm}
    \subfigure{
        \begin{minipage}[t]{0.1\linewidth}
            \centering
            \includegraphics[width=1\linewidth]{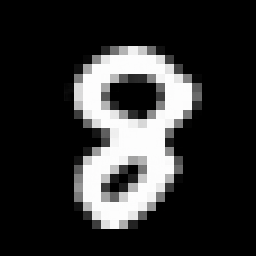}
        \end{minipage}
    }
    \hspace{-4mm}
    \subfigure{
        \begin{minipage}[t]{0.1\linewidth}
            \centering
            \includegraphics[width=1\linewidth]{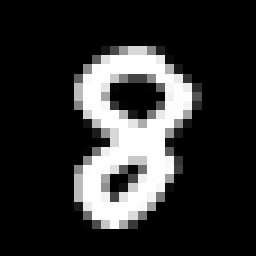}
        \end{minipage}
    }
    
    \vspace{-2.8mm}
    \setcounter{subfigure}{0}
    
    \subfigure{
        \rotatebox{90}{\footnotesize{\ \ 70;2;A;Wb}}
        \begin{minipage}[t]{0.1\linewidth} 
            \centering
            \includegraphics[width=1\linewidth]{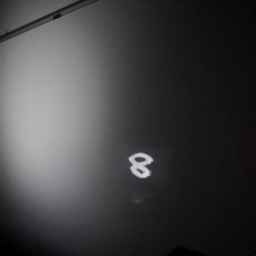}
        \end{minipage}
    }
    \hspace{-4mm}
    \subfigure{
        \begin{minipage}[t]{0.1\linewidth}
            \centering
            \includegraphics[width=1\linewidth]{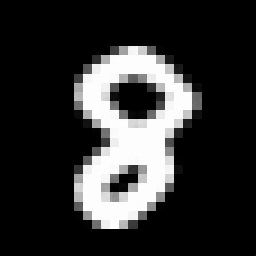}
        \end{minipage}
    }
    \hspace{-4mm}
    \subfigure{
        \begin{minipage}[t]{0.1\linewidth}
            \centering
            \includegraphics[width=1\linewidth]{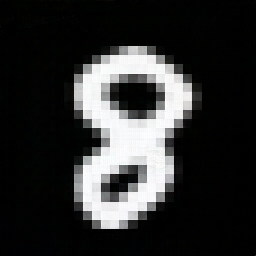}
        \end{minipage}
    }
    \hspace{-4mm}
    \subfigure{
        \begin{minipage}[t]{0.1\linewidth}
            \centering
            \includegraphics[width=1\linewidth]{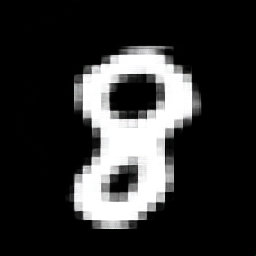}
        \end{minipage}
    }
    \hspace{-4mm}
    \subfigure{
        \begin{minipage}[t]{0.1\linewidth}
            \centering
            \includegraphics[width=1\linewidth]{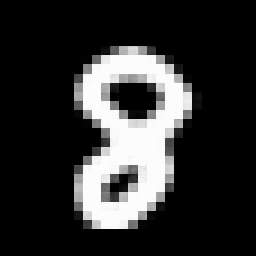}
        \end{minipage}
    }
    \hspace{-4mm}
    \subfigure{
        \begin{minipage}[t]{0.1\linewidth}
            \centering
            \includegraphics[width=1\linewidth]{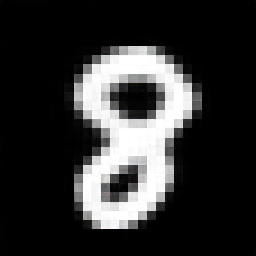}
        \end{minipage}
    }
    \hspace{-4mm}
    \subfigure{
        \begin{minipage}[t]{0.1\linewidth}
            \centering
            \includegraphics[width=1\linewidth]{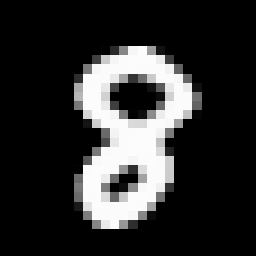}
        \end{minipage}
    }
    \hspace{-4mm}
    \subfigure{
        \begin{minipage}[t]{0.1\linewidth}
            \centering
            \includegraphics[width=1\linewidth]{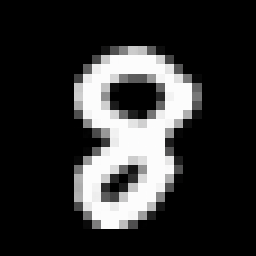}
        \end{minipage}
    }
    \hspace{-4mm}
    \subfigure{
        \begin{minipage}[t]{0.1\linewidth}
            \centering
            \includegraphics[width=1\linewidth]{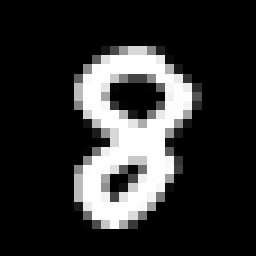}
        \end{minipage}
    }
    
    \vspace{-2.8mm}
    \setcounter{subfigure}{0}		
    
    \subfigure{
        \rotatebox{90}{\footnotesize{\ \ 100;2;A;Wb}}
        \begin{minipage}[t]{0.1\linewidth}
            \centering
            \includegraphics[width=1\linewidth]{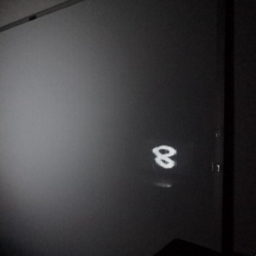}
            \centerline{\footnotesize{INPUT}}
        \end{minipage}
    }
    \hspace{-4mm}
    \subfigure{
        \begin{minipage}[t]{0.1\linewidth}
            \centering
            \includegraphics[width=1\linewidth]{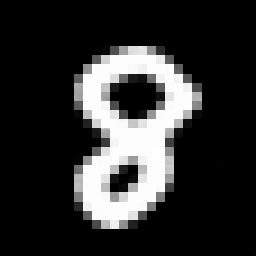}
            \centerline{\footnotesize{KBNet}}	
        \end{minipage}
    }
    \hspace{-4mm}
    \subfigure{
        \begin{minipage}[t]{0.1\linewidth}
            \centering
            \includegraphics[width=1\linewidth]{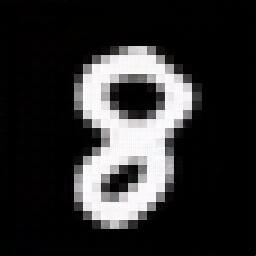}
            \centerline{\footnotesize{NAFNet}}	
        \end{minipage}
    }
    \hspace{-4mm}
    \subfigure{
        \begin{minipage}[t]{0.1\linewidth}
            \centering
            \includegraphics[width=1\linewidth]{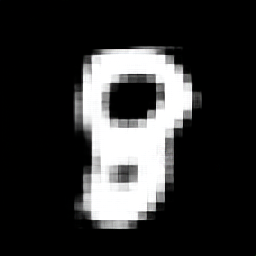}
            \centerline{\footnotesize{MSDINet}}	
        \end{minipage}
    }
    \hspace{-4mm}
    \subfigure{
        \begin{minipage}[t]{0.1\linewidth}
            \centering
            \includegraphics[width=1\linewidth]{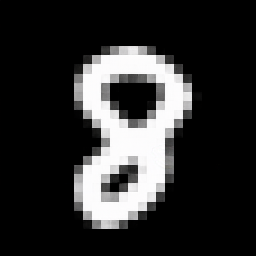}
            \centerline{\footnotesize{Uformer}}	
        \end{minipage}
    }
    \hspace{-4mm}
    \subfigure{
        \begin{minipage}[t]{0.1\linewidth}
            \centering
            \includegraphics[width=1\linewidth]{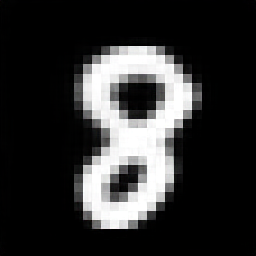}
            \centerline{\footnotesize{SwinIR}}	
        \end{minipage}
    }
    \hspace{-4mm}
    \subfigure{
        \begin{minipage}[t]{0.1\linewidth}
            \centering
            \includegraphics[width=1\linewidth]{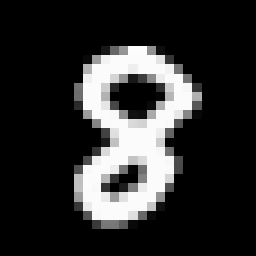}
            \centerline{\footnotesize{NLOS-OT}}	
        \end{minipage}
    }
    \hspace{-4mm}
    \subfigure{
        \begin{minipage}[t]{0.1\linewidth}
            \centering
            \includegraphics[width=1\linewidth]{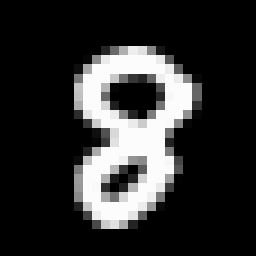}
            \centerline{\textbf{\footnotesize{NLOS-LTM}}}	
        \end{minipage}
    }
    \hspace{-4mm}
    \subfigure{
        \begin{minipage}[t]{0.1\linewidth}
            \centering
            \includegraphics[width=1\linewidth]{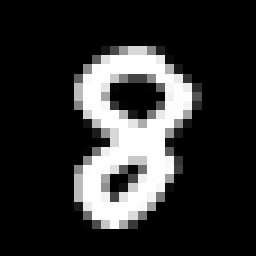}
            \centerline{\footnotesize{GT}}
        \end{minipage}
    }	
    \caption{Visual comparison of the results on the All-MNIST dataset under different light transport conditions.}
    \label{Fig-All-MNIST}
\end{figure*} 

\textit{3) Training Details:}
We use Pytorch on one NVIDIA GeForce RTX 4090 for our experiments. We label the projection images in the training set corresponding to different light transport conditions with 1, 2, 3, ..., $n_c$. Meanwhile, we randomly initialize the codebook $\mathcal{Z}$. For NLOS-LTM, we first train the auto-encoder composed of $E_h$ and $D_h$ using the hidden images. For this step, we use the Adam optimizer \cite{kingma2014adam}  (\begin{math}
    \beta_1=0.9
\end{math},
\begin{math}
    \beta_2=0.999
\end{math},
weight decay $=0$) with the initial learning rate 
\begin{math}
    1\times 10^{-4}
\end{math} gradually reduced to 
\begin{math}
    1\times 10^{-8}
\end{math} following the cosine annealing strategy \cite{loshchilov2016sgdr}. We then train the remaining parts of NLOS-LTM using the AdamW optimizer \cite{loshchilov2017decoupled} with the momentum terms of (0.9, 0.9) and the weight decay of
\begin{math}
    1\times 10^{-4}
\end{math}. The cosine decay is used to decrease the learning rate to 
\begin{math}
    1\times 10^{-7}
\end{math} with the initial learning rate 
\begin{math}
    1\times 10^{-4}
\end{math}.

\subsection{Reconstruction Results}
\textit{1) All-MNIST:}
We first compare the reconstruction results of NLOS-LTM and other baseline methods on All-MNIST and report the results in Table \ref{Table-All}. As shown in the table, NLOS-LTM achieves the best performance in all light transport conditions. There are big margins between NLOS-LTM and the second-best results on both PSNR and SSIM measures. Some examples for visual comparisons are shown in Fig. \ref{Fig-All-MNIST}, which demonstrates that the reconstruction results of NLOS-LTM are the closest to the ground truth, especially in terms of details. In contrast, image restoration methods produce blurry images or even incorrect semantic contents. Although NLOS-OT generates better images than other image restoration methods, for challenging conditions, i.e., using wall as the relay surface, there are still many artifacts in the reconstructions. NLOS-LTM reconstructs the hidden images well even under challenging conditions.

\begin{figure*}[!t]
    \centering
    \subfigure{
        \rotatebox{90}{\footnotesize{\ \ 100;1;A;Wall}}
        \begin{minipage}[t]{0.1\linewidth} 
            \centering
            \includegraphics[width=1\linewidth]{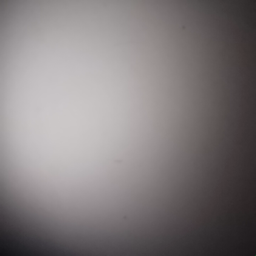}
        \end{minipage}
    }
    \hspace{-4mm}
    \subfigure{
        \begin{minipage}[t]{0.1\linewidth}
            \centering
            \includegraphics[width=1\linewidth]{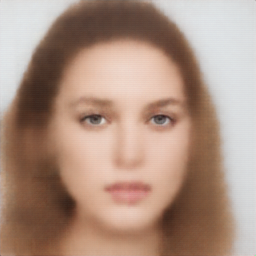}
        \end{minipage}
    }
    \hspace{-4mm}
    \subfigure{
        \begin{minipage}[t]{0.1\linewidth}
            \centering
            \includegraphics[width=1\linewidth]{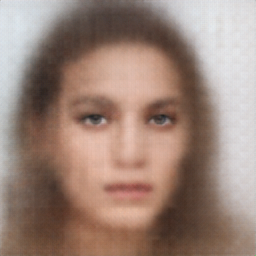}
        \end{minipage}
    }
    \hspace{-4mm}
    \subfigure{
        \begin{minipage}[t]{0.1\linewidth}
            \centering
            \includegraphics[width=1\linewidth]{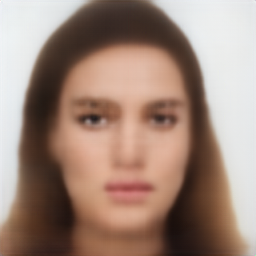}
        \end{minipage}
    }
    \hspace{-4mm}
    \subfigure{
        \begin{minipage}[t]{0.1\linewidth}
            \centering
            \includegraphics[width=1\linewidth]{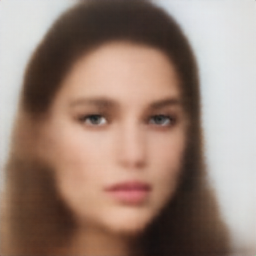}
        \end{minipage}
    }
    \hspace{-4mm}
    \subfigure{
        \begin{minipage}[t]{0.1\linewidth}
            \centering
            \includegraphics[width=1\linewidth]{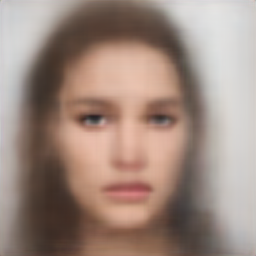}
        \end{minipage}
    }
    \hspace{-4mm}
    \subfigure{
        \begin{minipage}[t]{0.1\linewidth}
            \centering
            \includegraphics[width=1\linewidth]{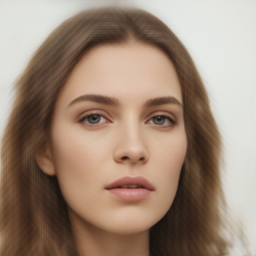}
        \end{minipage}
    }
    \hspace{-4mm}
    \subfigure{
        \begin{minipage}[t]{0.1\linewidth}
            \centering
            \includegraphics[width=1\linewidth]{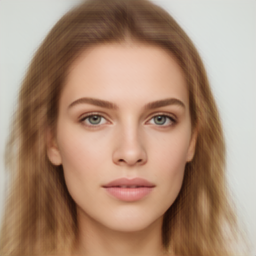}
        \end{minipage}
    }
    \hspace{-4mm}
    \subfigure{
        \begin{minipage}[t]{0.1\linewidth}
            \centering
            \includegraphics[width=1\linewidth]{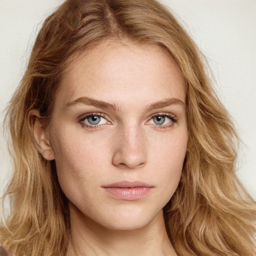}
        \end{minipage}
    }	
    
    \vspace{-2.8mm}
    \setcounter{subfigure}{0}
    
    \subfigure{
        \rotatebox{90}{\footnotesize{\ \ 70;2;A;Wall}}
        \begin{minipage}[t]{0.1\linewidth} 
            \centering
            \includegraphics[width=1\linewidth]{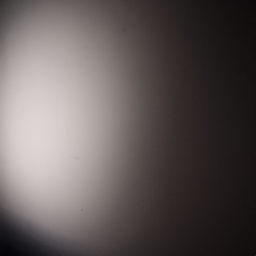}
        \end{minipage}
    }
    \hspace{-4mm}
    \subfigure{
        \begin{minipage}[t]{0.1\linewidth}
            \centering
            \includegraphics[width=1\linewidth]{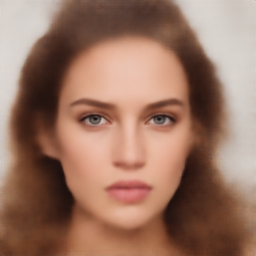}
        \end{minipage}
    }
    \hspace{-4mm}
    \subfigure{
        \begin{minipage}[t]{0.1\linewidth}
            \centering
            \includegraphics[width=1\linewidth]{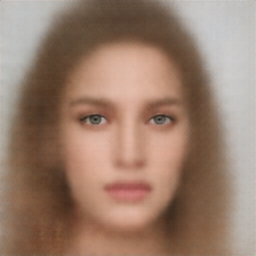}
        \end{minipage}
    }
    \hspace{-4mm}
    \subfigure{
        \begin{minipage}[t]{0.1\linewidth}
            \centering
            \includegraphics[width=1\linewidth]{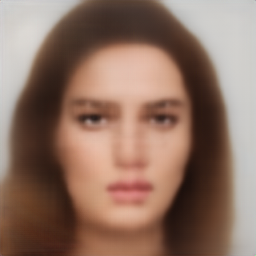}
        \end{minipage}
    }
    \hspace{-4mm}
    \subfigure{
        \begin{minipage}[t]{0.1\linewidth}
            \centering
            \includegraphics[width=1\linewidth]{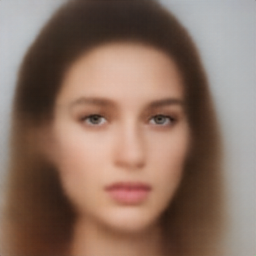}
        \end{minipage}
    }
    \hspace{-4mm}
    \subfigure{
        \begin{minipage}[t]{0.1\linewidth}
            \centering
            \includegraphics[width=1\linewidth]{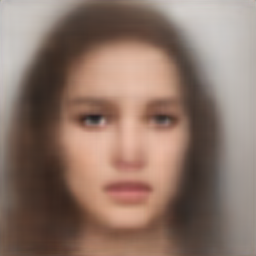}
        \end{minipage}
    }
    \hspace{-4mm}
    \subfigure{
        \begin{minipage}[t]{0.1\linewidth}
            \centering
            \includegraphics[width=1\linewidth]{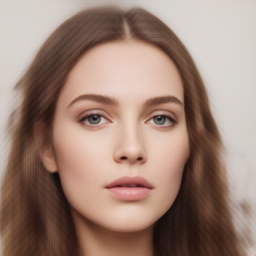}
        \end{minipage}
    }
    \hspace{-4mm}
    \subfigure{
        \begin{minipage}[t]{0.1\linewidth}
            \centering
            \includegraphics[width=1\linewidth]{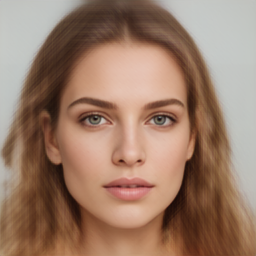}
        \end{minipage}
    }
    \hspace{-4mm}
    \subfigure{
        \begin{minipage}[t]{0.1\linewidth}
            \centering
            \includegraphics[width=1\linewidth]{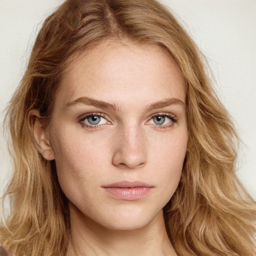}
        \end{minipage}
    }
    
    \vspace{-2.8mm}
    \setcounter{subfigure}{0}
    
    \subfigure{
        \rotatebox{90}{\footnotesize{\ \ 100;2;A;Wall}}
        \begin{minipage}[t]{0.1\linewidth} 
            \centering
            \includegraphics[width=1\linewidth]{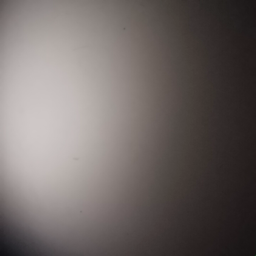}
        \end{minipage}
    }
    \hspace{-4mm}
    \subfigure{
        \begin{minipage}[t]{0.1\linewidth}
            \centering
            \includegraphics[width=1\linewidth]{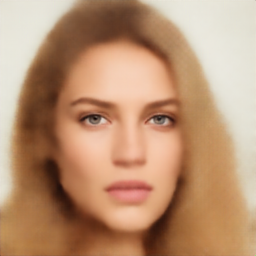}
        \end{minipage}
    }
    \hspace{-4mm}
    \subfigure{
        \begin{minipage}[t]{0.1\linewidth}
            \centering
            \includegraphics[width=1\linewidth]{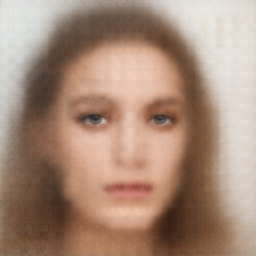}
        \end{minipage}
    }
    \hspace{-4mm}
    \subfigure{
        \begin{minipage}[t]{0.1\linewidth}
            \centering
            \includegraphics[width=1\linewidth]{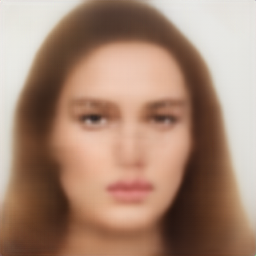}
        \end{minipage}
    }
    \hspace{-4mm}
    \subfigure{
        \begin{minipage}[t]{0.1\linewidth}
            \centering
            \includegraphics[width=1\linewidth]{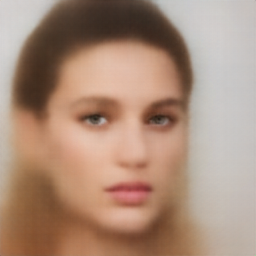}
        \end{minipage}
    }
    \hspace{-4mm}
    \subfigure{
        \begin{minipage}[t]{0.1\linewidth}
            \centering
            \includegraphics[width=1\linewidth]{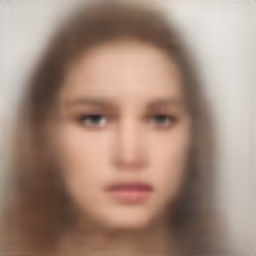}
        \end{minipage}
    }
    \hspace{-4mm}
    \subfigure{
        \begin{minipage}[t]{0.1\linewidth}
            \centering
            \includegraphics[width=1\linewidth]{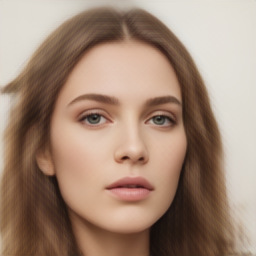}
        \end{minipage}
    }
    \hspace{-4mm}
    \subfigure{
        \begin{minipage}[t]{0.1\linewidth}
            \centering
            \includegraphics[width=1\linewidth]{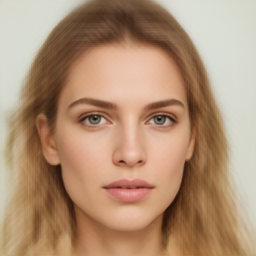}
        \end{minipage}
    }
    \hspace{-4mm}
    \subfigure{
        \begin{minipage}[t]{0.1\linewidth}
            \centering
            \includegraphics[width=1\linewidth]{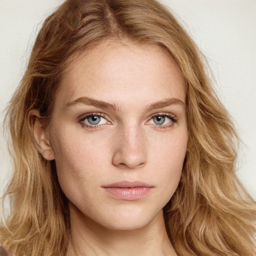}
        \end{minipage}
    }
    
    \vspace{-2.8mm}
    \setcounter{subfigure}{0}
    
    \subfigure{
        \rotatebox{90}{\footnotesize{\ \ 70;2;L;Wall}}
        \begin{minipage}[t]{0.1\linewidth} 
            \centering
            \includegraphics[width=1\linewidth]{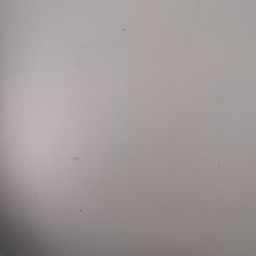}
        \end{minipage}
    }
    \hspace{-4mm}
    \subfigure{
        \begin{minipage}[t]{0.1\linewidth}
            \centering
            \includegraphics[width=1\linewidth]{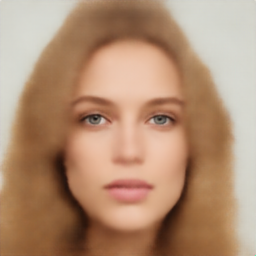}
        \end{minipage}
    }
    \hspace{-4mm}
    \subfigure{
        \begin{minipage}[t]{0.1\linewidth}
            \centering
            \includegraphics[width=1\linewidth]{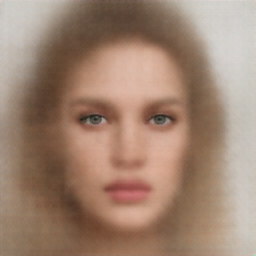}
        \end{minipage}
    }
    \hspace{-4mm}
    \subfigure{
        \begin{minipage}[t]{0.1\linewidth}
            \centering
            \includegraphics[width=1\linewidth]{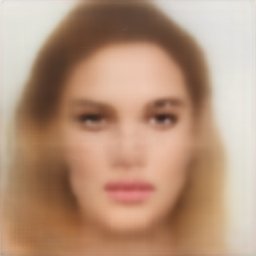}
        \end{minipage}
    }
    \hspace{-4mm}
    \subfigure{
        \begin{minipage}[t]{0.1\linewidth}
            \centering
            \includegraphics[width=1\linewidth]{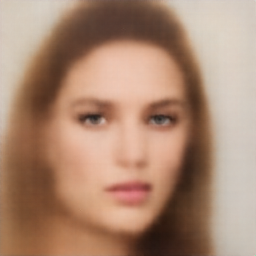}
        \end{minipage}
    }
    \hspace{-4mm}
    \subfigure{
        \begin{minipage}[t]{0.1\linewidth}
            \centering
            \includegraphics[width=1\linewidth]{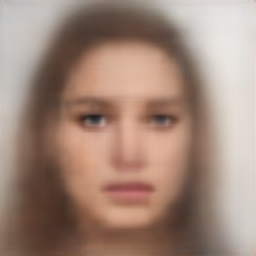}
        \end{minipage}
    }
    \hspace{-4mm}
    \subfigure{
        \begin{minipage}[t]{0.1\linewidth}
            \centering
            \includegraphics[width=1\linewidth]{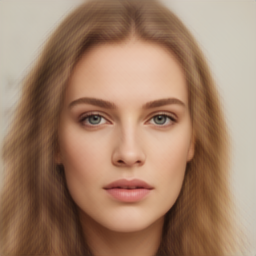}
        \end{minipage}
    }
    \hspace{-4mm}
    \subfigure{
        \begin{minipage}[t]{0.1\linewidth}
            \centering
            \includegraphics[width=1\linewidth]{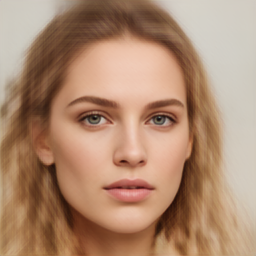}
        \end{minipage}
    }
    \hspace{-4mm}
    \subfigure{
        \begin{minipage}[t]{0.1\linewidth}
            \centering
            \includegraphics[width=1\linewidth]{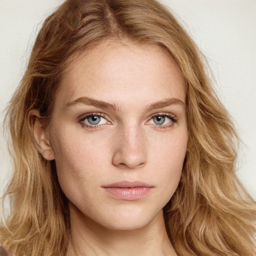}
        \end{minipage}
    }
    
    \vspace{-2.8mm}
    \setcounter{subfigure}{0}
    
    \subfigure{
        \rotatebox{90}{\footnotesize{\ \ 70;1;A;Wb}}
        \begin{minipage}[t]{0.1\linewidth} 
            \centering
            \includegraphics[width=1\linewidth]{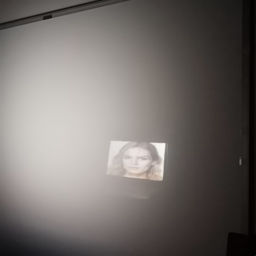}
        \end{minipage}
    }
    \hspace{-4mm}
    \subfigure{
        \begin{minipage}[t]{0.1\linewidth}
            \centering
            \includegraphics[width=1\linewidth]{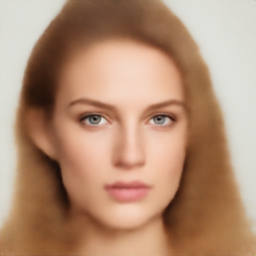}
        \end{minipage}
    }
    \hspace{-4mm}
    \subfigure{
        \begin{minipage}[t]{0.1\linewidth}
            \centering
            \includegraphics[width=1\linewidth]{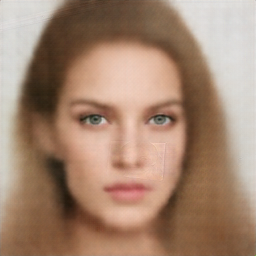}
        \end{minipage}
    }
    \hspace{-4mm}
    \subfigure{
        \begin{minipage}[t]{0.1\linewidth}
            \centering
            \includegraphics[width=1\linewidth]{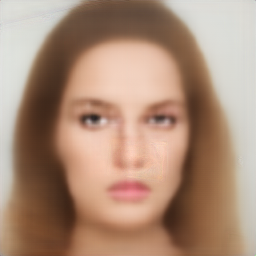}
        \end{minipage}
    }
    \hspace{-4mm}
    \subfigure{
        \begin{minipage}[t]{0.1\linewidth}
            \centering
            \includegraphics[width=1\linewidth]{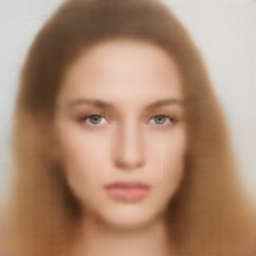}
        \end{minipage}
    }
    \hspace{-4mm}
    \subfigure{
        \begin{minipage}[t]{0.1\linewidth}
            \centering
            \includegraphics[width=1\linewidth]{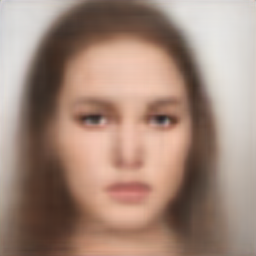}
        \end{minipage}
    }
    \hspace{-4mm}
    \subfigure{
        \begin{minipage}[t]{0.1\linewidth}
            \centering
            \includegraphics[width=1\linewidth]{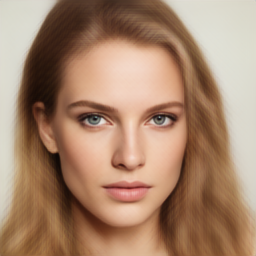}
        \end{minipage}
    }
    \hspace{-4mm}
    \subfigure{
        \begin{minipage}[t]{0.1\linewidth}
            \centering
            \includegraphics[width=1\linewidth]{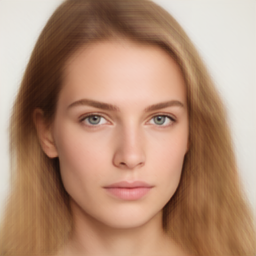}
        \end{minipage}
    }
    \hspace{-4mm}
    \subfigure{
        \begin{minipage}[t]{0.1\linewidth}
            \centering
            \includegraphics[width=1\linewidth]{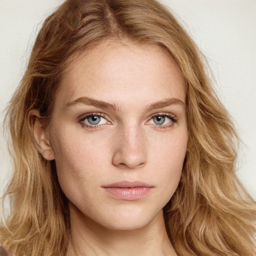}
        \end{minipage}
    }
    
    \vspace{-2.8mm}
    \setcounter{subfigure}{0}
    
    \subfigure{
        \rotatebox{90}{\footnotesize{\ \ 70;2;A;Wb}}
        \begin{minipage}[t]{0.1\linewidth} 
            \centering
            \includegraphics[width=1\linewidth]{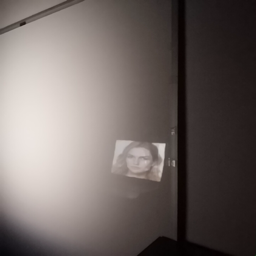}
        \end{minipage}
    }
    \hspace{-4mm}
    \subfigure{
        \begin{minipage}[t]{0.1\linewidth}
            \centering
            \includegraphics[width=1\linewidth]{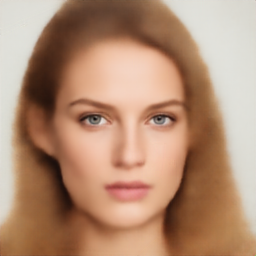}
        \end{minipage}
    }
    \hspace{-4mm}
    \subfigure{
        \begin{minipage}[t]{0.1\linewidth}
            \centering
            \includegraphics[width=1\linewidth]{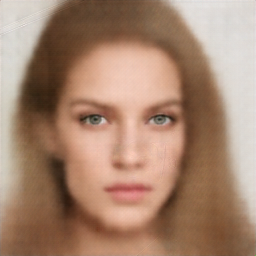}
        \end{minipage}
    }
    \hspace{-4mm}
    \subfigure{
        \begin{minipage}[t]{0.1\linewidth}
            \centering
            \includegraphics[width=1\linewidth]{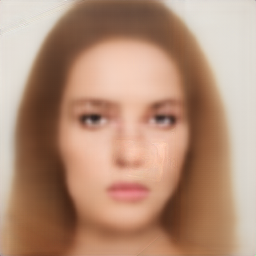}
        \end{minipage}
    }
    \hspace{-4mm}
    \subfigure{
        \begin{minipage}[t]{0.1\linewidth}
            \centering
            \includegraphics[width=1\linewidth]{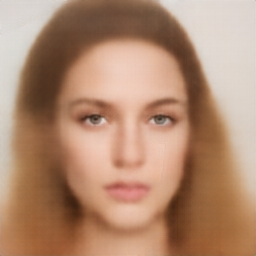}
        \end{minipage}
    }
    \hspace{-4mm}
    \subfigure{
        \begin{minipage}[t]{0.1\linewidth}
            \centering
            \includegraphics[width=1\linewidth]{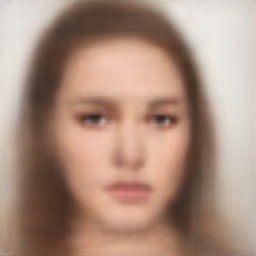}
        \end{minipage}
    }
    \hspace{-4mm}
    \subfigure{
        \begin{minipage}[t]{0.1\linewidth}
            \centering
            \includegraphics[width=1\linewidth]{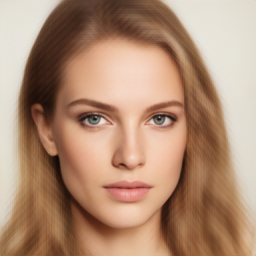}
        \end{minipage}
    }
    \hspace{-4mm}
    \subfigure{
        \begin{minipage}[t]{0.1\linewidth}
            \centering
            \includegraphics[width=1\linewidth]{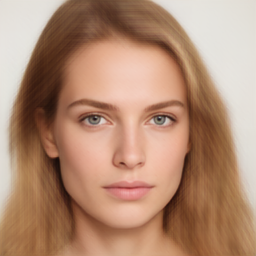}
        \end{minipage}
    }
    \hspace{-4mm}
    \subfigure{
        \begin{minipage}[t]{0.1\linewidth}
            \centering
            \includegraphics[width=1\linewidth]{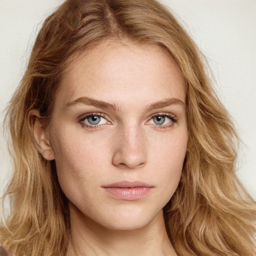}
        \end{minipage}
    }
    
    \vspace{-2.8mm}
    \setcounter{subfigure}{0}
    
    \subfigure{
        \rotatebox{90}{\footnotesize{\ \ 100;2;A;Wb}}
        \begin{minipage}[t]{0.1\linewidth} 
            \centering
            \includegraphics[width=1\linewidth]{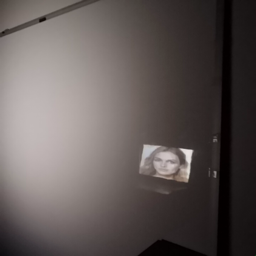}
        \end{minipage}
    }
    \hspace{-4mm}
    \subfigure{
        \begin{minipage}[t]{0.1\linewidth}
            \centering
            \includegraphics[width=1\linewidth]{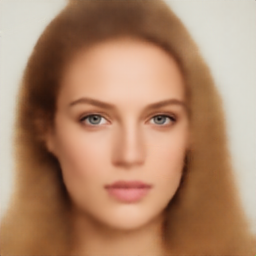}
        \end{minipage}
    }
    \hspace{-4mm}
    \subfigure{
        \begin{minipage}[t]{0.1\linewidth}
            \centering
            \includegraphics[width=1\linewidth]{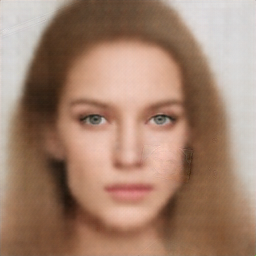}
        \end{minipage}
    }
    \hspace{-4mm}
    \subfigure{
        \begin{minipage}[t]{0.1\linewidth}
            \centering
            \includegraphics[width=1\linewidth]{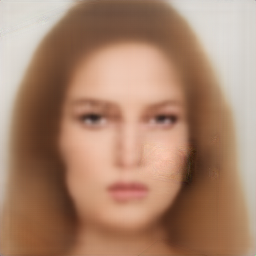}
        \end{minipage}
    }
    \hspace{-4mm}
    \subfigure{
        \begin{minipage}[t]{0.1\linewidth}
            \centering
            \includegraphics[width=1\linewidth]{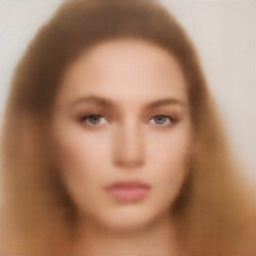}
        \end{minipage}
    }
    \hspace{-4mm}
    \subfigure{
        \begin{minipage}[t]{0.1\linewidth}
            \centering
            \includegraphics[width=1\linewidth]{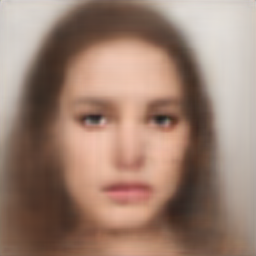}
        \end{minipage}
    }
    \hspace{-4mm}
    \subfigure{
        \begin{minipage}[t]{0.1\linewidth}
            \centering
            \includegraphics[width=1\linewidth]{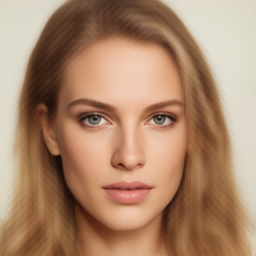}
        \end{minipage}
    }
    \hspace{-4mm}
    \subfigure{
        \begin{minipage}[t]{0.1\linewidth}
            \centering
            \includegraphics[width=1\linewidth]{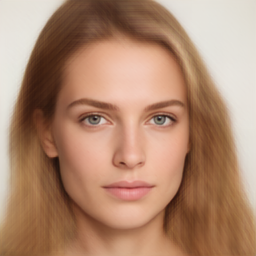}
        \end{minipage}
    }
    \hspace{-4mm}
    \subfigure{
        \begin{minipage}[t]{0.1\linewidth}
            \centering
            \includegraphics[width=1\linewidth]{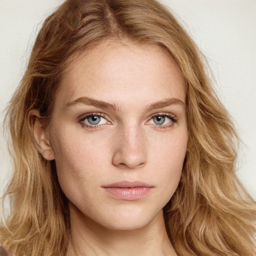}
        \end{minipage}
    }
    
    \vspace{-2.8mm}
    \setcounter{subfigure}{0}		
    
    \subfigure{
        \rotatebox{90}{\footnotesize{\ \ 70;1;L;Wb}}
        \begin{minipage}[t]{0.1\linewidth}
            \centering
            \includegraphics[width=1\linewidth]{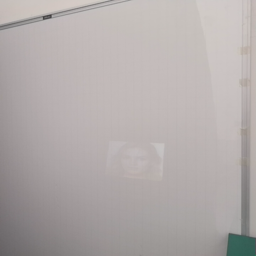}
            \centerline{\footnotesize{INPUT}}
        \end{minipage}
    }
    \hspace{-4mm}
    \subfigure{
        \begin{minipage}[t]{0.1\linewidth}
            \centering
            \includegraphics[width=1\linewidth]{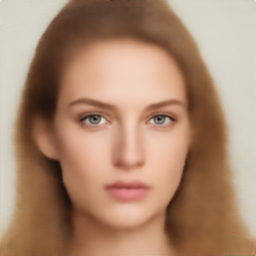}
            \centerline{\footnotesize{KBNet}}
        \end{minipage}
    }
    \hspace{-4mm}
    \subfigure{
        \begin{minipage}[t]{0.1\linewidth}
            \centering
            \includegraphics[width=1\linewidth]{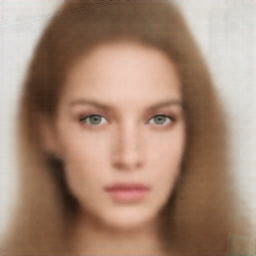}
            \centerline{\footnotesize{NAFNet}}
        \end{minipage}
    }
    \hspace{-4mm}
    \subfigure{
        \begin{minipage}[t]{0.1\linewidth}
            \centering
            \includegraphics[width=1\linewidth]{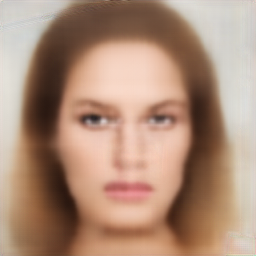}
            \centerline{\footnotesize{MSDINet}}
        \end{minipage}
    }
    \hspace{-4mm}
    \subfigure{
        \begin{minipage}[t]{0.1\linewidth}
            \centering
            \includegraphics[width=1\linewidth]{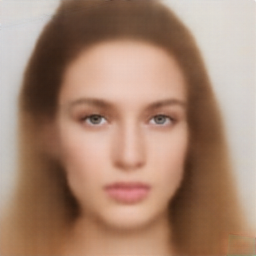}
            \centerline{\footnotesize{Uformer}}
        \end{minipage}
    }
    \hspace{-4mm}
    \subfigure{
        \begin{minipage}[t]{0.1\linewidth}
            \centering
            \includegraphics[width=1\linewidth]{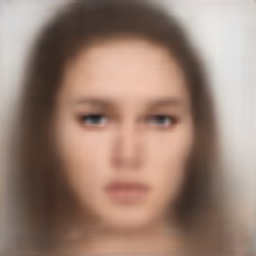}
            \centerline{\footnotesize{SwinIR}}
        \end{minipage}
    }
    \hspace{-4mm}
    \subfigure{
        \begin{minipage}[t]{0.1\linewidth}
            \centering
            \includegraphics[width=1\linewidth]{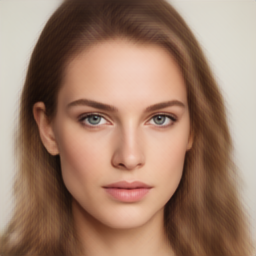}
            \centerline{\footnotesize{NLOS-OT}}
        \end{minipage}
    }
    \hspace{-4mm}
    \subfigure{
        \begin{minipage}[t]{0.1\linewidth}
            \centering
            \includegraphics[width=1\linewidth]{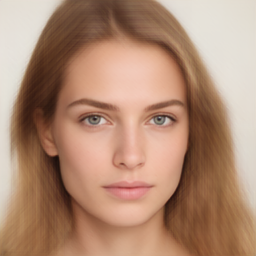}
            \centerline{\footnotesize{\textbf{NLOS-LTM}}}
        \end{minipage}
    }
    \hspace{-4mm}
    \subfigure{
        \begin{minipage}[t]{0.1\linewidth}
            \centering
            \includegraphics[width=1\linewidth]{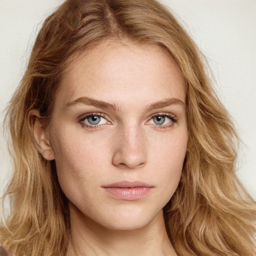}
            \centerline{\footnotesize{GT}}
        \end{minipage}
    }	
    \caption{Visual comparison of the results on the All-Supermodel dataset under different light transport conditions.}
    \label{Fig-All-Supermodel}
\end{figure*}

\textit{2) All-Supermodel:}
For All-Supermodel, NLOS-LTM surpasses most baseline methods on almost all light transport conditions considered. Fig. \ref{Fig-All-Supermodel} illustrates the qualitative comparisons of the results, in which NLOS-LTM not only successfully reconstructs the hidden images with greater clarity, but also preserves more consistent details of the faces, such as the shade on the face and the color of the hair.

\begin{figure*}[!t]
    \centering
    \subfigure{
        \rotatebox{90}{\footnotesize{\ \ 100;1;A;Wall}}
        \begin{minipage}[t]{0.1\linewidth} 
            \centering
            \includegraphics[width=1\linewidth]{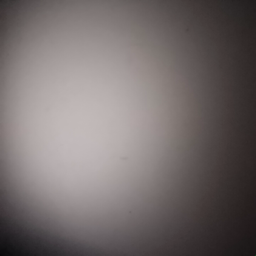}
        \end{minipage}
    }
    \hspace{-4mm}
    \subfigure{
        \begin{minipage}[t]{0.1\linewidth}
            \centering
            \includegraphics[width=1\linewidth]{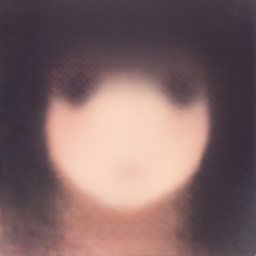}
        \end{minipage}
    }
    \hspace{-4mm}
    \subfigure{
        \begin{minipage}[t]{0.1\linewidth}
            \centering
            \includegraphics[width=1\linewidth]{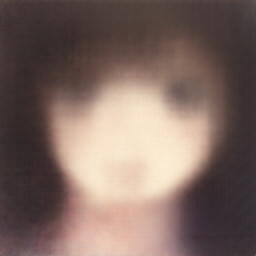}
        \end{minipage}
    }
    \hspace{-4mm}
    \subfigure{
        \begin{minipage}[t]{0.1\linewidth}
            \centering
            \includegraphics[width=1\linewidth]{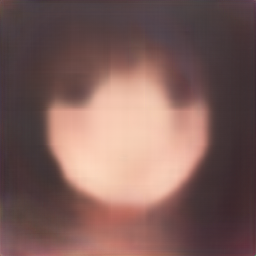}
        \end{minipage}
    }
    \hspace{-4mm}
    \subfigure{
        \begin{minipage}[t]{0.1\linewidth}
            \centering
            \includegraphics[width=1\linewidth]{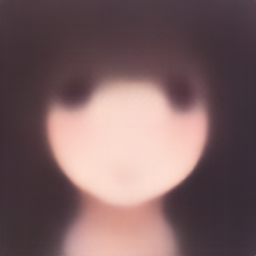}
        \end{minipage}
    }
    \hspace{-4mm}
    \subfigure{
        \begin{minipage}[t]{0.1\linewidth}
            \centering
            \includegraphics[width=1\linewidth]{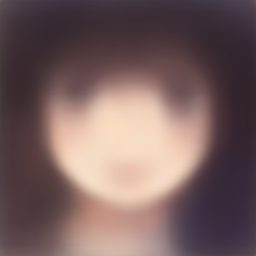}
        \end{minipage}
    }
    \hspace{-4mm}
    \subfigure{
        \begin{minipage}[t]{0.1\linewidth}
            \centering
            \includegraphics[width=1\linewidth]{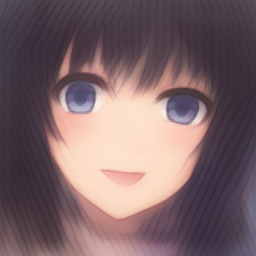}
        \end{minipage}
    }
    \hspace{-4mm}
    \subfigure{
        \begin{minipage}[t]{0.1\linewidth}
            \centering
            \includegraphics[width=1\linewidth]{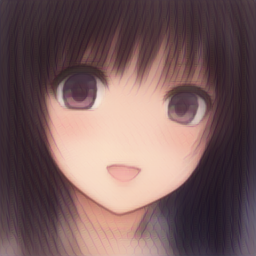}
        \end{minipage}
    }
    \hspace{-4mm}
    \subfigure{
        \begin{minipage}[t]{0.1\linewidth}
            \centering
            \includegraphics[width=1\linewidth]{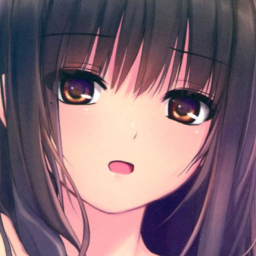}
        \end{minipage}
    }	
    
    \vspace{-2.8mm}
    \setcounter{subfigure}{0}
    
    \subfigure{
        \rotatebox{90}{\footnotesize{\ \ 70;2;A;Wall}}
        \begin{minipage}[t]{0.1\linewidth} 
            \centering
            \includegraphics[width=1\linewidth]{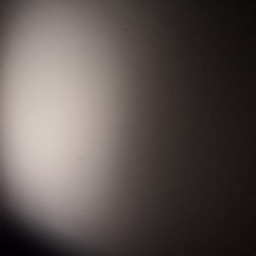}
        \end{minipage}
    }
    \hspace{-4mm}
    \subfigure{
        \begin{minipage}[t]{0.1\linewidth}
            \centering
            \includegraphics[width=1\linewidth]{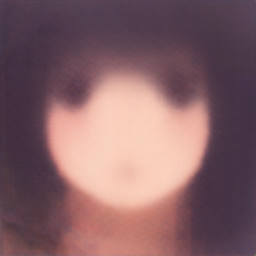}
        \end{minipage}
    }
    \hspace{-4mm}
    \subfigure{
        \begin{minipage}[t]{0.1\linewidth}
            \centering
            \includegraphics[width=1\linewidth]{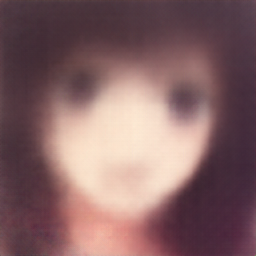}
        \end{minipage}
    }
    \hspace{-4mm}
    \subfigure{
        \begin{minipage}[t]{0.1\linewidth}
            \centering
            \includegraphics[width=1\linewidth]{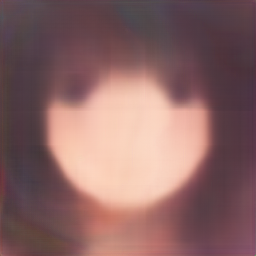}
        \end{minipage}
    }
    \hspace{-4mm}
    \subfigure{
        \begin{minipage}[t]{0.1\linewidth}
            \centering
            \includegraphics[width=1\linewidth]{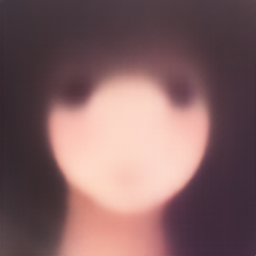}
        \end{minipage}
    }
    \hspace{-4mm}
    \subfigure{
        \begin{minipage}[t]{0.1\linewidth}
            \centering
            \includegraphics[width=1\linewidth]{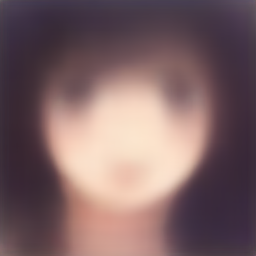}
        \end{minipage}
    }
    \hspace{-4mm}
    \subfigure{
        \begin{minipage}[t]{0.1\linewidth}
            \centering
            \includegraphics[width=1\linewidth]{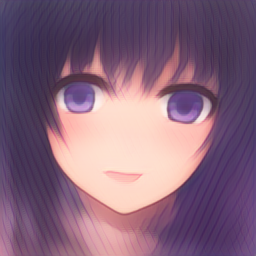}
        \end{minipage}
    }
    \hspace{-4mm}
    \subfigure{
        \begin{minipage}[t]{0.1\linewidth}
            \centering
            \includegraphics[width=1\linewidth]{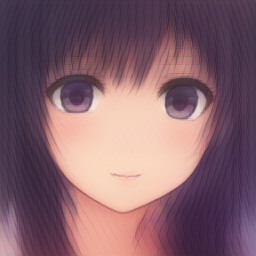}
        \end{minipage}
    }
    \hspace{-4mm}
    \subfigure{
        \begin{minipage}[t]{0.1\linewidth}
            \centering
            \includegraphics[width=1\linewidth]{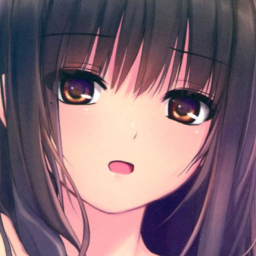}
        \end{minipage}
    }	
    
    \vspace{-2.8mm}
    \setcounter{subfigure}{0}
    
    \subfigure{
        \rotatebox{90}{\footnotesize{\ \ 100;2;A;Wall}}
        \begin{minipage}[t]{0.1\linewidth} 
            \centering
            \includegraphics[width=1\linewidth]{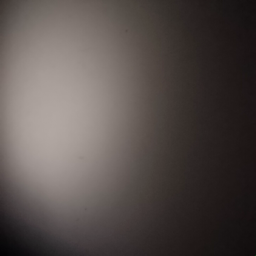}
        \end{minipage}
    }
    \hspace{-4mm}
    \subfigure{
        \begin{minipage}[t]{0.1\linewidth}
            \centering
            \includegraphics[width=1\linewidth]{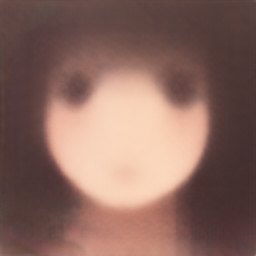}
        \end{minipage}
    }
    \hspace{-4mm}
    \subfigure{
        \begin{minipage}[t]{0.1\linewidth}
            \centering
            \includegraphics[width=1\linewidth]{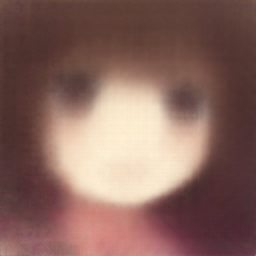}
        \end{minipage}
    }
    \hspace{-4mm}
    \subfigure{
        \begin{minipage}[t]{0.1\linewidth}
            \centering
            \includegraphics[width=1\linewidth]{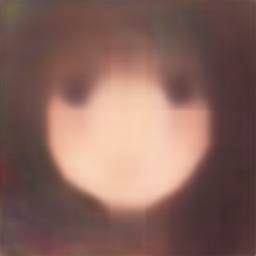}
        \end{minipage}
    }
    \hspace{-4mm}
    \subfigure{
        \begin{minipage}[t]{0.1\linewidth}
            \centering
            \includegraphics[width=1\linewidth]{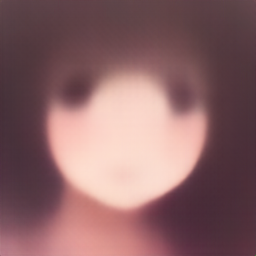}
        \end{minipage}
    }
    \hspace{-4mm}
    \subfigure{
        \begin{minipage}[t]{0.1\linewidth}
            \centering
            \includegraphics[width=1\linewidth]{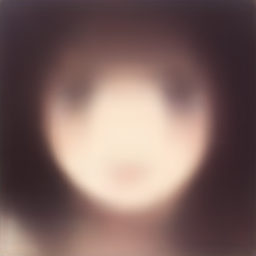}
        \end{minipage}
    }
    \hspace{-4mm}
    \subfigure{
        \begin{minipage}[t]{0.1\linewidth}
            \centering
            \includegraphics[width=1\linewidth]{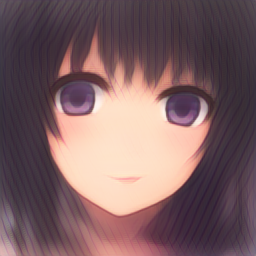}
        \end{minipage}
    }
    \hspace{-4mm}
    \subfigure{
        \begin{minipage}[t]{0.1\linewidth}
            \centering
            \includegraphics[width=1\linewidth]{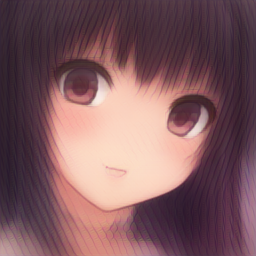}
        \end{minipage}
    }
    \hspace{-4mm}
    \subfigure{
        \begin{minipage}[t]{0.1\linewidth}
            \centering
            \includegraphics[width=1\linewidth]{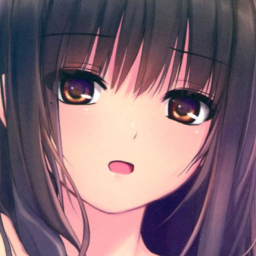}
        \end{minipage}
    }	
    
    \vspace{-3.5mm}
    \setcounter{subfigure}{0}
    
    \subfigure{
        \rotatebox{90}{\footnotesize{\ \ 70;1;L;Wall}}
        \begin{minipage}[t]{0.1\linewidth} 
            \centering
            \includegraphics[width=1\linewidth]{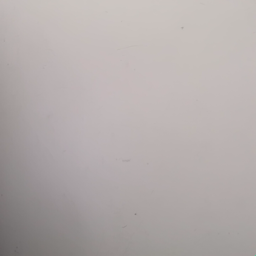}
        \end{minipage}
    }
    \hspace{-4mm}
    \subfigure{
        \begin{minipage}[t]{0.1\linewidth}
            \centering
            \includegraphics[width=1\linewidth]{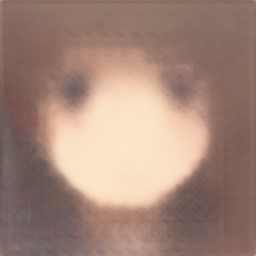}
        \end{minipage}
    }
    \hspace{-4mm}
    \subfigure{
        \begin{minipage}[t]{0.1\linewidth}
            \centering
            \includegraphics[width=1\linewidth]{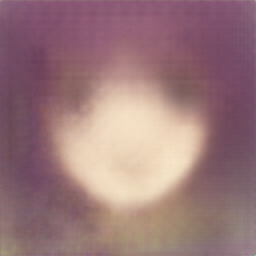}
        \end{minipage}
    }
    \hspace{-4mm}
    \subfigure{
        \begin{minipage}[t]{0.1\linewidth}
            \centering
            \includegraphics[width=1\linewidth]{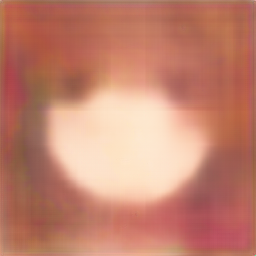}
        \end{minipage}
    }
    \hspace{-4mm}
    \subfigure{
        \begin{minipage}[t]{0.1\linewidth}
            \centering
            \includegraphics[width=1\linewidth]{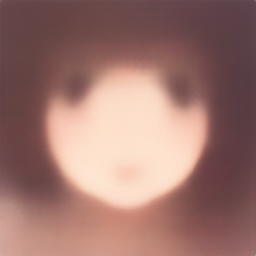}
        \end{minipage}
    }
    \hspace{-4mm}
    \subfigure{
        \begin{minipage}[t]{0.1\linewidth}
            \centering
            \includegraphics[width=1\linewidth]{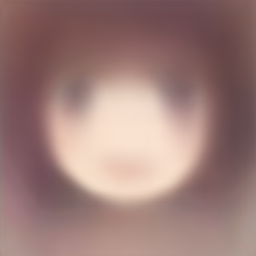}
        \end{minipage}
    }
    \hspace{-4mm}
    \subfigure{
        \begin{minipage}[t]{0.1\linewidth}
            \centering
            \includegraphics[width=1\linewidth]{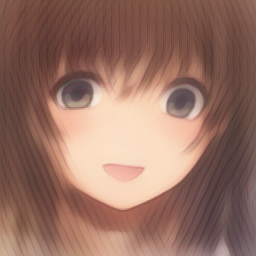}
        \end{minipage}
    }
    \hspace{-4mm}
    \subfigure{
        \begin{minipage}[t]{0.1\linewidth}
            \centering
            \includegraphics[width=1\linewidth]{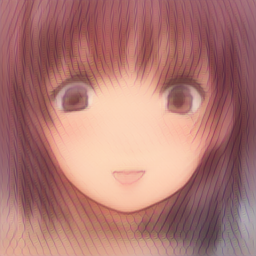}
        \end{minipage}
    }
    \hspace{-4mm}
    \subfigure{
        \begin{minipage}[t]{0.1\linewidth}
            \centering
            \includegraphics[width=1\linewidth]{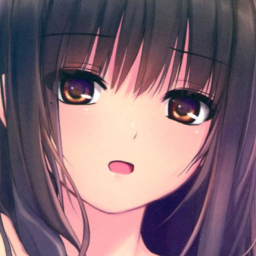}
        \end{minipage}
    }	
    
    \vspace{-2.8mm}
    \setcounter{subfigure}{0}
    
    \subfigure{
        \rotatebox{90}{\footnotesize{\ \ 70;1;A;Wb}}
        \begin{minipage}[t]{0.1\linewidth} 
            \centering
            \includegraphics[width=1\linewidth]{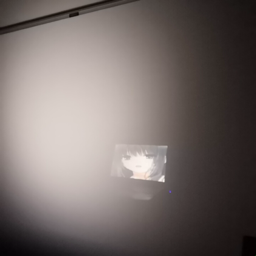}
        \end{minipage}
    }
    \hspace{-4mm}
    \subfigure{
        \begin{minipage}[t]{0.1\linewidth}
            \centering
            \includegraphics[width=1\linewidth]{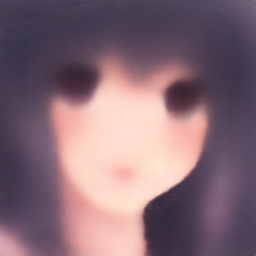}
        \end{minipage}
    }
    \hspace{-4mm}
    \subfigure{
        \begin{minipage}[t]{0.1\linewidth}
            \centering
            \includegraphics[width=1\linewidth]{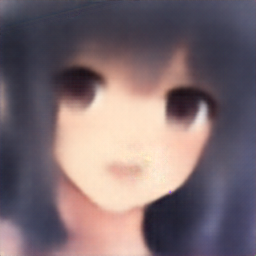}
        \end{minipage}
    }
    \hspace{-4mm}
    \subfigure{
        \begin{minipage}[t]{0.1\linewidth}
            \centering
            \includegraphics[width=1\linewidth]{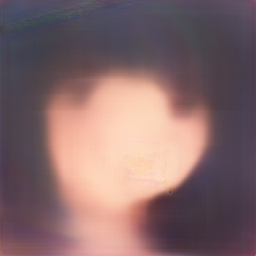}
        \end{minipage}
    }
    \hspace{-4mm}
    \subfigure{
        \begin{minipage}[t]{0.1\linewidth}
            \centering
            \includegraphics[width=1\linewidth]{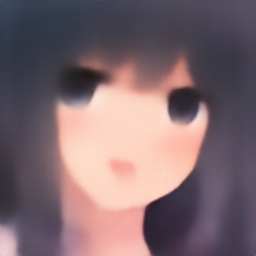}
        \end{minipage}
    }
    \hspace{-4mm}
    \subfigure{
        \begin{minipage}[t]{0.1\linewidth}
            \centering
            \includegraphics[width=1\linewidth]{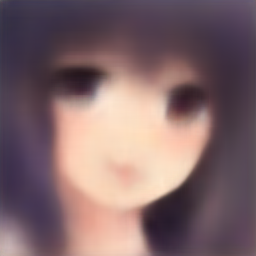}
        \end{minipage}
    }
    \hspace{-4mm}
    \subfigure{
        \begin{minipage}[t]{0.1\linewidth}
            \centering
            \includegraphics[width=1\linewidth]{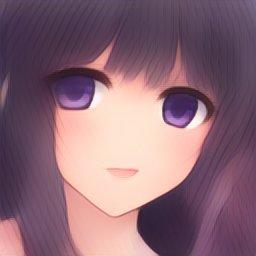}
        \end{minipage}
    }
    \hspace{-4mm}
    \subfigure{
        \begin{minipage}[t]{0.1\linewidth}
            \centering
            \includegraphics[width=1\linewidth]{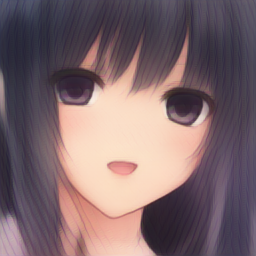}
        \end{minipage}
    }
    \hspace{-4mm}
    \subfigure{
        \begin{minipage}[t]{0.1\linewidth}
            \centering
            \includegraphics[width=1\linewidth]{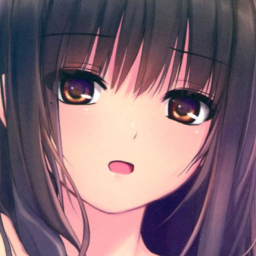}
        \end{minipage}
    }	
    
    \vspace{-2.8mm}
    \setcounter{subfigure}{0}
    
    \subfigure{
        \rotatebox{90}{\footnotesize{\ \ 70;2;A;Wb}}
        \begin{minipage}[t]{0.1\linewidth} 
            \centering
            \includegraphics[width=1\linewidth]{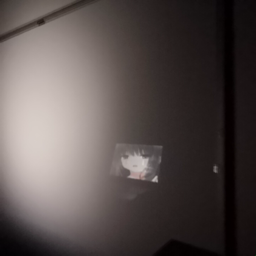}
        \end{minipage}
    }
    \hspace{-4mm}
    \subfigure{
        \begin{minipage}[t]{0.1\linewidth}
            \centering
            \includegraphics[width=1\linewidth]{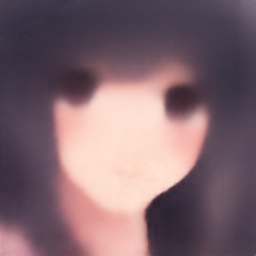}
        \end{minipage}
    }
    \hspace{-4mm}
    \subfigure{
        \begin{minipage}[t]{0.1\linewidth}
            \centering
            \includegraphics[width=1\linewidth]{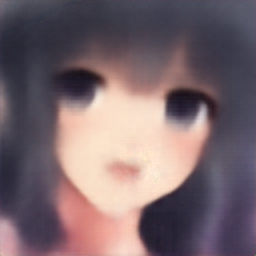}
        \end{minipage}
    }
    \hspace{-4mm}
    \subfigure{
        \begin{minipage}[t]{0.1\linewidth}
            \centering
            \includegraphics[width=1\linewidth]{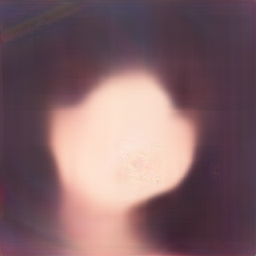}
        \end{minipage}
    }
    \hspace{-4mm}
    \subfigure{
        \begin{minipage}[t]{0.1\linewidth}
            \centering
            \includegraphics[width=1\linewidth]{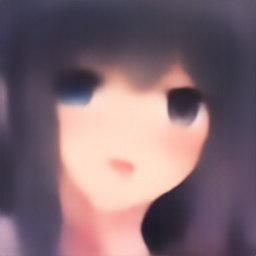}
        \end{minipage}
    }
    \hspace{-4mm}
    \subfigure{
        \begin{minipage}[t]{0.1\linewidth}
            \centering
            \includegraphics[width=1\linewidth]{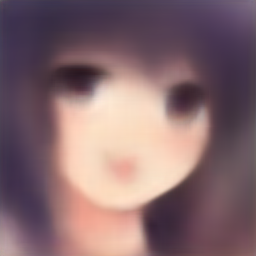}
        \end{minipage}
    }
    \hspace{-4mm}
    \subfigure{
        \begin{minipage}[t]{0.1\linewidth}
            \centering
            \includegraphics[width=1\linewidth]{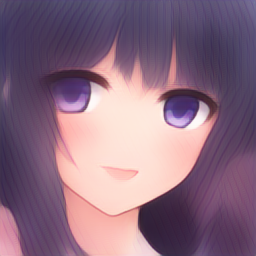}
        \end{minipage}
    }
    \hspace{-4mm}
    \subfigure{
        \begin{minipage}[t]{0.1\linewidth}
            \centering
            \includegraphics[width=1\linewidth]{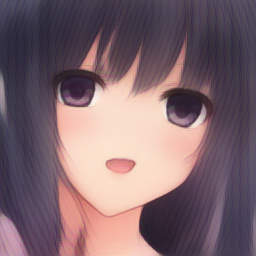}
        \end{minipage}
    }
    \hspace{-4mm}
    \subfigure{
        \begin{minipage}[t]{0.1\linewidth}
            \centering
            \includegraphics[width=1\linewidth]{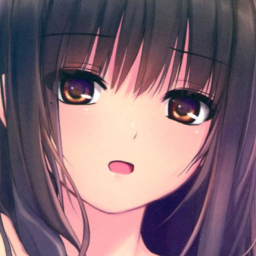}
        \end{minipage}
    }	
    
    \vspace{-2.8mm}
    \setcounter{subfigure}{0}	
    
    \subfigure{
        \rotatebox{90}{\footnotesize{\ \ 100;2;A;Wb}}
        \begin{minipage}[t]{0.1\linewidth} 
            \centering
            \includegraphics[width=1\linewidth]{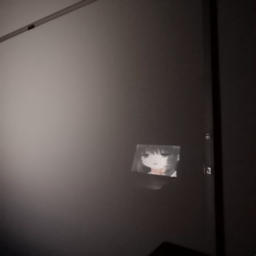}
        \end{minipage}
    }
    \hspace{-4mm}
    \subfigure{
        \begin{minipage}[t]{0.1\linewidth}
            \centering
            \includegraphics[width=1\linewidth]{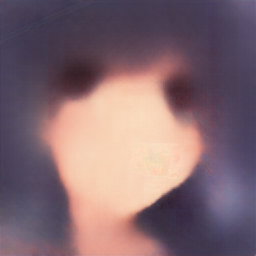}
        \end{minipage}
    }
    \hspace{-4mm}
    \subfigure{
        \begin{minipage}[t]{0.1\linewidth}
            \centering
            \includegraphics[width=1\linewidth]{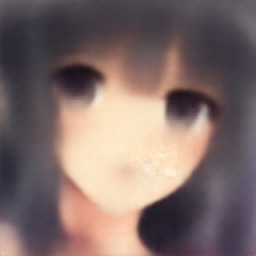}
        \end{minipage}
    }
    \hspace{-4mm}
    \subfigure{
        \begin{minipage}[t]{0.1\linewidth}
            \centering
            \includegraphics[width=1\linewidth]{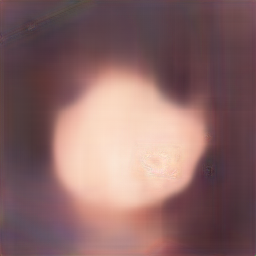}
        \end{minipage}
    }
    \hspace{-4mm}
    \subfigure{
        \begin{minipage}[t]{0.1\linewidth}
            \centering
            \includegraphics[width=1\linewidth]{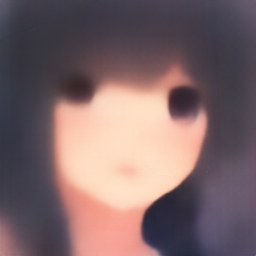}
        \end{minipage}
    }
    \hspace{-4mm}
    \subfigure{
        \begin{minipage}[t]{0.1\linewidth}
            \centering
            \includegraphics[width=1\linewidth]{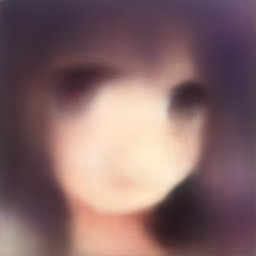}
        \end{minipage}
    }
    \hspace{-4mm}
    \subfigure{
        \begin{minipage}[t]{0.1\linewidth}
            \centering
            \includegraphics[width=1\linewidth]{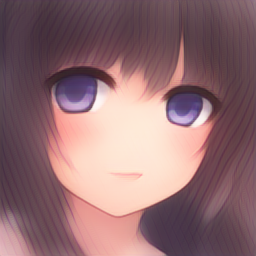}
        \end{minipage}
    }
    \hspace{-4mm}
    \subfigure{
        \begin{minipage}[t]{0.1\linewidth}
            \centering
            \includegraphics[width=1\linewidth]{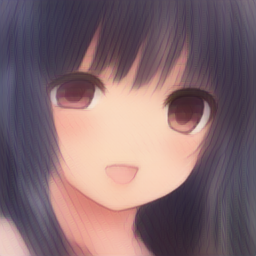}
        \end{minipage}
    }
    \hspace{-4mm}
    \subfigure{
        \begin{minipage}[t]{0.1\linewidth}
            \centering
            \includegraphics[width=1\linewidth]{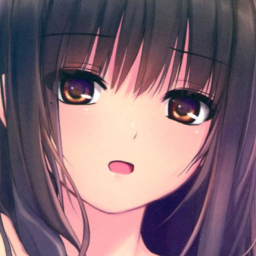}
        \end{minipage}
    }	
    
    \vspace{-2.8mm}
    \setcounter{subfigure}{0}	
    
    \subfigure{
        \rotatebox{90}{\footnotesize{\ \ 70;1;L;Wb}}
        \begin{minipage}[t]{0.1\linewidth}
            \centering
            \includegraphics[width=1\linewidth]{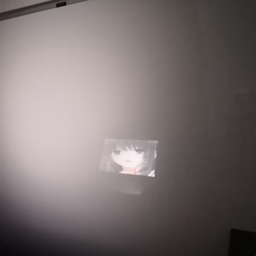}
            \centerline{\footnotesize{INPUT}}
        \end{minipage}
    }
    \hspace{-4mm}
    \subfigure{
        \begin{minipage}[t]{0.1\linewidth}
            \centering
            \includegraphics[width=1\linewidth]{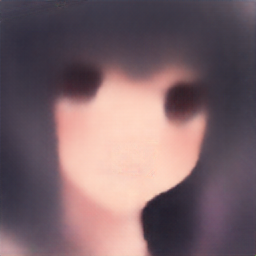}
            \centerline{\footnotesize{KBNet}}
        \end{minipage}
    }
    \hspace{-4mm}
    \subfigure{
        \begin{minipage}[t]{0.1\linewidth}
            \centering
            \includegraphics[width=1\linewidth]{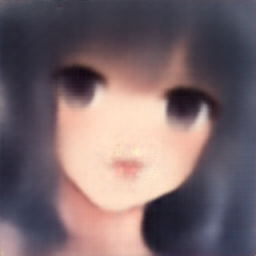}
            \centerline{\footnotesize{NAFNet}}
        \end{minipage}
    }
    \hspace{-4mm}
    \subfigure{
        \begin{minipage}[t]{0.1\linewidth}
            \centering
            \includegraphics[width=1\linewidth]{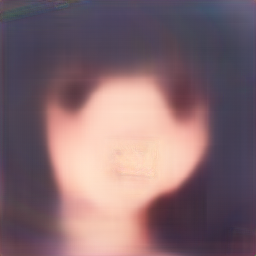}
            \centerline{\footnotesize{MSDINet}}
        \end{minipage}
    }
    \hspace{-4mm}
    \subfigure{
        \begin{minipage}[t]{0.1\linewidth}
            \centering
            \includegraphics[width=1\linewidth]{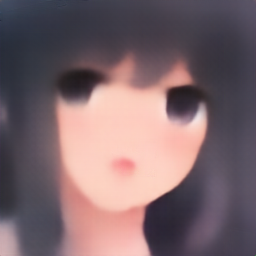}
            \centerline{\footnotesize{Uformer}}
        \end{minipage}
    }
    \hspace{-4mm}
    \subfigure{
        \begin{minipage}[t]{0.1\linewidth}
            \centering
            \includegraphics[width=1\linewidth]{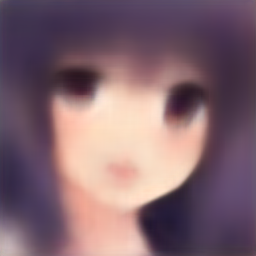}
            \centerline{\footnotesize{SwinIR}}
        \end{minipage}
    }
    \hspace{-4mm}
    \subfigure{
        \begin{minipage}[t]{0.1\linewidth}
            \centering
            \includegraphics[width=1\linewidth]{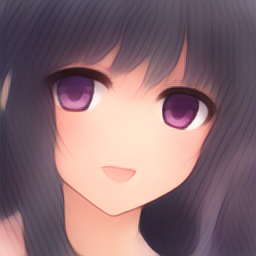}
            \centerline{\footnotesize{NLOS-OT}}
        \end{minipage}
    }
    \hspace{-4mm}
    \subfigure{
        \begin{minipage}[t]{0.1\linewidth}
            \centering
            \includegraphics[width=1\linewidth]{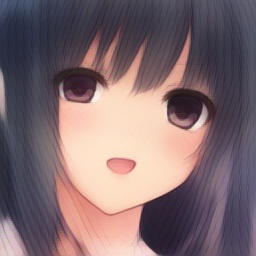}
            \centerline{\footnotesize{\textbf{NLOS-LTM}}}
        \end{minipage}
    }
    \hspace{-4mm}
    \subfigure{
        \begin{minipage}[t]{0.1\linewidth}
            \centering
            \includegraphics[width=1\linewidth]{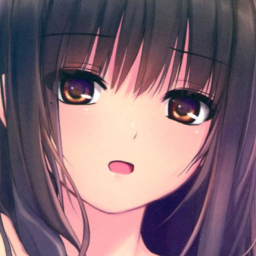}
            \centerline{\footnotesize{{GT}}}
        \end{minipage}
    }
    \caption{Visual comparison of the results on the All-Anime dataset under different light transport conditions.}
    \label{Fig-All-Anime}
\end{figure*}

\textit{3) All-Anime:}
For the All-Anime dataset, NLOS-LTM achieves the highest SSIM scores on all light transport conditions. For PSNR, out of the eight conditions, NLOS-LTM gets the highest for three conditions and the second highest for another three conditions. Compared with PSNR, which measures the mean squared error of pixel values, SSIM is more aligned with human perception. The SSIM scores indicate that the visual quality of the images reconstructed by NLOS-LTM is the best. This is confirmed by the qualitative comparisons in Fig. \ref{Fig-All-Anime}. Although some restoration methods get higher PSNR than NLOS-LTM, their reconstructions are quite blurry and lacking details. Although the clarity of the NLOS-LTM reconstructions is similar with the NLOS-OT reconstructions, for some important attributes of the faces, such as the pose and the color of the eyes, the former is more consistent with the ground truth than the latter.

To illustrate the effectiveness of NLOS-LTM for handling multiple light transport conditions in an all-in-one framework, we also compare its performance with NLOS-OT trained with its original setting that learns a condition specific model for each light transport condition (The results are denoted as NLOS-OT-S in Table \ref{Table-All}). As we can see from the results, NLOS-OT-S achieves better reconstructions than NLOS-OT that trains a single model for multiple conditions. On the other hand, NLOS-LTM outperforms NLOS-OT-S on all three datasets with notable margins. This is remarkable that we only train a single reconstruction model with NLOS-LTM and we do not need to know under which condition a projection image is taken during testing.

\subsection{Assessment of Generalization Ability}
\begin{figure*}[!t]
    \centering
    \subfigcapskip=5pt 
    \subfigure{
            \rotatebox{90}{\makebox[0pt][c]{\vspace{1.5cm}\scriptsize{70;1;A;Wall}}}
        \begin{minipage}[t]{0.054\linewidth} 
            \centering
            \includegraphics[width=1\linewidth]{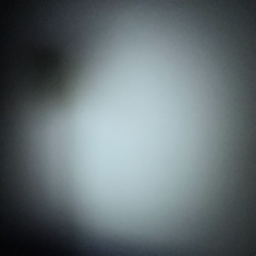}
            \includegraphics[width=1\linewidth]{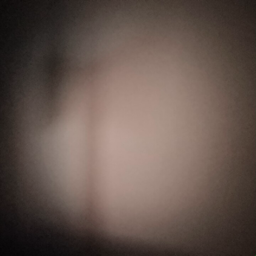}
        \end{minipage}
        \hspace{-2.8mm}
        \begin{minipage}[t]{0.054\linewidth}
            \centering
            \includegraphics[width=1\linewidth]{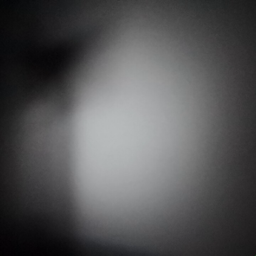}
            \includegraphics[width=1\linewidth]{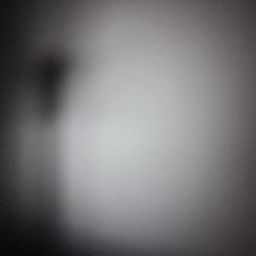}
        \end{minipage}
    }
    \hspace{-4.5mm}
    \subfigure{
        \begin{minipage}[t]{0.054\linewidth}
            \centering
            \includegraphics[width=1\linewidth]{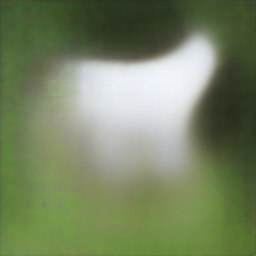}
            \includegraphics[width=1\linewidth]{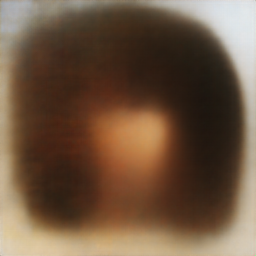}
        \end{minipage}
        \hspace{-2.8mm}      
        \begin{minipage}[t]{0.054\linewidth}
            \centering
            \includegraphics[width=1\linewidth]{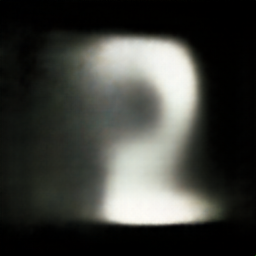}
            \includegraphics[width=1\linewidth]{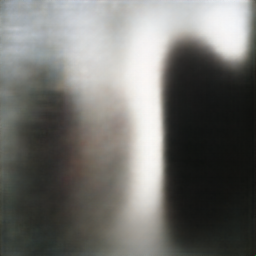}
        \end{minipage}
    }
    \hspace{-4.5mm}
    \subfigure{
        \begin{minipage}[t]{0.054\linewidth}
            \centering
            \includegraphics[width=1\linewidth]{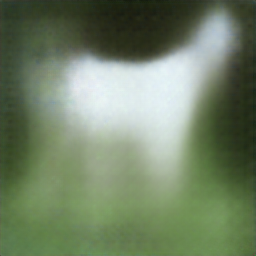}
            \includegraphics[width=1\linewidth]{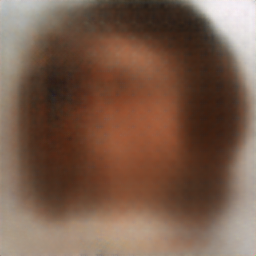}  
        \end{minipage}
        \hspace{-2.8mm}      
        \begin{minipage}[t]{0.054\linewidth} 
            \centering
            \includegraphics[width=1\linewidth]{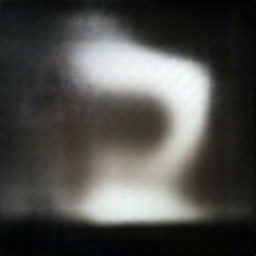}
            \includegraphics[width=1\linewidth]{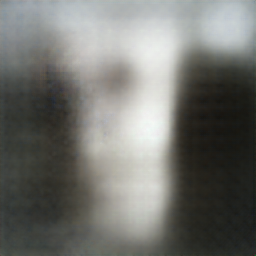}
        \end{minipage}
    }
    \hspace{-4.5mm}
    \subfigure{
        \begin{minipage}[t]{0.054\linewidth}
            \centering
            \includegraphics[width=1\linewidth]{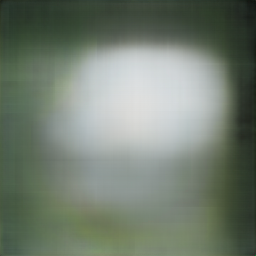}
            \includegraphics[width=1\linewidth]{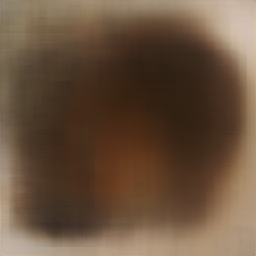}
        \end{minipage}
        \hspace{-2.8mm}      
        \begin{minipage}[t]{0.054\linewidth}
            \centering
            \includegraphics[width=1\linewidth]{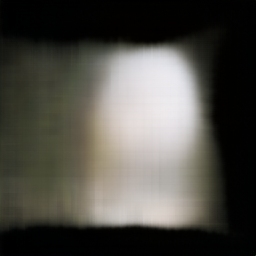}
           \includegraphics[width=1\linewidth]{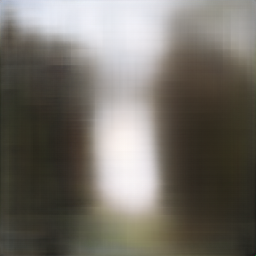}
        \end{minipage}
    }
    \hspace{-4.5mm}
    \subfigure{
        \begin{minipage}[t]{0.054\linewidth}
            \centering
            \includegraphics[width=1\linewidth]{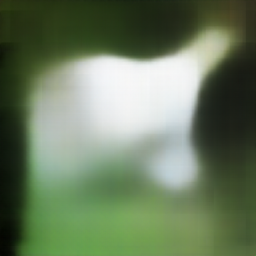}
            \includegraphics[width=1\linewidth]{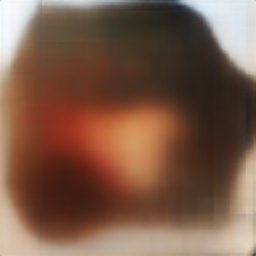}
        \end{minipage}
        \hspace{-2.8mm}      
        \begin{minipage}[t]{0.054\linewidth}
            \centering
            \includegraphics[width=1\linewidth]{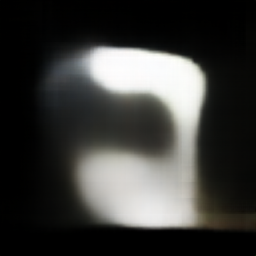}
            \includegraphics[width=1\linewidth]{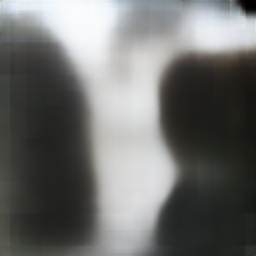}
        \end{minipage}
    }
    \hspace{-4.5mm}
    \subfigure{
        \begin{minipage}[t]{0.054\linewidth}
            \centering
            \includegraphics[width=1\linewidth]{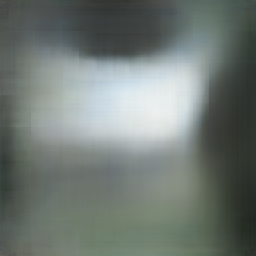}
            \includegraphics[width=1\linewidth]{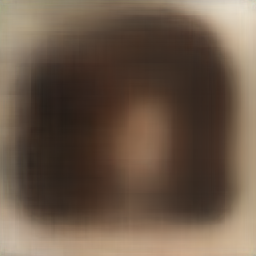}
        \end{minipage}
        \hspace{-2.8mm}      
        \begin{minipage}[t]{0.054\linewidth}
            \centering
            \includegraphics[width=1\linewidth]{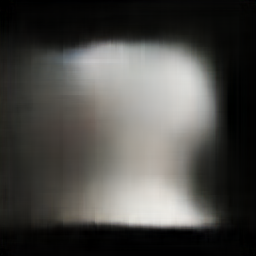}
            \includegraphics[width=1\linewidth]{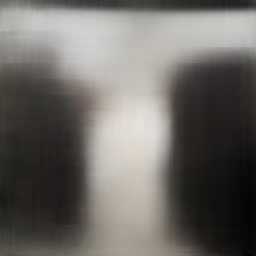}
        \end{minipage}
    }
    \hspace{-4.5mm}
    \subfigure{
        \begin{minipage}[t]{0.054\linewidth}
            \centering
            \includegraphics[width=1\linewidth]{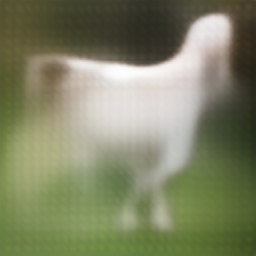}
            \includegraphics[width=1\linewidth]{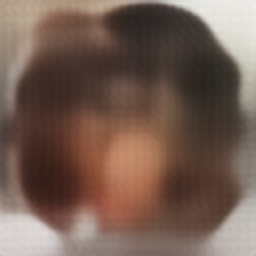}
        \end{minipage}
        \hspace{-2.8mm}      
        \begin{minipage}[t]{0.054\linewidth}
            \centering
            \includegraphics[width=1\linewidth]{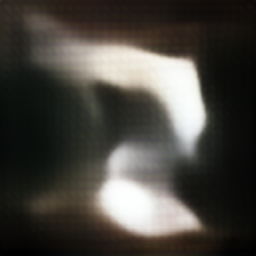}
            \includegraphics[width=1\linewidth]{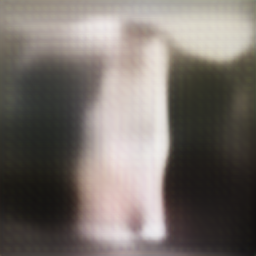}
        \end{minipage}
    }
    \hspace{-4.5mm}
    \subfigure{
        \begin{minipage}[t]{0.054\linewidth}
            \centering
            \includegraphics[width=1\linewidth]{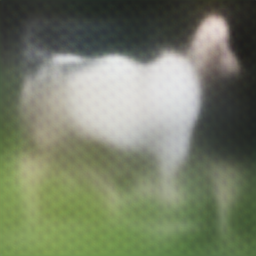}
            \includegraphics[width=1\linewidth]{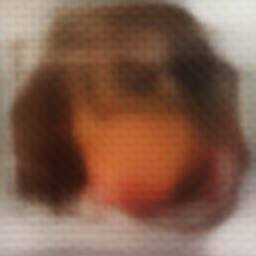}
        \end{minipage}
        \hspace{-2.8mm}      
        \begin{minipage}[t]{0.054\linewidth}
            \centering
            \includegraphics[width=1\linewidth]{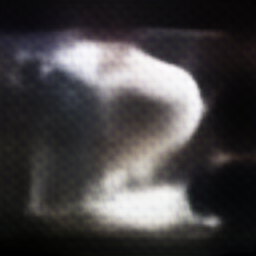}
            \includegraphics[width=1\linewidth]{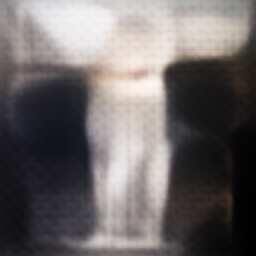}
        \end{minipage}
    }
    \hspace{-4.5mm}
    \subfigure{
        \begin{minipage}[t]{0.054\linewidth}
            \centering
            \includegraphics[width=1\linewidth]{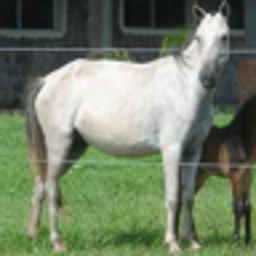}
            \includegraphics[width=1\linewidth]{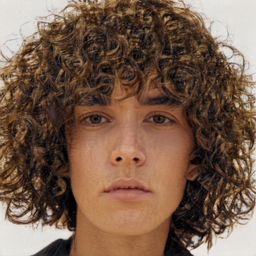}
        \end{minipage}
        \hspace{-2.8mm}      
        \begin{minipage}[t]{0.054\linewidth}
            \centering
            \includegraphics[width=1\linewidth]{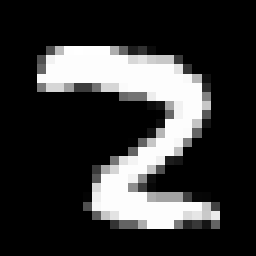}
            \includegraphics[width=1\linewidth]{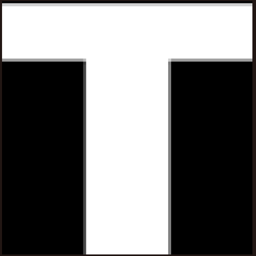}
        \end{minipage}
    }
    
    \vspace{-3mm}
    \setcounter{subfigure}{0}

    \subfigure{
       \rotatebox{90}{\makebox[0pt][c]{\vspace{1.5cm}\scriptsize{100;1;A;Wall}}}
        \begin{minipage}[t]{0.054\linewidth} 
            \centering
            \includegraphics[width=1\linewidth]{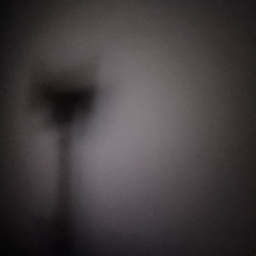}
            \includegraphics[width=1\linewidth]{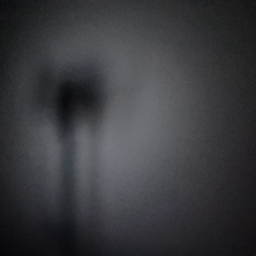}
        \end{minipage}
        \hspace{-2.8mm}
        \begin{minipage}[t]{0.054\linewidth}
            \centering
            \includegraphics[width=1\linewidth]{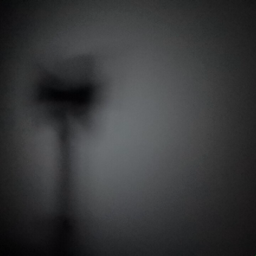}
            \includegraphics[width=1\linewidth]{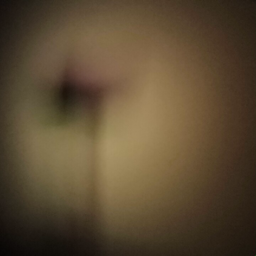}
        \end{minipage}
    }
    \hspace{-4.5mm}
    \subfigure{
        \begin{minipage}[t]{0.054\linewidth}
            \centering
            \includegraphics[width=1\linewidth]{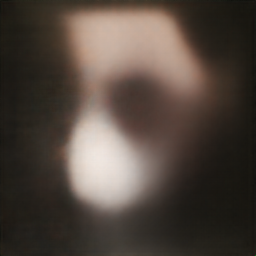}
            \includegraphics[width=1\linewidth]{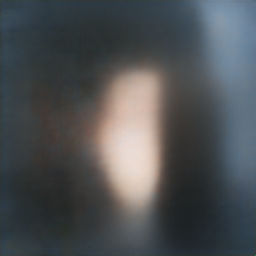}
        \end{minipage}
        \hspace{-2.8mm}      
        \begin{minipage}[t]{0.054\linewidth}
            \centering
            \includegraphics[width=1\linewidth]{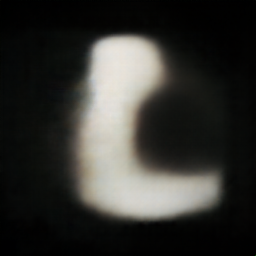}
            \includegraphics[width=1\linewidth]{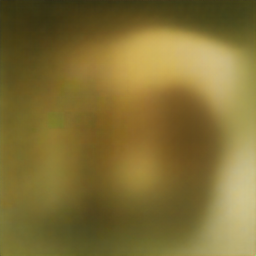}
        \end{minipage}
    }
    \hspace{-4.5mm}
    \subfigure{
        \begin{minipage}[t]{0.054\linewidth}
            \centering
            \includegraphics[width=1\linewidth]{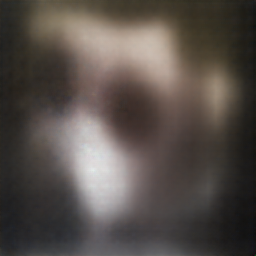}
            \includegraphics[width=1\linewidth]{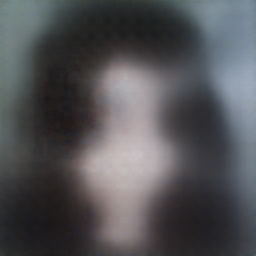}
        \end{minipage}
        \hspace{-2.8mm}      
        \begin{minipage}[t]{0.054\linewidth}
            \centering
            \includegraphics[width=1\linewidth]{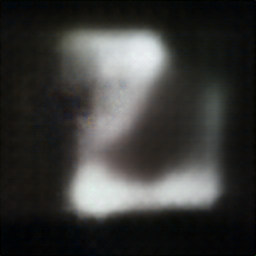}
            \includegraphics[width=1\linewidth]{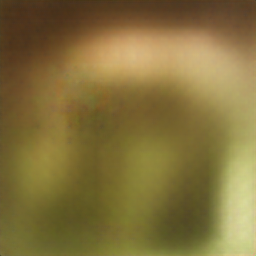}
        \end{minipage}
    }
    \hspace{-4.5mm}
    \subfigure{
        \begin{minipage}[t]{0.054\linewidth}
            \centering
            \includegraphics[width=1\linewidth]{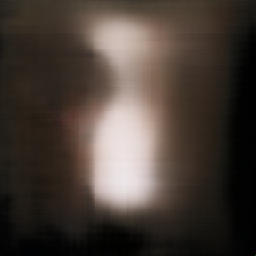}
            \includegraphics[width=1\linewidth]{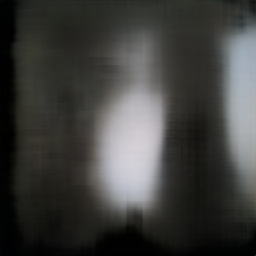}
        \end{minipage}
        \hspace{-2.8mm}      
        \begin{minipage}[t]{0.054\linewidth}
            \centering
            \includegraphics[width=1\linewidth]{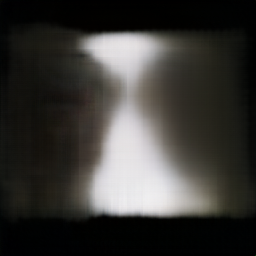}
            \includegraphics[width=1\linewidth]{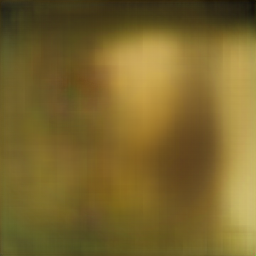}
        \end{minipage}
    }
    \hspace{-4.5mm}
    \subfigure{
        \begin{minipage}[t]{0.054\linewidth}
            \centering
            \includegraphics[width=1\linewidth]{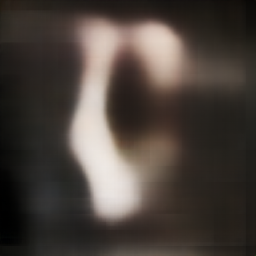}
            \includegraphics[width=1\linewidth]{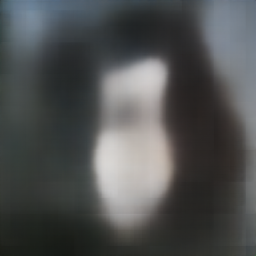}
        \end{minipage}
        \hspace{-2.8mm}      
        \begin{minipage}[t]{0.054\linewidth}
            \centering
            \includegraphics[width=1\linewidth]{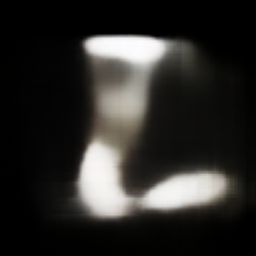}
            \includegraphics[width=1\linewidth]{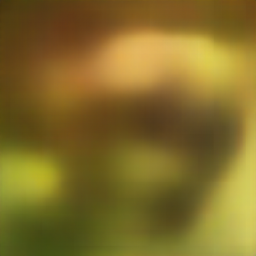}
        \end{minipage}
    }
    \hspace{-4.5mm}
    \subfigure{
        \begin{minipage}[t]{0.054\linewidth}
            \centering
            \includegraphics[width=1\linewidth]{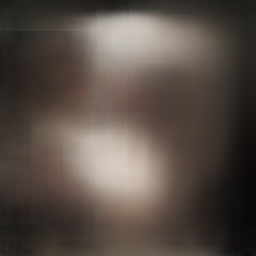}
            \includegraphics[width=1\linewidth]{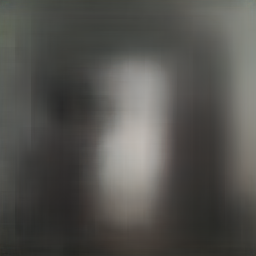}
        \end{minipage}
        \hspace{-2.8mm}      
        \begin{minipage}[t]{0.054\linewidth}
            \centering
            \includegraphics[width=1\linewidth]{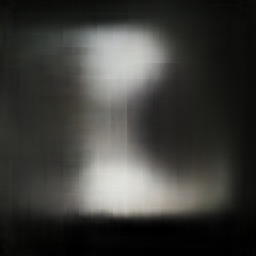}
            \includegraphics[width=1\linewidth]{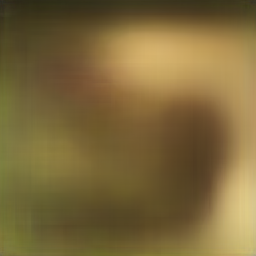}
        \end{minipage}
    }
    \hspace{-4.5mm}
    \subfigure{
        \begin{minipage}[t]{0.054\linewidth}
            \centering
            \includegraphics[width=1\linewidth]{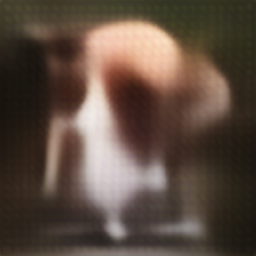}
            \includegraphics[width=1\linewidth]{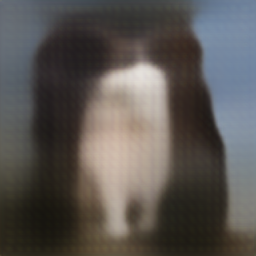}
        \end{minipage}
        \hspace{-2.8mm}      
        \begin{minipage}[t]{0.054\linewidth}
            \centering
            \includegraphics[width=1\linewidth]{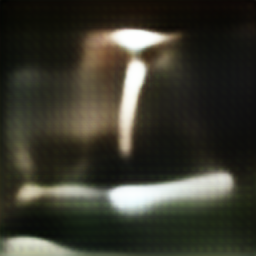}
            \includegraphics[width=1\linewidth]{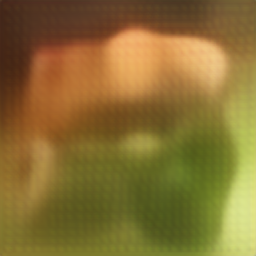}
        \end{minipage}
    }
    \hspace{-4.5mm}
    \subfigure{
        \begin{minipage}[t]{0.054\linewidth}
            \centering
            \includegraphics[width=1\linewidth]{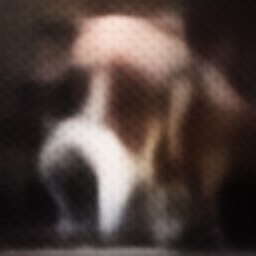}
            \includegraphics[width=1\linewidth]{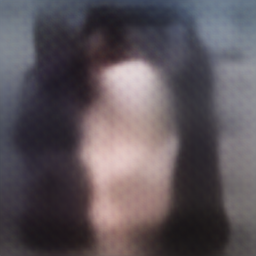}  
        \end{minipage}
        \hspace{-2.8mm}      
        \begin{minipage}[t]{0.054\linewidth} 
            \centering
            \includegraphics[width=1\linewidth]{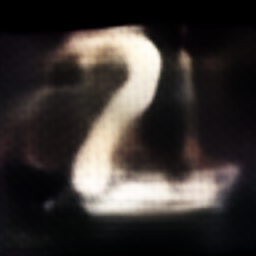}
            \includegraphics[width=1\linewidth]{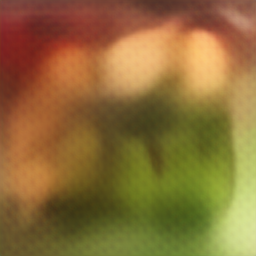}
        \end{minipage}
    }
    \hspace{-4.5mm}
    \subfigure{
        \begin{minipage}[t]{0.054\linewidth}
            \centering
            \includegraphics[width=1\linewidth]{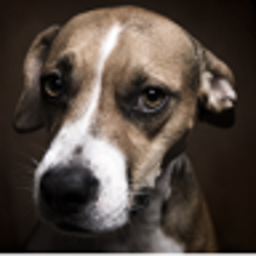}
            \includegraphics[width=1\linewidth]{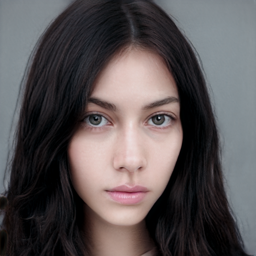}
        \end{minipage}
        \hspace{-2.8mm}      
        \begin{minipage}[t]{0.054\linewidth}
            \centering
            \includegraphics[width=1\linewidth]{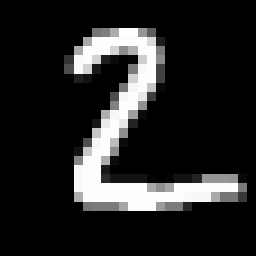}
            \includegraphics[width=1\linewidth]{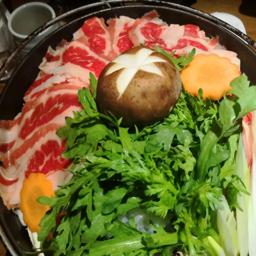}
        \end{minipage}
    }
    
    \vspace{-3mm}
    \setcounter{subfigure}{0}

    \subfigure{
      \rotatebox{90}{\makebox[0pt][c]{\vspace{1.5cm}\scriptsize{70;2;A;Wall}}}
        \begin{minipage}[t]{0.054\linewidth} 
            \centering
            \includegraphics[width=1\linewidth]{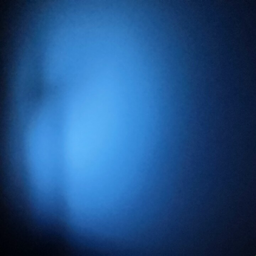}
            \includegraphics[width=1\linewidth]{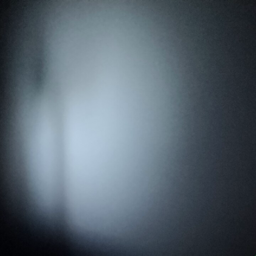}
        \end{minipage}
        \hspace{-2.8mm}
        \begin{minipage}[t]{0.054\linewidth}
            \centering
            \includegraphics[width=1\linewidth]{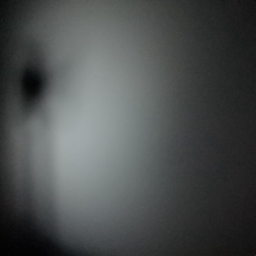}
            \includegraphics[width=1\linewidth]{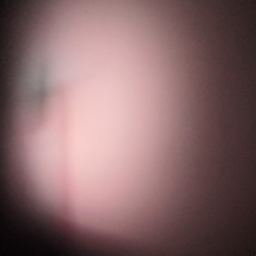}
        \end{minipage}
    }
    \hspace{-4.5mm}
    \subfigure{
        \begin{minipage}[t]{0.054\linewidth}
            \centering
            \includegraphics[width=1\linewidth]{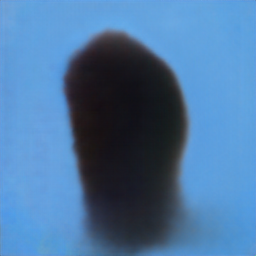}
            \includegraphics[width=1\linewidth]{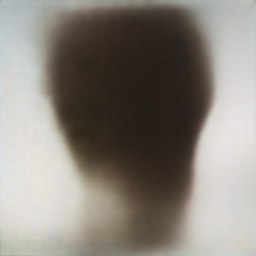}
        \end{minipage}
        \hspace{-2.8mm}      
        \begin{minipage}[t]{0.054\linewidth}
            \centering
            \includegraphics[width=1\linewidth]{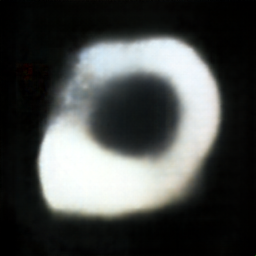}
            \includegraphics[width=1\linewidth]{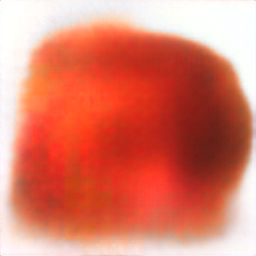}
        \end{minipage}
    }
    \hspace{-4.5mm}
    \subfigure{
        \begin{minipage}[t]{0.054\linewidth}
            \centering
            \includegraphics[width=1\linewidth]{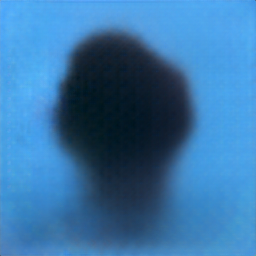}
            \includegraphics[width=1\linewidth]{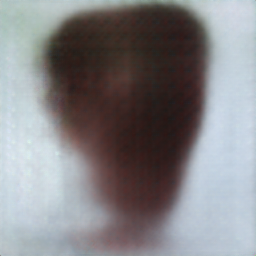}  
        \end{minipage}
        \hspace{-2.8mm}      
        \begin{minipage}[t]{0.054\linewidth} 
            \centering
            \includegraphics[width=1\linewidth]{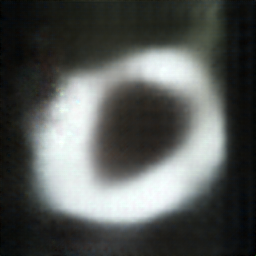}
            \includegraphics[width=1\linewidth]{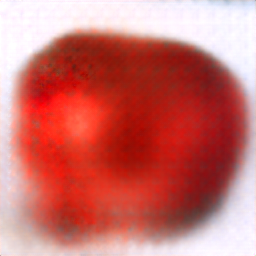}
        \end{minipage}
    }
    \hspace{-4.5mm}
    \subfigure{
        \begin{minipage}[t]{0.054\linewidth}
            \centering
            \includegraphics[width=1\linewidth]{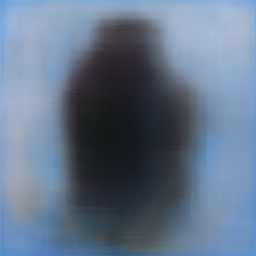}
            \includegraphics[width=1\linewidth]{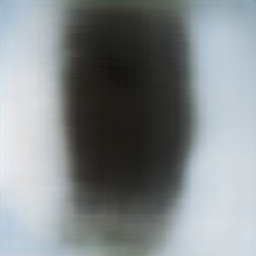}
        \end{minipage}
        \hspace{-2.8mm}      
        \begin{minipage}[t]{0.054\linewidth}
            \centering
            \includegraphics[width=1\linewidth]{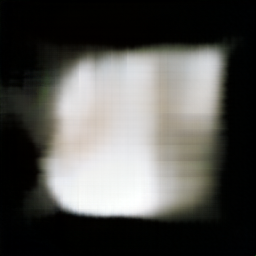}
            \includegraphics[width=1\linewidth]{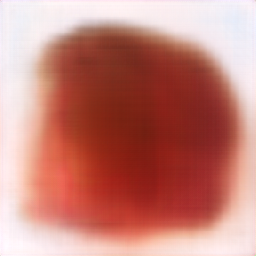}
        \end{minipage}
    }
    \hspace{-4.5mm}
    \subfigure{
        \begin{minipage}[t]{0.054\linewidth}
            \centering
            \includegraphics[width=1\linewidth]{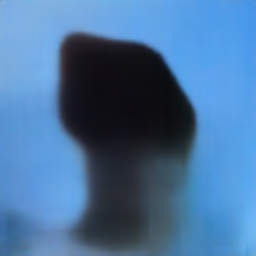}
            \includegraphics[width=1\linewidth]{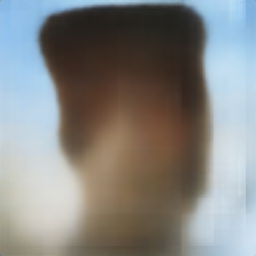}
        \end{minipage}
        \hspace{-2.8mm}      
        \begin{minipage}[t]{0.054\linewidth}
            \centering
            \includegraphics[width=1\linewidth]{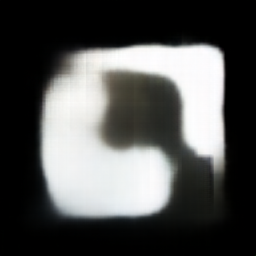}
            \includegraphics[width=1\linewidth]{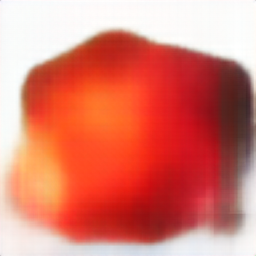}
        \end{minipage}
    }
    \hspace{-4.5mm}
    \subfigure{
        \begin{minipage}[t]{0.054\linewidth}
            \centering
            \includegraphics[width=1\linewidth]{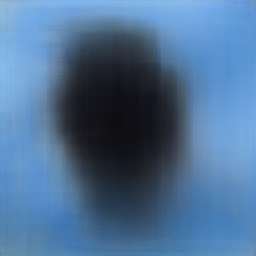}
            \includegraphics[width=1\linewidth]{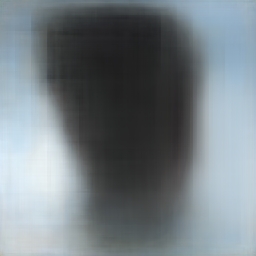}
        \end{minipage}
        \hspace{-2.8mm}      
        \begin{minipage}[t]{0.054\linewidth}
            \centering
            \includegraphics[width=1\linewidth]{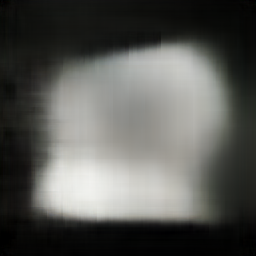}
            \includegraphics[width=1\linewidth]{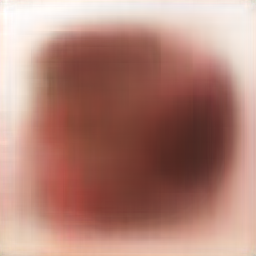}
        \end{minipage}
    }   
    \hspace{-4.5mm}
    \subfigure{
        \begin{minipage}[t]{0.054\linewidth}
            \centering
            \includegraphics[width=1\linewidth]{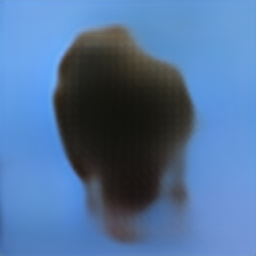}
            \includegraphics[width=1\linewidth]{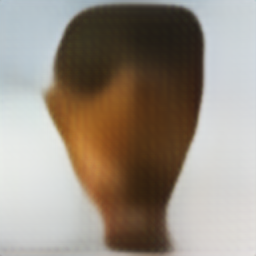}
        \end{minipage}
        \hspace{-2.8mm}      
        \begin{minipage}[t]{0.054\linewidth}
            \centering
            \includegraphics[width=1\linewidth]{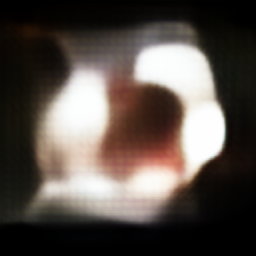}
            \includegraphics[width=1\linewidth]{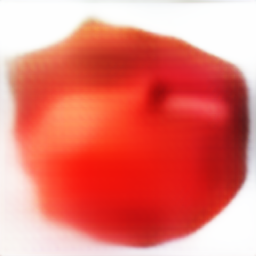}
        \end{minipage}
    }    
    \hspace{-4.5mm}
    \subfigure{
        \begin{minipage}[t]{0.054\linewidth}
            \centering
            \includegraphics[width=1\linewidth]{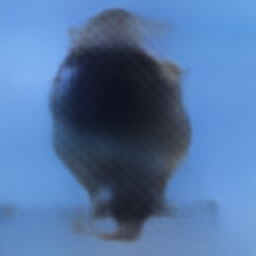}
            \includegraphics[width=1\linewidth]{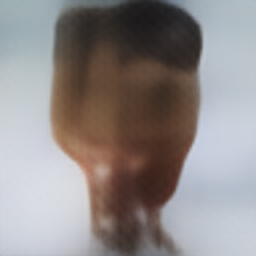}
        \end{minipage}
        \hspace{-2.8mm}      
        \begin{minipage}[t]{0.054\linewidth}
            \centering
            \includegraphics[width=1\linewidth]{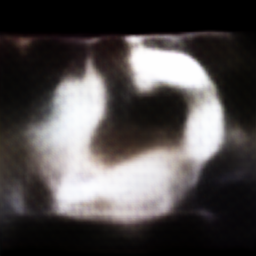}
            \includegraphics[width=1\linewidth]{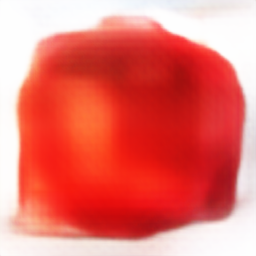}
        \end{minipage}
    }    
    \hspace{-4.5mm}
    \subfigure{
        \begin{minipage}[t]{0.054\linewidth}
            \centering
            \includegraphics[width=1\linewidth]{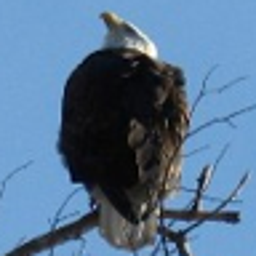}
            \includegraphics[width=1\linewidth]{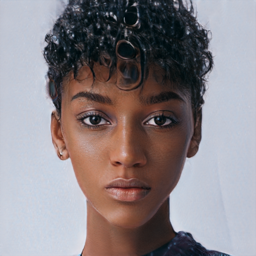}
        \end{minipage}
        \hspace{-2.8mm}      
        \begin{minipage}[t]{0.054\linewidth}
            \centering
            \includegraphics[width=1\linewidth]{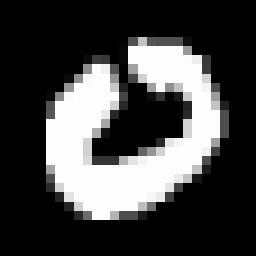}
            \includegraphics[width=1\linewidth]{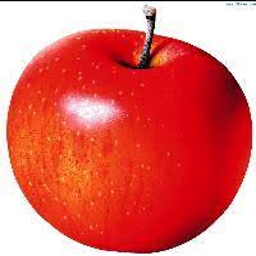}
        \end{minipage}
    }
    
    \vspace{-3mm}
    \setcounter{subfigure}{0}

    \subfigure{
         \rotatebox{90}{\makebox[0pt][c]{\vspace{1.5cm}\scriptsize{100;2;A;Wall}}}
        \begin{minipage}[t]{0.054\linewidth} 
            \centering
            \includegraphics[width=1\linewidth]{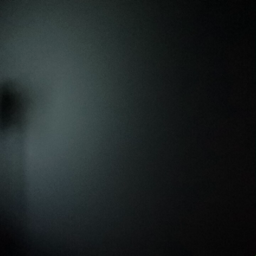}
            \includegraphics[width=1\linewidth]{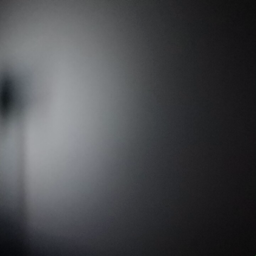}
        \end{minipage}
        \hspace{-2.8mm}
        \begin{minipage}[t]{0.054\linewidth}
            \centering
            \includegraphics[width=1\linewidth]{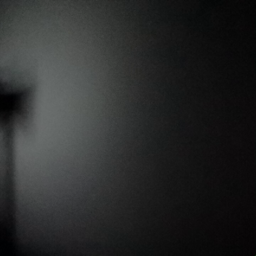}
            \includegraphics[width=1\linewidth]{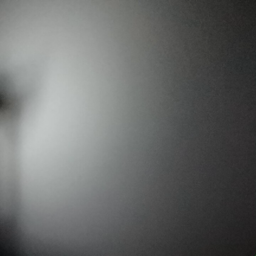}
        \end{minipage}
    }
    \hspace{-4.5mm}
    \subfigure{
        \begin{minipage}[t]{0.054\linewidth}
            \centering
            \includegraphics[width=1\linewidth]{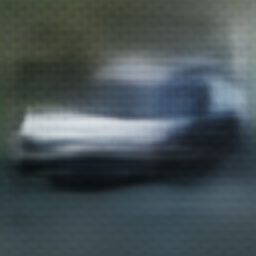}
            \includegraphics[width=1\linewidth]{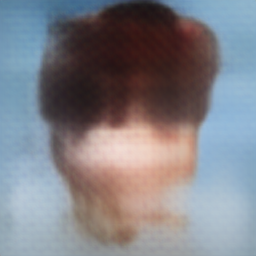}
        \end{minipage}
        \hspace{-2.8mm}      
        \begin{minipage}[t]{0.054\linewidth}
            \centering
            \includegraphics[width=1\linewidth]{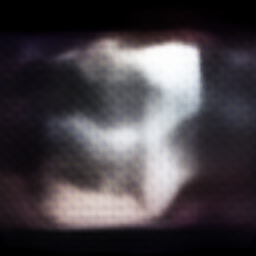}
            \includegraphics[width=1\linewidth]{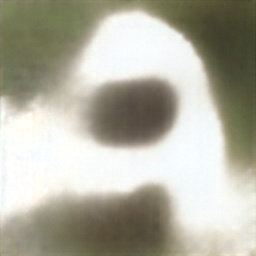}
        \end{minipage}
    }
    \hspace{-4.5mm}
    \subfigure{
        \begin{minipage}[t]{0.054\linewidth}
            \centering
            \includegraphics[width=1\linewidth]{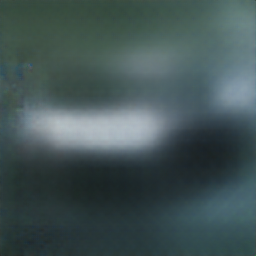}
            \includegraphics[width=1\linewidth]{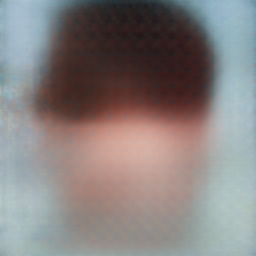}  
        \end{minipage}
        \hspace{-2.8mm}      
        \begin{minipage}[t]{0.054\linewidth} 
            \centering
            \includegraphics[width=1\linewidth]{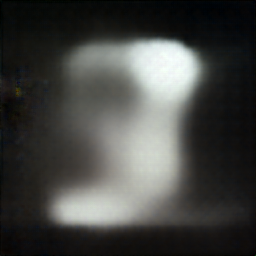}
            \includegraphics[width=1\linewidth]{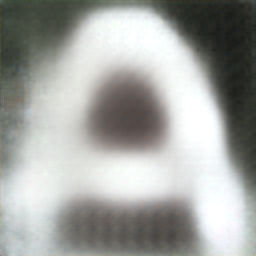}
        \end{minipage}
    }
    \hspace{-4.5mm}
    \subfigure{
        \begin{minipage}[t]{0.054\linewidth}
            \centering
            \includegraphics[width=1\linewidth]{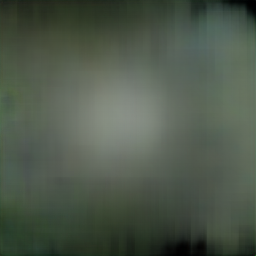}
            \includegraphics[width=1\linewidth]{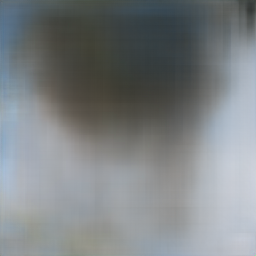}  
        \end{minipage}
        \hspace{-2.8mm}      
        \begin{minipage}[t]{0.054\linewidth} 
            \centering
            \includegraphics[width=1\linewidth]{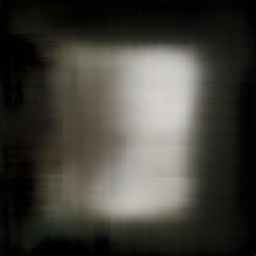}
            \includegraphics[width=1\linewidth]{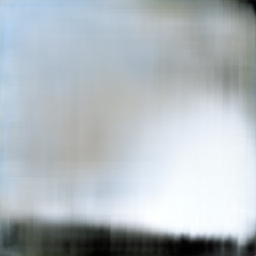}
        \end{minipage}
    }
     \hspace{-4.5mm}
    \subfigure{
        \begin{minipage}[t]{0.054\linewidth}
            \centering
            \includegraphics[width=1\linewidth]{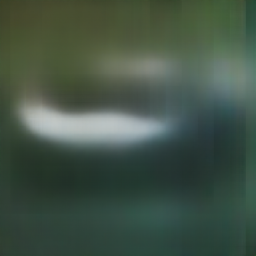}
            \includegraphics[width=1\linewidth]{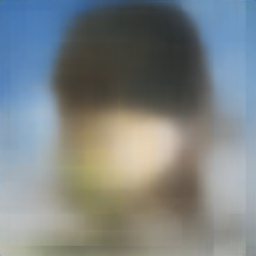}  
        \end{minipage}
        \hspace{-2.8mm}      
        \begin{minipage}[t]{0.054\linewidth} 
            \centering
            \includegraphics[width=1\linewidth]{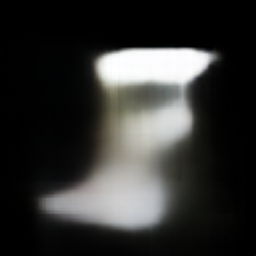}
            \includegraphics[width=1\linewidth]{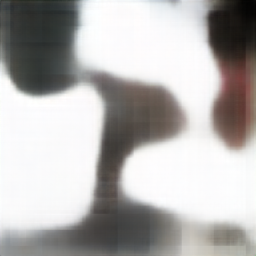}
        \end{minipage}
    }
     \hspace{-4.5mm}
    \subfigure{
        \begin{minipage}[t]{0.054\linewidth}
            \centering
            \includegraphics[width=1\linewidth]{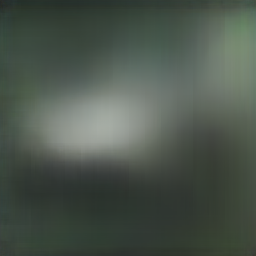}
            \includegraphics[width=1\linewidth]{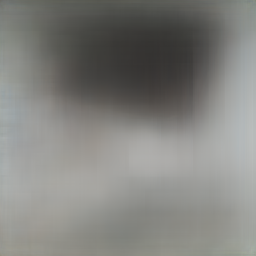}  
        \end{minipage}
        \hspace{-2.8mm}      
        \begin{minipage}[t]{0.054\linewidth} 
            \centering
            \includegraphics[width=1\linewidth]{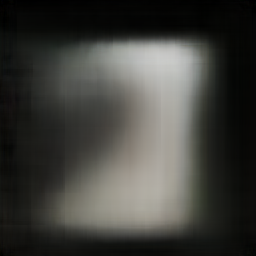}
            \includegraphics[width=1\linewidth]{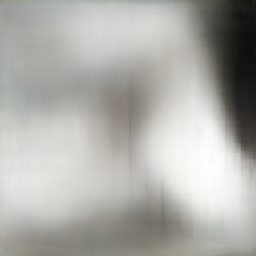}
        \end{minipage}
    }
     \hspace{-4.5mm}
    \subfigure{
        \begin{minipage}[t]{0.054\linewidth}
            \centering
            \includegraphics[width=1\linewidth]{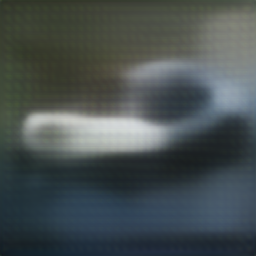}
            \includegraphics[width=1\linewidth]{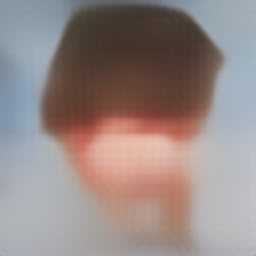}  
        \end{minipage}
        \hspace{-2.8mm}      
        \begin{minipage}[t]{0.054\linewidth} 
            \centering
            \includegraphics[width=1\linewidth]{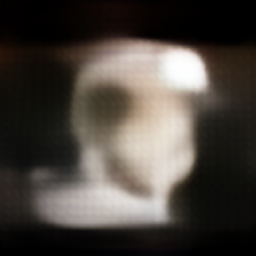}
            \includegraphics[width=1\linewidth]{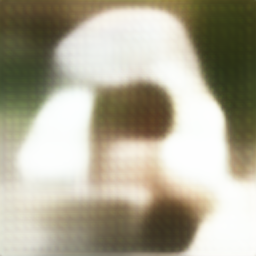}
        \end{minipage}
    }
     \hspace{-4.5mm}
    \subfigure{
        \begin{minipage}[t]{0.054\linewidth}
            \centering
            \includegraphics[width=1\linewidth]{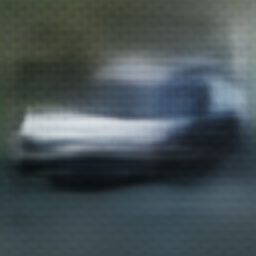}
            \includegraphics[width=1\linewidth]{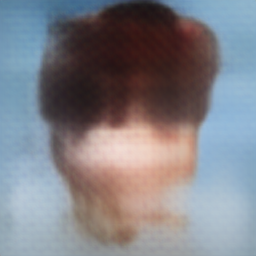}  
        \end{minipage}
        \hspace{-2.8mm}      
        \begin{minipage}[t]{0.054\linewidth} 
            \centering
            \includegraphics[width=1\linewidth]{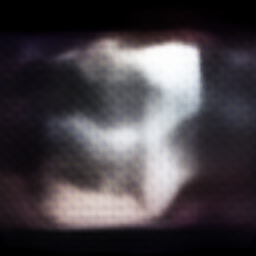}
            \includegraphics[width=1\linewidth]{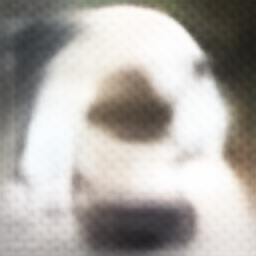}
        \end{minipage}
    }
    \hspace{-4.5mm}
    \subfigure{
        \begin{minipage}[t]{0.054\linewidth}
            \centering
            \includegraphics[width=1\linewidth]{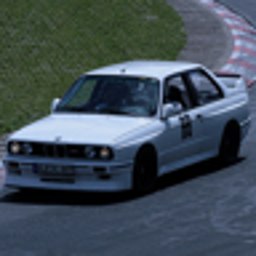}
            \includegraphics[width=1\linewidth]{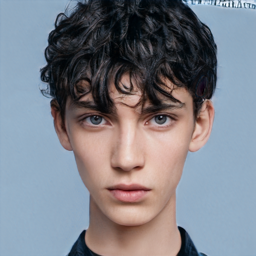}
        \end{minipage}
        \hspace{-2.8mm}      
        \begin{minipage}[t]{0.054\linewidth}
            \centering
            \includegraphics[width=1\linewidth]{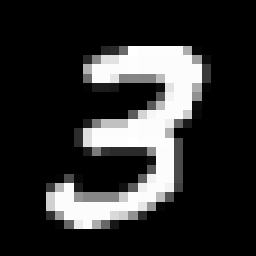}
            \includegraphics[width=1\linewidth]{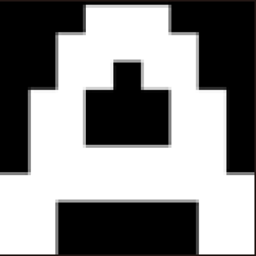}
        \end{minipage}
    }
        
    \vspace{-3mm}
    \setcounter{subfigure}{0}

    \subfigure{
         \rotatebox{90}{\makebox[0pt][c]{\vspace{1.5cm}\scriptsize{70;1;A;Wb}}}
        \begin{minipage}[t]{0.054\linewidth} 
            \centering
            \includegraphics[width=1\linewidth]{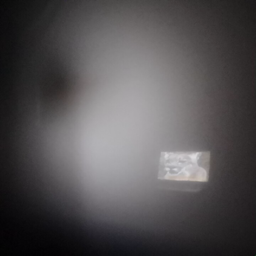}
            \includegraphics[width=1\linewidth]{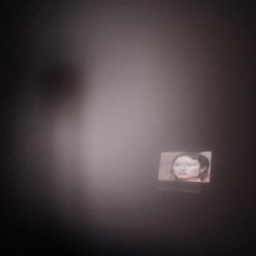}
        \end{minipage}
        \hspace{-2.8mm}
        \begin{minipage}[t]{0.054\linewidth}
            \centering
            \includegraphics[width=1\linewidth]{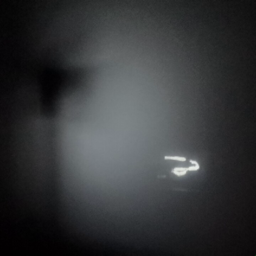}
            \includegraphics[width=1\linewidth]{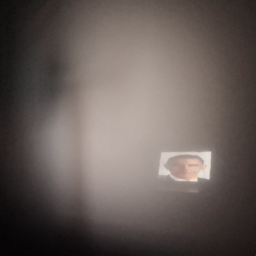}
        \end{minipage}
    }
    \hspace{-4.5mm}
    \subfigure{
        \begin{minipage}[t]{0.054\linewidth}
            \centering
            \includegraphics[width=1\linewidth]{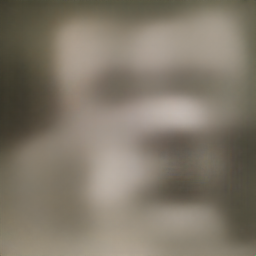}
            \includegraphics[width=1\linewidth]{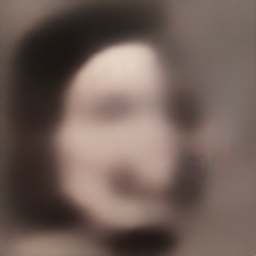}
        \end{minipage}
        \hspace{-2.8mm}      
        \begin{minipage}[t]{0.054\linewidth}
            \centering
            \includegraphics[width=1\linewidth]{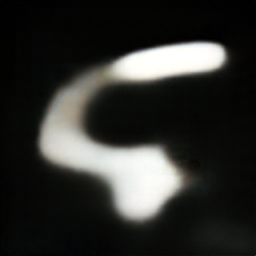}
            \includegraphics[width=1\linewidth]{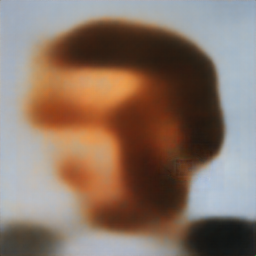}
        \end{minipage}
    }
    \hspace{-4.5mm}
    \subfigure{
        \begin{minipage}[t]{0.054\linewidth}
            \centering
            \includegraphics[width=1\linewidth]{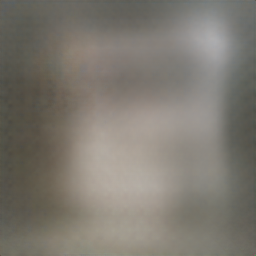}
            \includegraphics[width=1\linewidth]{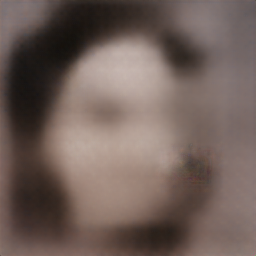}  
        \end{minipage}
        \hspace{-2.8mm}      
        \begin{minipage}[t]{0.054\linewidth} 
            \centering
            \includegraphics[width=1\linewidth]{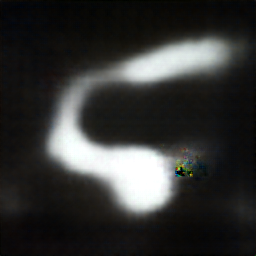}
            \includegraphics[width=1\linewidth]{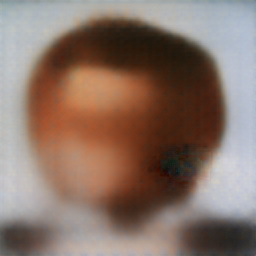}
        \end{minipage}
    }
    \hspace{-4.5mm}
    \subfigure{
        \begin{minipage}[t]{0.054\linewidth}
            \centering
            \includegraphics[width=1\linewidth]{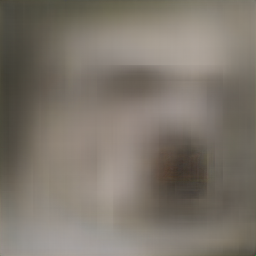}
            \includegraphics[width=1\linewidth]{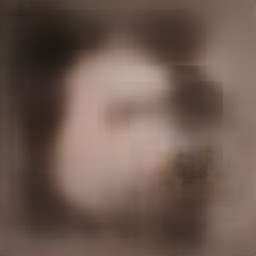}  
        \end{minipage}
        \hspace{-2.8mm}      
        \begin{minipage}[t]{0.054\linewidth} 
            \centering
            \includegraphics[width=1\linewidth]{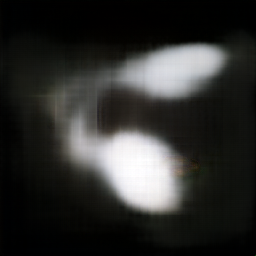}
            \includegraphics[width=1\linewidth]{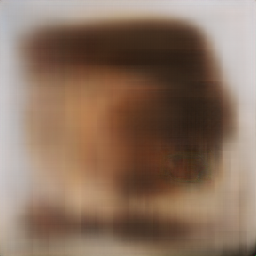}
        \end{minipage}
    }
     \hspace{-4.5mm}
    \subfigure{
        \begin{minipage}[t]{0.054\linewidth}
            \centering
            \includegraphics[width=1\linewidth]{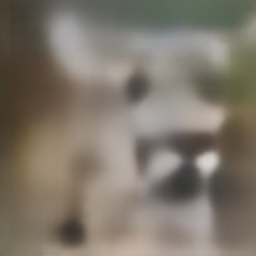}
            \includegraphics[width=1\linewidth]{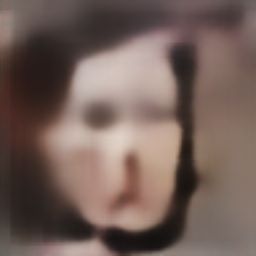}  
        \end{minipage}
        \hspace{-2.8mm}      
        \begin{minipage}[t]{0.054\linewidth} 
            \centering
            \includegraphics[width=1\linewidth]{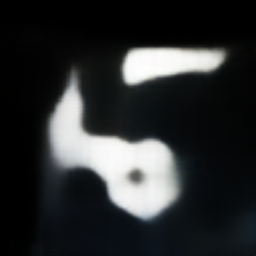}
            \includegraphics[width=1\linewidth]{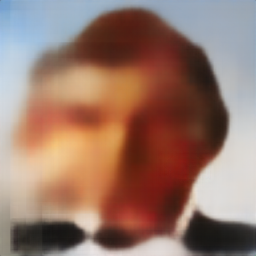}
        \end{minipage}
    }
     \hspace{-4.5mm}
    \subfigure{
        \begin{minipage}[t]{0.054\linewidth}
            \centering
            \includegraphics[width=1\linewidth]{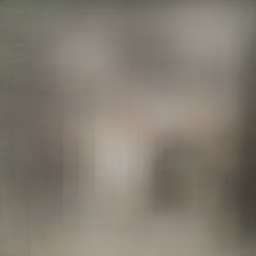}
            \includegraphics[width=1\linewidth]{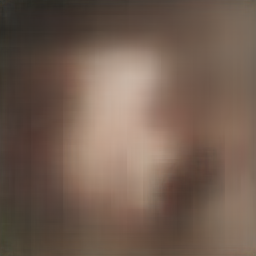}  
        \end{minipage}
        \hspace{-2.8mm}      
        \begin{minipage}[t]{0.054\linewidth} 
            \centering
            \includegraphics[width=1\linewidth]{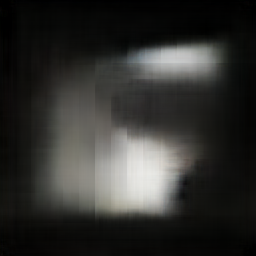}
            \includegraphics[width=1\linewidth]{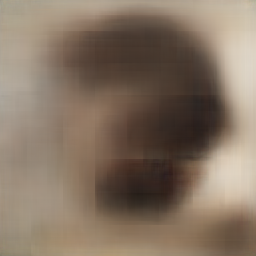}
        \end{minipage}
    }
     \hspace{-4.5mm}
    \subfigure{
        \begin{minipage}[t]{0.054\linewidth}
            \centering
            \includegraphics[width=1\linewidth]{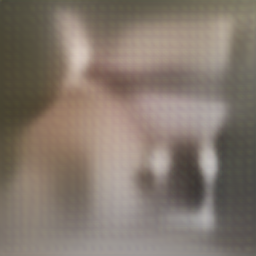}
            \includegraphics[width=1\linewidth]{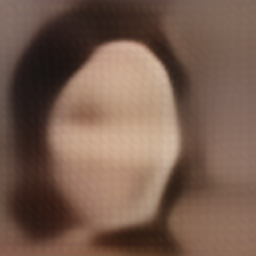}  
        \end{minipage}
        \hspace{-2.8mm}      
        \begin{minipage}[t]{0.054\linewidth} 
            \centering
            \includegraphics[width=1\linewidth]{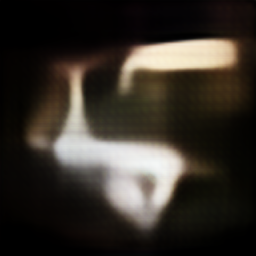}
            \includegraphics[width=1\linewidth]{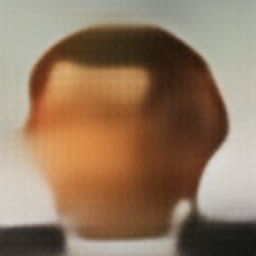}
        \end{minipage}
    }
     \hspace{-4.5mm}
    \subfigure{
        \begin{minipage}[t]{0.054\linewidth}
            \centering
            \includegraphics[width=1\linewidth]{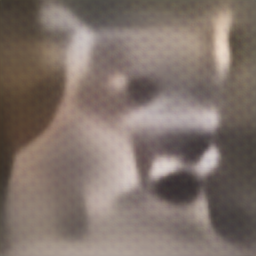}
            \includegraphics[width=1\linewidth]{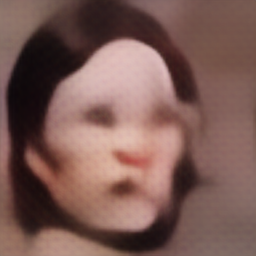}  
        \end{minipage}
        \hspace{-2.8mm}      
        \begin{minipage}[t]{0.054\linewidth} 
            \centering
            \includegraphics[width=1\linewidth]{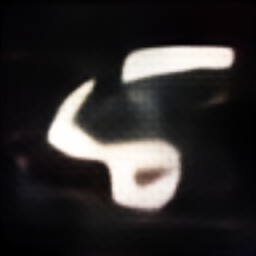}
            \includegraphics[width=1\linewidth]{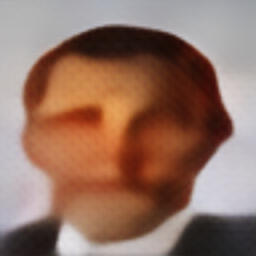}
        \end{minipage}
    }
    \hspace{-4.5mm}
    \subfigure{
        \begin{minipage}[t]{0.054\linewidth}
            \centering
            \includegraphics[width=1\linewidth]{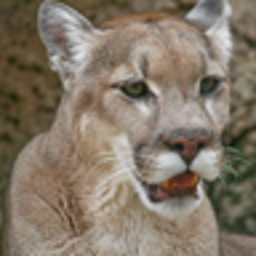}
            \includegraphics[width=1\linewidth]{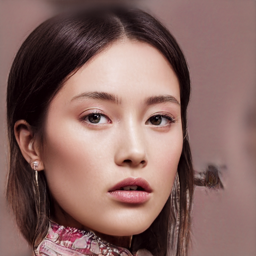}
        \end{minipage}
        \hspace{-2.8mm}      
        \begin{minipage}[t]{0.054\linewidth}
            \centering
            \includegraphics[width=1\linewidth]{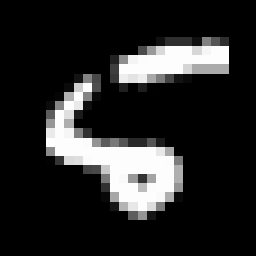}
            \includegraphics[width=1\linewidth]{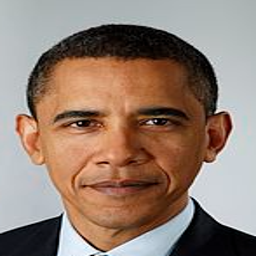}
        \end{minipage}
    }
        
    \vspace{-3mm}
    \setcounter{subfigure}{0}

    \subfigure{
         \rotatebox{90}{\makebox[0pt][c]{\vspace{1.5cm}\scriptsize{100;1;A;Wb}}}
        \begin{minipage}[t]{0.054\linewidth} 
            \centering
            \includegraphics[width=1\linewidth]{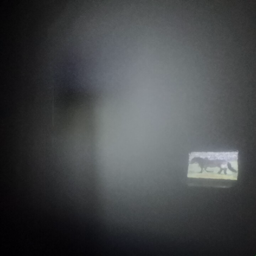}
            \includegraphics[width=1\linewidth]{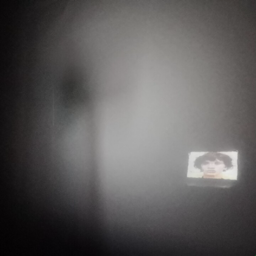}
        \end{minipage}
        \hspace{-2.8mm}
        \begin{minipage}[t]{0.054\linewidth}
            \centering
            \includegraphics[width=1\linewidth]{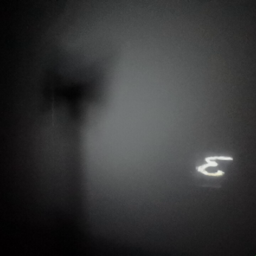}
            \includegraphics[width=1\linewidth]{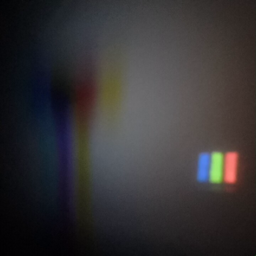}
        \end{minipage}
    }
    \hspace{-4.5mm}
    \subfigure{
        \begin{minipage}[t]{0.054\linewidth}
            \centering
            \includegraphics[width=1\linewidth]{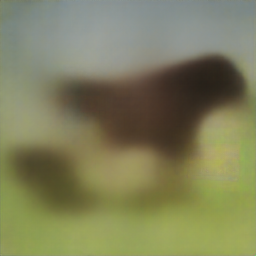}
            \includegraphics[width=1\linewidth]{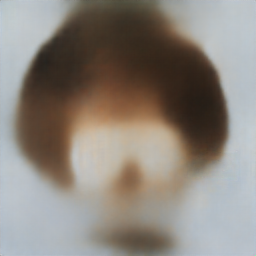}
        \end{minipage}
        \hspace{-2.8mm}      
        \begin{minipage}[t]{0.054\linewidth}
            \centering
            \includegraphics[width=1\linewidth]{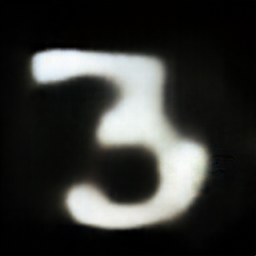}
            \includegraphics[width=1\linewidth]{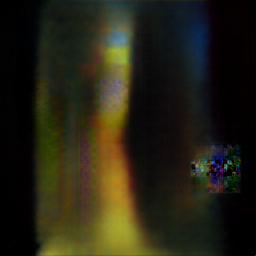}
        \end{minipage}
    }
    \hspace{-4.5mm}
    \subfigure{
        \begin{minipage}[t]{0.054\linewidth}
            \centering
            \includegraphics[width=1\linewidth]{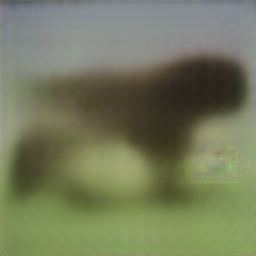}
            \includegraphics[width=1\linewidth]{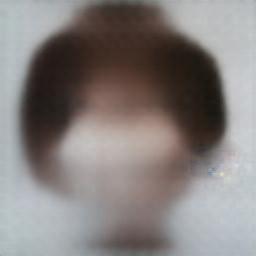}  
        \end{minipage}
        \hspace{-2.8mm}      
        \begin{minipage}[t]{0.054\linewidth} 
            \centering
            \includegraphics[width=1\linewidth]{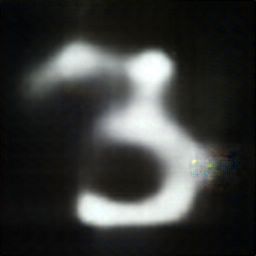}
            \includegraphics[width=1\linewidth]{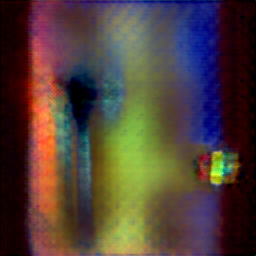}
        \end{minipage}
    }
    \hspace{-4.5mm}
    \subfigure{
        \begin{minipage}[t]{0.054\linewidth}
            \centering
            \includegraphics[width=1\linewidth]{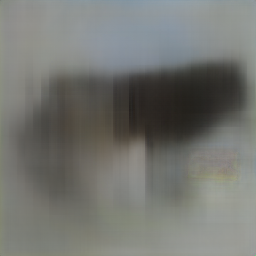}
            \includegraphics[width=1\linewidth]{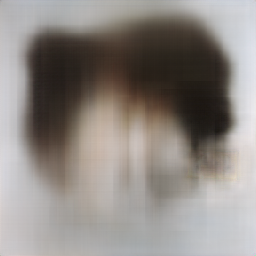}  
        \end{minipage}
        \hspace{-2.8mm}      
        \begin{minipage}[t]{0.054\linewidth} 
            \centering
            \includegraphics[width=1\linewidth]{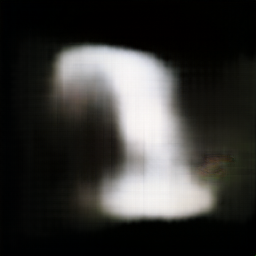}
            \includegraphics[width=1\linewidth]{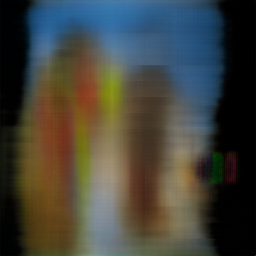}
        \end{minipage}
    }
     \hspace{-4.5mm}
    \subfigure{
        \begin{minipage}[t]{0.054\linewidth}
            \centering
            \includegraphics[width=1\linewidth]{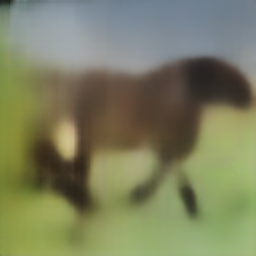}
            \includegraphics[width=1\linewidth]{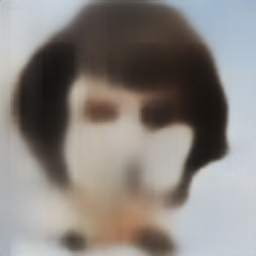}  
        \end{minipage}
        \hspace{-2.8mm}      
        \begin{minipage}[t]{0.054\linewidth} 
            \centering
            \includegraphics[width=1\linewidth]{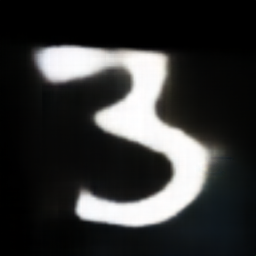}
            \includegraphics[width=1\linewidth]{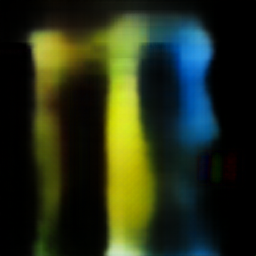}
        \end{minipage}
    }
     \hspace{-4.5mm}
    \subfigure{
        \begin{minipage}[t]{0.054\linewidth}
            \centering
            \includegraphics[width=1\linewidth]{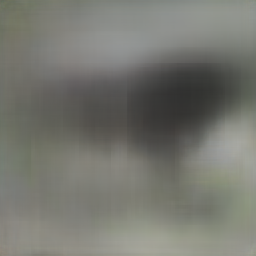}
            \includegraphics[width=1\linewidth]{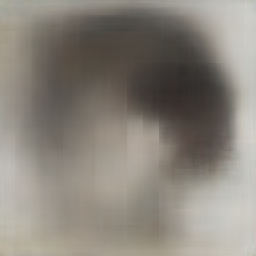}  
        \end{minipage}
        \hspace{-2.8mm}      
        \begin{minipage}[t]{0.054\linewidth} 
            \centering
            \includegraphics[width=1\linewidth]{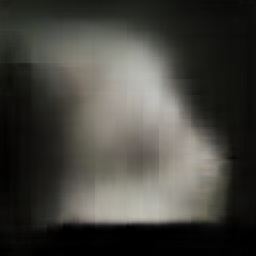}
            \includegraphics[width=1\linewidth]{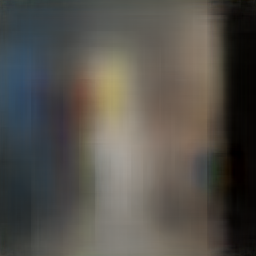}
        \end{minipage}
    }
     \hspace{-4.5mm}
    \subfigure{
        \begin{minipage}[t]{0.054\linewidth}
            \centering
            \includegraphics[width=1\linewidth]{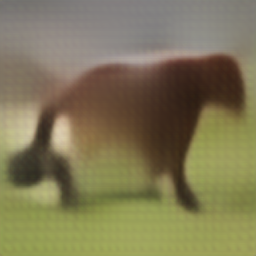}
            \includegraphics[width=1\linewidth]{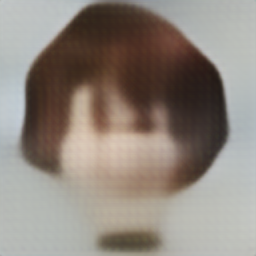}  
        \end{minipage}
        \hspace{-2.8mm}      
        \begin{minipage}[t]{0.054\linewidth} 
            \centering
            \includegraphics[width=1\linewidth]{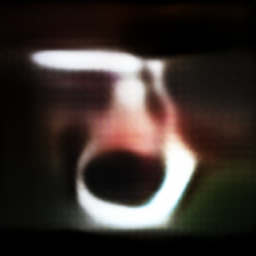}
            \includegraphics[width=1\linewidth]{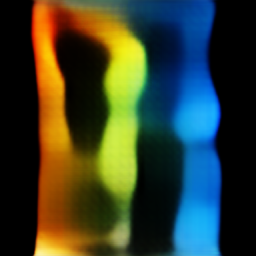}
        \end{minipage}
    }
     \hspace{-4.5mm}
    \subfigure{
        \begin{minipage}[t]{0.054\linewidth}
            \centering
            \includegraphics[width=1\linewidth]{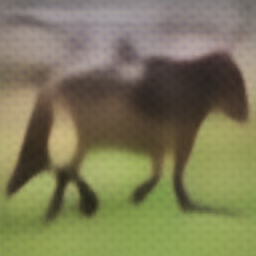}
            \includegraphics[width=1\linewidth]{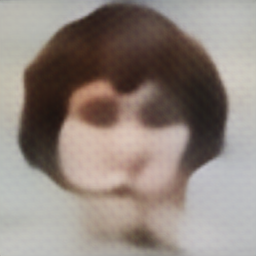}  
        \end{minipage}
        \hspace{-2.8mm}      
        \begin{minipage}[t]{0.054\linewidth} 
            \centering
            \includegraphics[width=1\linewidth]{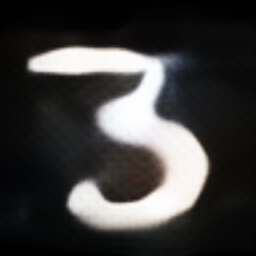}
            \includegraphics[width=1\linewidth]{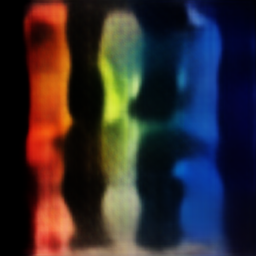}
        \end{minipage}
    }
    \hspace{-4.5mm}
    \subfigure{
        \begin{minipage}[t]{0.054\linewidth}
            \centering
            \includegraphics[width=1\linewidth]{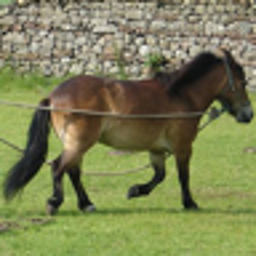}
            \includegraphics[width=1\linewidth]{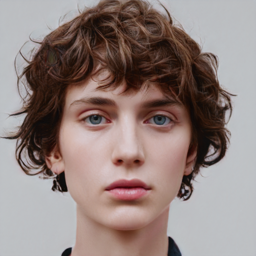}
        \end{minipage}
        \hspace{-2.8mm}      
        \begin{minipage}[t]{0.054\linewidth}
            \centering
            \includegraphics[width=1\linewidth]{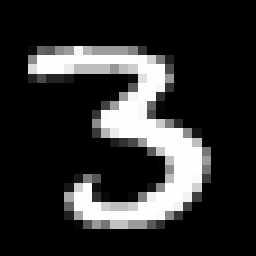}
            \includegraphics[width=1\linewidth]{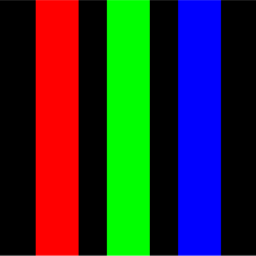}
        \end{minipage}
    }   
    
    \vspace{-3mm}
    \setcounter{subfigure}{0}

    \subfigure{
         \rotatebox{90}{\makebox[0pt][c]{\vspace{1.5cm}\scriptsize{70;2;A;Wb}}}
        \begin{minipage}[t]{0.054\linewidth} 
            \centering
            \includegraphics[width=1\linewidth]{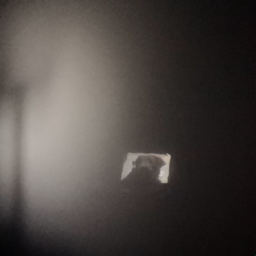}
            \includegraphics[width=1\linewidth]{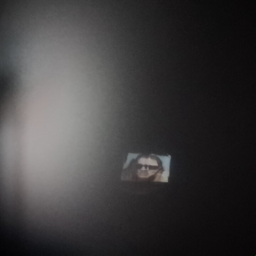}
        \end{minipage}
        \hspace{-2.8mm}
        \begin{minipage}[t]{0.054\linewidth}
            \centering
            \includegraphics[width=1\linewidth]{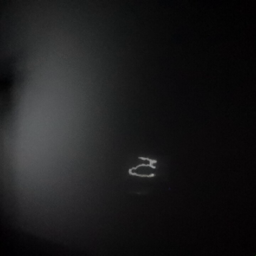}
            \includegraphics[width=1\linewidth]{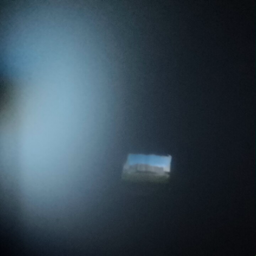}
        \end{minipage}
    }
    \hspace{-4.5mm}
    \subfigure{
        \begin{minipage}[t]{0.054\linewidth}
            \includegraphics[width=1\linewidth]{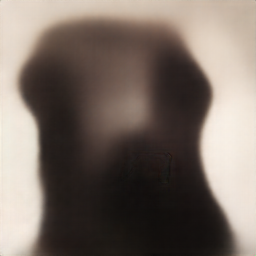}
            \includegraphics[width=1\linewidth]{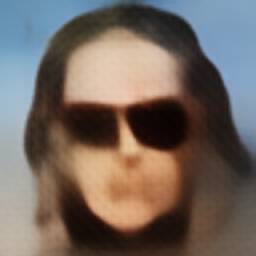}
        \end{minipage}
        \hspace{-2.8mm}      
        \begin{minipage}[t]{0.054\linewidth}
            \centering
            \includegraphics[width=1\linewidth]{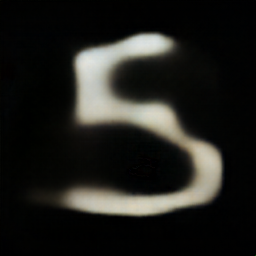}
            \includegraphics[width=1\linewidth]{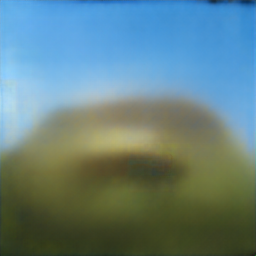}
        \end{minipage}
    }
    \hspace{-4.5mm}
    \subfigure{
        \begin{minipage}[t]{0.054\linewidth}
            \centering
            \includegraphics[width=1\linewidth]{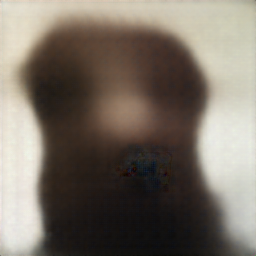}
            \includegraphics[width=1\linewidth]{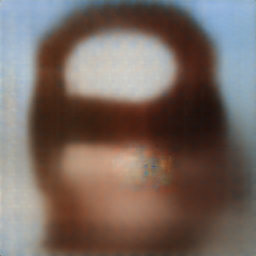}  
        \end{minipage}
        \hspace{-2.8mm}      
        \begin{minipage}[t]{0.054\linewidth} 
            \centering
            \includegraphics[width=1\linewidth]{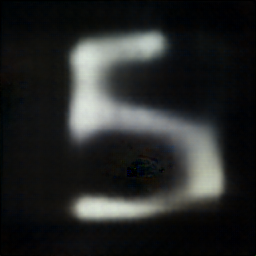}
            \includegraphics[width=1\linewidth]{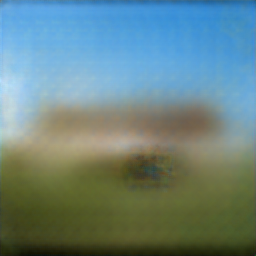}
        \end{minipage}
    }
    \hspace{-4.5mm}
    \subfigure{
        \begin{minipage}[t]{0.054\linewidth}
            \centering
            \includegraphics[width=1\linewidth]{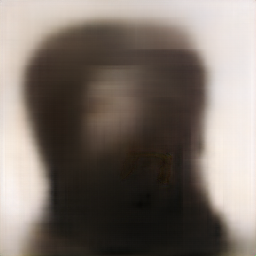}
            \includegraphics[width=1\linewidth]{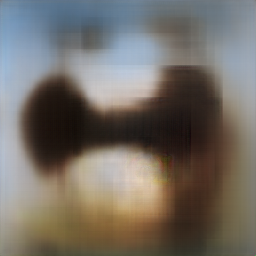}  
        \end{minipage}
        \hspace{-2.8mm}      
        \begin{minipage}[t]{0.054\linewidth} 
            \centering
            \includegraphics[width=1\linewidth]{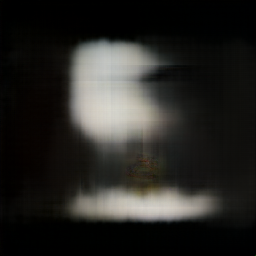}
            \includegraphics[width=1\linewidth]{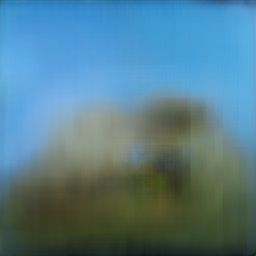}
        \end{minipage}
    }
     \hspace{-4.5mm}
    \subfigure{
        \begin{minipage}[t]{0.054\linewidth}
            \centering
            \includegraphics[width=1\linewidth]{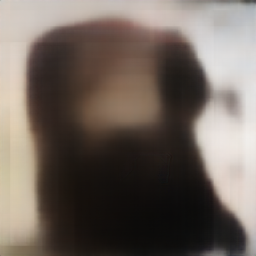}
            \includegraphics[width=1\linewidth]{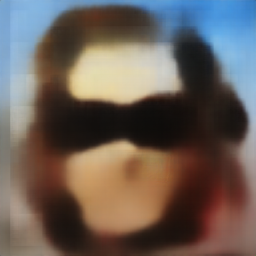}  
        \end{minipage}
        \hspace{-2.8mm}      
        \begin{minipage}[t]{0.054\linewidth} 
            \centering
            \includegraphics[width=1\linewidth]{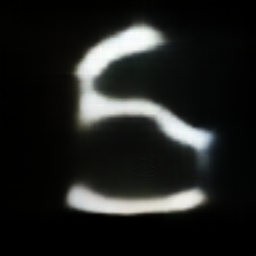}
            \includegraphics[width=1\linewidth]{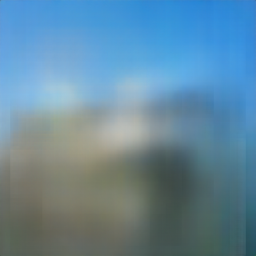}
        \end{minipage}
    }
     \hspace{-4.5mm}
    \subfigure{
        \begin{minipage}[t]{0.054\linewidth}
            \centering
            \includegraphics[width=1\linewidth]{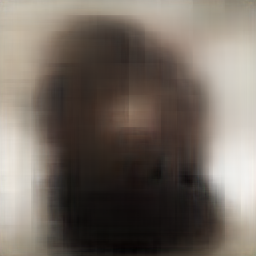}
            \includegraphics[width=1\linewidth]{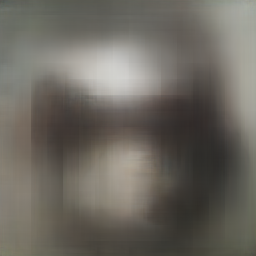}  
        \end{minipage}
        \hspace{-2.8mm}      
        \begin{minipage}[t]{0.054\linewidth} 
            \centering
            \includegraphics[width=1\linewidth]{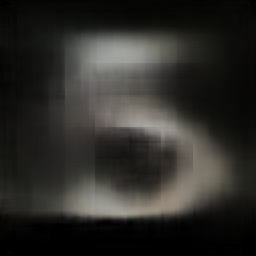}
            \includegraphics[width=1\linewidth]{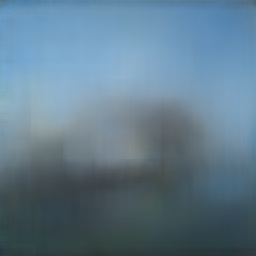}
        \end{minipage}
    }
     \hspace{-4.5mm}
    \subfigure{
        \begin{minipage}[t]{0.054\linewidth}
            \centering
            \includegraphics[width=1\linewidth]{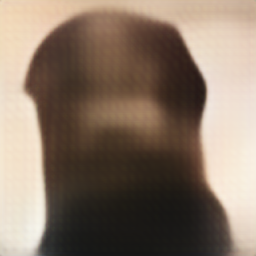}
            \includegraphics[width=1\linewidth]{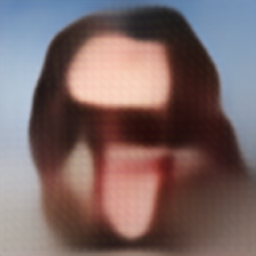}  
        \end{minipage}
        \hspace{-2.8mm}      
        \begin{minipage}[t]{0.054\linewidth} 
            \centering
            \includegraphics[width=1\linewidth]{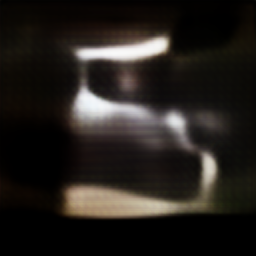}
            \includegraphics[width=1\linewidth]{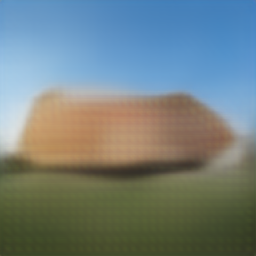}
        \end{minipage}
    }
     \hspace{-4.5mm}
    \subfigure{
        \begin{minipage}[t]{0.054\linewidth}
            \centering
            \includegraphics[width=1\linewidth]{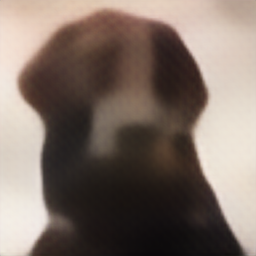}
            \includegraphics[width=1\linewidth]{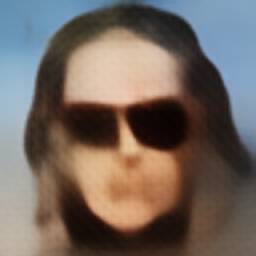}
        \end{minipage}
        \hspace{-2.8mm}      
        \begin{minipage}[t]{0.054\linewidth} 
            \centering
            \includegraphics[width=1\linewidth]{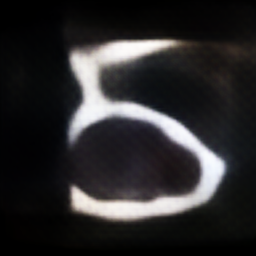}
            \includegraphics[width=1\linewidth]{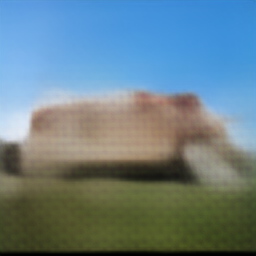}
        \end{minipage}
    }
    \hspace{-4.5mm}
    \subfigure{
        \begin{minipage}[t]{0.054\linewidth}
            \centering
            \includegraphics[width=1\linewidth]{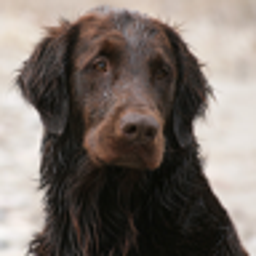}
            \includegraphics[width=1\linewidth]{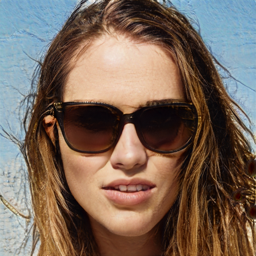}
        \end{minipage}
        \hspace{-2.8mm}      
        \begin{minipage}[t]{0.054\linewidth}
            \centering
            \includegraphics[width=1\linewidth]{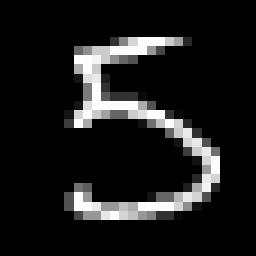}
            \includegraphics[width=1\linewidth]{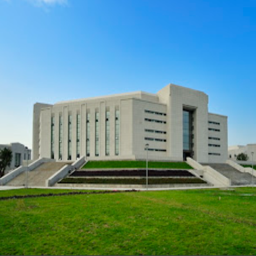}
        \end{minipage}
    }

    \vspace{-3mm}
    \setcounter{subfigure}{0}

    \subfigure[\tiny{INPUT}]{
         \rotatebox{90}{\makebox[0pt][c]{\vspace{1.5cm}\scriptsize{100;2;A;Wb}}}
        \begin{minipage}[t]{0.054\linewidth} 
            \centering
            \includegraphics[width=1\linewidth]{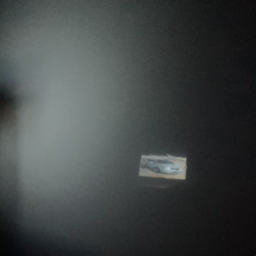}
            \includegraphics[width=1\linewidth]{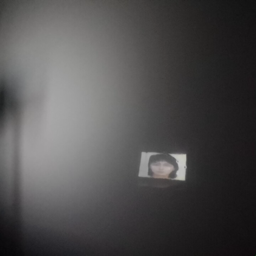}
        \end{minipage}
        \hspace{-2.8mm}
        \begin{minipage}[t]{0.054\linewidth}
            \centering
            \includegraphics[width=1\linewidth]{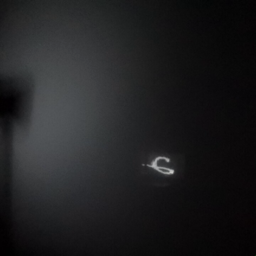}
            \includegraphics[width=1\linewidth]{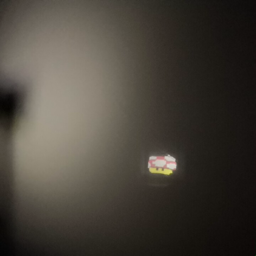}
        \end{minipage}
    }
    \hspace{-4.5mm}
    \subfigure[\tiny{KBNet}]{
        \begin{minipage}[t]{0.054\linewidth}
            \centering
            \includegraphics[width=1\linewidth]{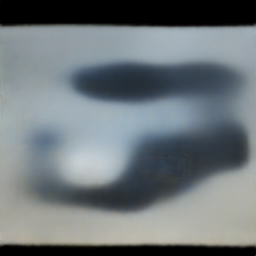}
            \includegraphics[width=1\linewidth]{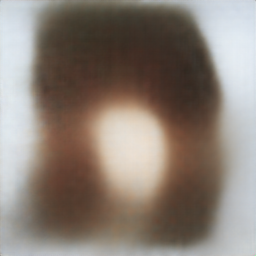}
        \end{minipage}
        \hspace{-2.8mm}      
        \begin{minipage}[t]{0.054\linewidth}
            \centering
            \includegraphics[width=1\linewidth]{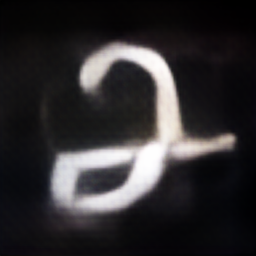}
            \includegraphics[width=1\linewidth]{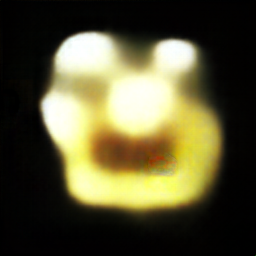}
        \end{minipage}
    }
    \hspace{-4.5mm}
    \subfigure[\tiny{NAFNet}]{
        \begin{minipage}[t]{0.054\linewidth}
            \centering
            \includegraphics[width=1\linewidth]{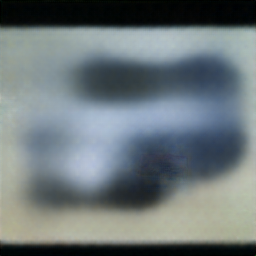}
            \includegraphics[width=1\linewidth]{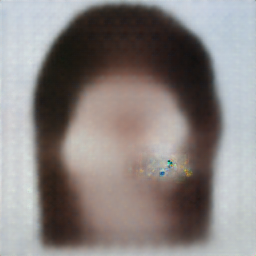}  
        \end{minipage}
        \hspace{-2.8mm}      
        \begin{minipage}[t]{0.054\linewidth} 
            \centering
            \includegraphics[width=1\linewidth]{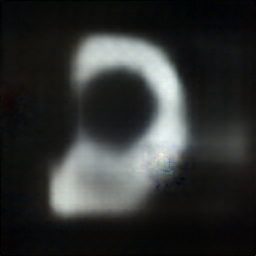}
            \includegraphics[width=1\linewidth]{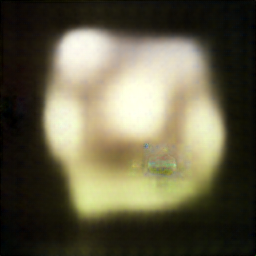}
        \end{minipage}
    }
    \hspace{-4.5mm}
    \subfigure[\tiny{MSDINet}]{
        \begin{minipage}[t]{0.054\linewidth}
            \centering
            \includegraphics[width=1\linewidth]{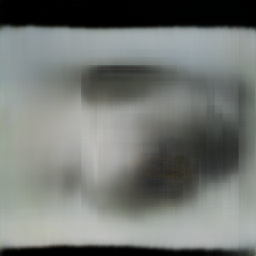}
            \includegraphics[width=1\linewidth]{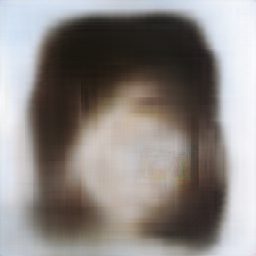}  
        \end{minipage}
        \hspace{-2.8mm}      
        \begin{minipage}[t]{0.054\linewidth} 
            \centering
            \includegraphics[width=1\linewidth]{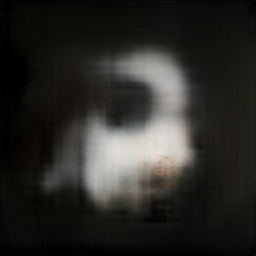}
            \includegraphics[width=1\linewidth]{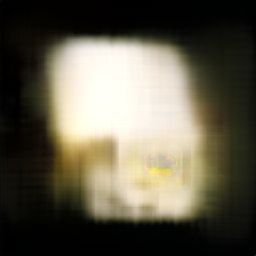}
        \end{minipage}
    }
     \hspace{-4.5mm}
    \subfigure[\tiny{Uformer}]{
        \begin{minipage}[t]{0.054\linewidth}
            \centering
            \includegraphics[width=1\linewidth]{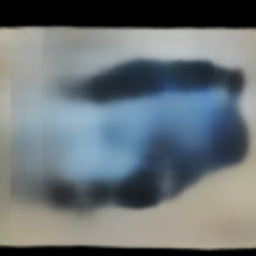}
            \includegraphics[width=1\linewidth]{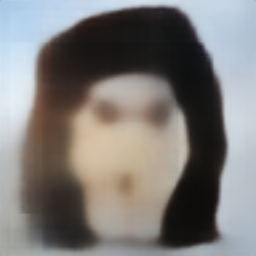}  
        \end{minipage}
        \hspace{-2.8mm}      
        \begin{minipage}[t]{0.054\linewidth} 
            \centering
            \includegraphics[width=1\linewidth]{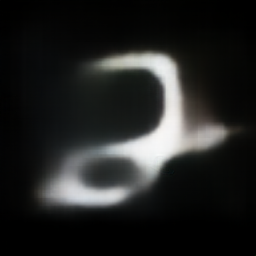}
            \includegraphics[width=1\linewidth]{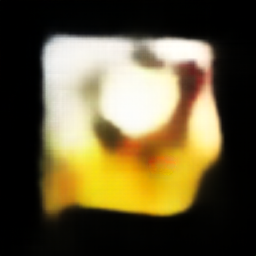}
        \end{minipage}
    }
     \hspace{-4.5mm}
    \subfigure[\tiny{SwinIR}]{
        \begin{minipage}[t]{0.054\linewidth}
            \centering
            \includegraphics[width=1\linewidth]{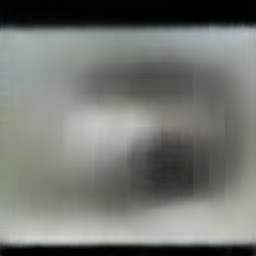}
            \includegraphics[width=1\linewidth]{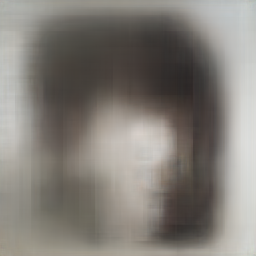}  
        \end{minipage}
        \hspace{-2.8mm}      
        \begin{minipage}[t]{0.054\linewidth} 
            \centering
            \includegraphics[width=1\linewidth]{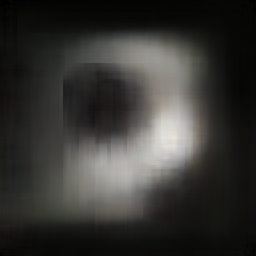}
            \includegraphics[width=1\linewidth]{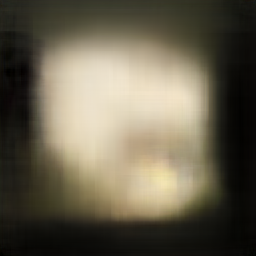}
        \end{minipage}
    }
     \hspace{-4.5mm}
    \subfigure[\tiny{NLOS-OT}]{
        \begin{minipage}[t]{0.054\linewidth}
            \centering
            \includegraphics[width=1\linewidth]{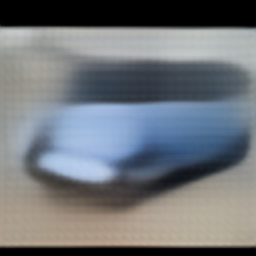}
            \includegraphics[width=1\linewidth]{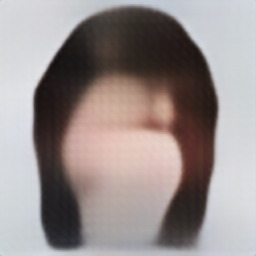}  
        \end{minipage}
        \hspace{-2.8mm}      
        \begin{minipage}[t]{0.054\linewidth} 
            \centering
            \includegraphics[width=1\linewidth]{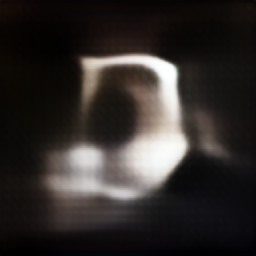}
            \includegraphics[width=1\linewidth]{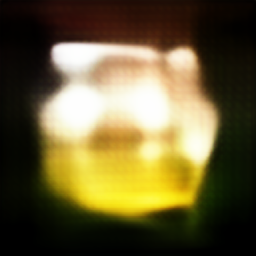}
        \end{minipage}
    }
     \hspace{-4.5mm}
    \subfigure[\tiny{\textbf{NLOS-LTM}}]{
        \begin{minipage}[t]{0.054\linewidth}
            \centering
            \includegraphics[width=1\linewidth]{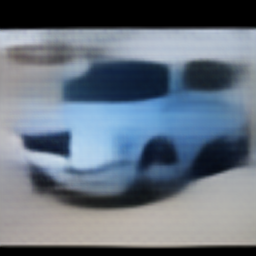}
            \includegraphics[width=1\linewidth]{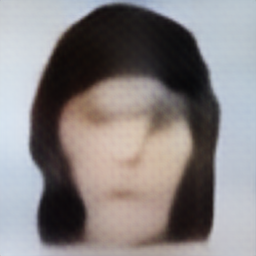}  
        \end{minipage}
        \hspace{-2.8mm}      
        \begin{minipage}[t]{0.054\linewidth} 
            \centering
            \includegraphics[width=1\linewidth]{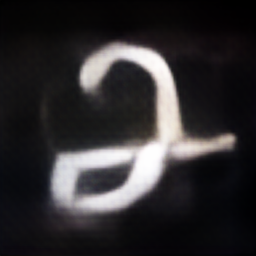}
            \includegraphics[width=1\linewidth]{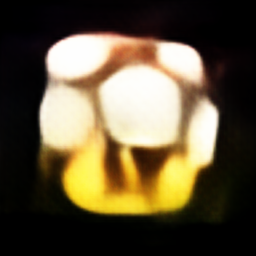}
        \end{minipage}
    }
    \hspace{-4.5mm}
    \subfigure[\tiny{GT}]{
        \begin{minipage}[t]{0.054\linewidth}
            \centering
            \includegraphics[width=1\linewidth]{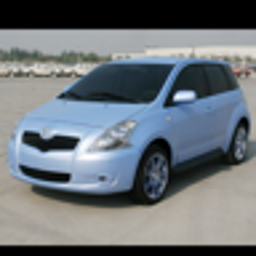}
            \includegraphics[width=1\linewidth]{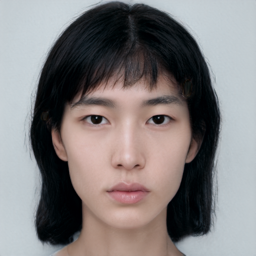}
        \end{minipage}
        \hspace{-2.8mm}      
        \begin{minipage}[t]{0.054\linewidth}
            \centering
            \includegraphics[width=1\linewidth]{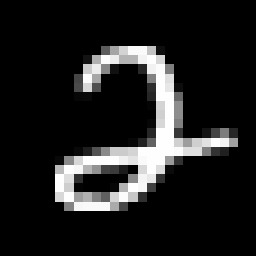}
            \includegraphics[width=1\linewidth]{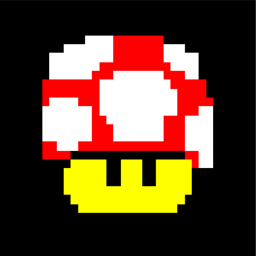}
        \end{minipage}
    }
    
    \caption{Visual comparison of the generalization ability. The models are trained on STL-10, and tested on STL-10 as well as images from other datasets. The data for each method under each optical transport condition comes from four datasets, listed from top to bottom and left to right as STL-10, MNIST, Supermodel, and Real.}
    \label{Fig-Generalization}
\end{figure*}

\textit{1) Cross-Datasets Generalization Ability:}
To evaluate the cross-datasets generalization ability of our method, we conduct experiments using the projections of the STL-10 dataset as the training set. Since in the publicly available dataset NLOS-Passive, the STL-10 data were collected only under two conditions, which can not reflect the multi-condition handling ability of our method, we recollect a new version of the dataset containing more light transport conditions (similar to NLOS-Passive, an occluder is added in the scene to reduce the complexity of the problem). We then test the model on projection images corresponding to other datasets. In addition to the MNIST and Supermodel images, we add another set of images for this test. It contains the four testing images used by \cite{saunders2019computational} and the ``Real Images'' dataset used in \cite{geng2021passive}, which is named as Real. The results are reported in Table \ref{Table-Generalization-8} and Fig. \ref{Fig-Generalization}. We find that although the STL-10 images are quite different from the MNIST, Supermodel and Real images, we are still able to get reasonable reconstructions with NLOS-LTM. 
Although the images are blurry, the shape and profile of the objects are similar with the ground truth.

\begin{figure}[!t]
    \centering
    \subfigure{
        \rotatebox{90}{\tiny{\quad \ INPUT}}
        \begin{minipage}[t]{0.105\linewidth} 
            \centering
            \includegraphics[width=0.95\linewidth]{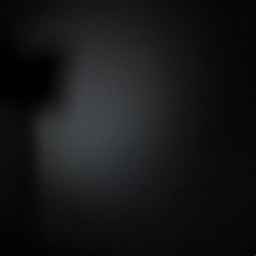}
        \end{minipage}
    }
    \hspace{-5mm}
    \subfigure{
        \begin{minipage}[t]{0.105\linewidth}
            \centering
            \includegraphics[width=0.95\linewidth]{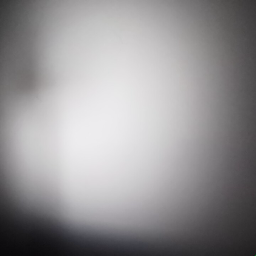}
        \end{minipage}
    }
    \hspace{-5mm}
    \subfigure{
        \begin{minipage}[t]{0.105\linewidth}
            \centering
            \includegraphics[width=0.95\linewidth]{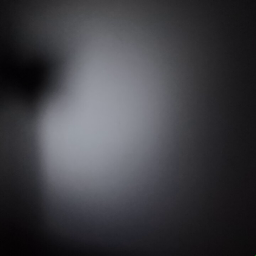}
        \end{minipage}
    }
    \hspace{-5mm}
    \subfigure{
        \begin{minipage}[t]{0.105\linewidth}
            \centering
            \includegraphics[width=0.95\linewidth]{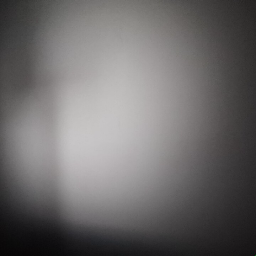}
        \end{minipage}
    }
    \hspace{-5mm}
    \subfigure{
        \begin{minipage}[t]{0.105\linewidth}
            \centering
            \includegraphics[width=0.95\linewidth]{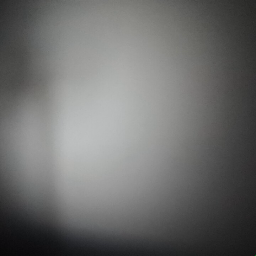}
        \end{minipage}
    }
    \hspace{-5mm}
    \subfigure{
        \begin{minipage}[t]{0.105\linewidth}
            \centering
            \includegraphics[width=0.95\linewidth]{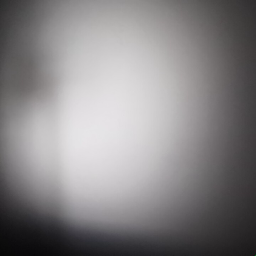}
        \end{minipage}
    }
    \hspace{-5mm}
    \subfigure{
        \begin{minipage}[t]{0.105\linewidth}
            \centering
            \includegraphics[width=0.95\linewidth]{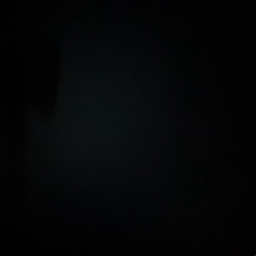}
        \end{minipage}
    }
    \hspace{-5mm}
    \subfigure{
        \begin{minipage}[t]{0.105\linewidth}
            \centering
            \includegraphics[width=0.95\linewidth]{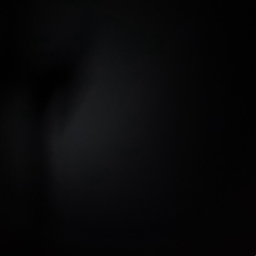}
        \end{minipage}
    }
    \hspace{-5mm}
    \subfigure{
        \begin{minipage}[t]{0.105\linewidth}
            \centering
            \includegraphics[width=0.95\linewidth]{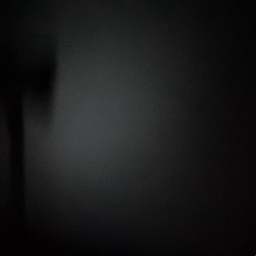}
        \end{minipage}
    }
    
    \vspace{-3mm}
    \setcounter{subfigure}{0}
    
    \subfigure{
        \rotatebox{90}{\tiny{\textbf{NLOS-LTM}}}
        \begin{minipage}[t]{0.105\linewidth} 
            \centering
            \includegraphics[width=0.95\linewidth]{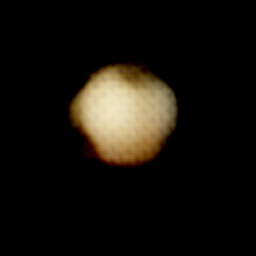}
        \end{minipage}
    }
    \hspace{-5mm}
    \subfigure{
        \begin{minipage}[t]{0.105\linewidth}
            \centering
            \includegraphics[width=0.95\linewidth]{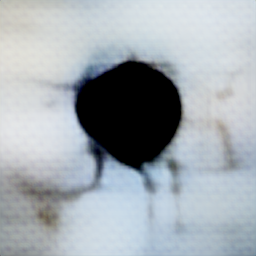}
        \end{minipage}
    }
    \hspace{-5mm}
    \subfigure{
        \begin{minipage}[t]{0.105\linewidth}
            \centering
            \includegraphics[width=0.95\linewidth]{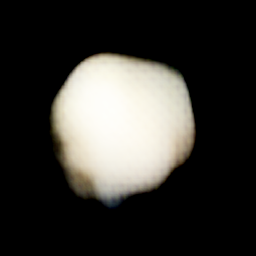}
        \end{minipage}
    }
    \hspace{-5mm}
    \subfigure{
        \begin{minipage}[t]{0.105\linewidth}
            \centering
            \includegraphics[width=0.95\linewidth]{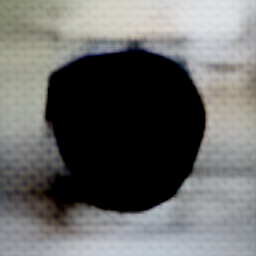}
        \end{minipage}
    }
    \hspace{-5mm}
    \subfigure{
        \begin{minipage}[t]{0.105\linewidth}
            \centering
            \includegraphics[width=0.95\linewidth]{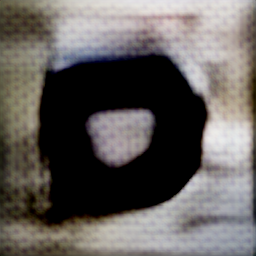}
        \end{minipage}
    }
    \hspace{-5mm}
    \subfigure{
        \begin{minipage}[t]{0.105\linewidth}
            \centering
            \includegraphics[width=0.95\linewidth]{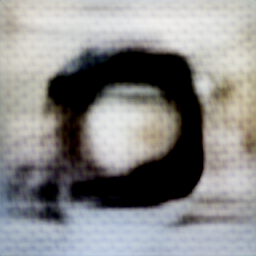}
        \end{minipage}
    }
    \hspace{-5mm}
    \subfigure{
        \begin{minipage}[t]{0.105\linewidth}
            \centering
            \includegraphics[width=0.95\linewidth]{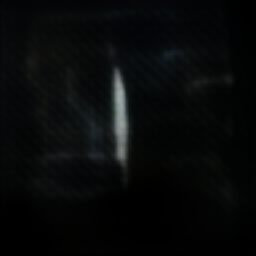}
        \end{minipage}
    }
    \hspace{-5mm}
    \subfigure{
        \begin{minipage}[t]{0.105\linewidth}
            \centering
            \includegraphics[width=0.95\linewidth]{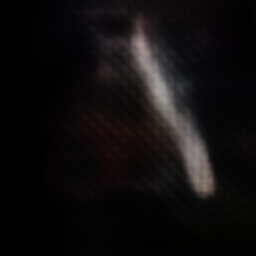}
        \end{minipage}
    }
    \hspace{-5mm}
    \subfigure{
        \begin{minipage}[t]{0.105\linewidth}
            \centering
            \includegraphics[width=0.95\linewidth]{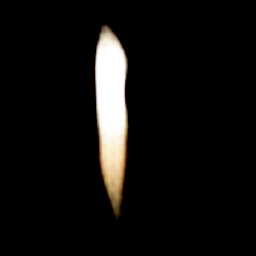}
        \end{minipage}
    }
    
    \vspace{-3mm}
    \setcounter{subfigure}{0}
    
    \subfigure{
        \rotatebox{90}{\tiny{\quad \quad GT}}
        \begin{minipage}[t]{0.105\linewidth}
            \centering
            \includegraphics[width=0.95\linewidth]{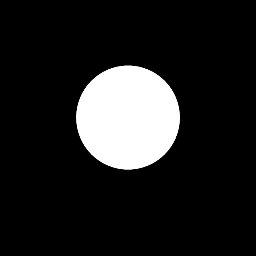}
        \end{minipage}
    }
    \hspace{-5mm}
    \subfigure{
        \begin{minipage}[t]{0.105\linewidth}
            \centering
            \includegraphics[width=0.95\linewidth]{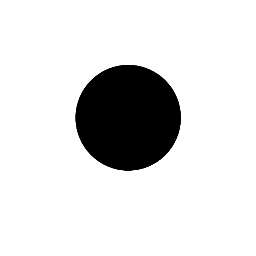}
        \end{minipage}
    }
    \hspace{-5mm}
    \subfigure{
        \begin{minipage}[t]{0.105\linewidth}
            \centering
            \includegraphics[width=0.95\linewidth]{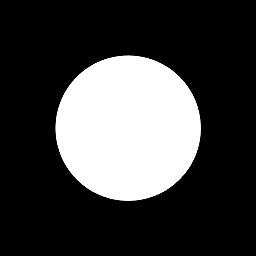}
        \end{minipage}
    }
    \hspace{-5mm}
    \subfigure{
        \begin{minipage}[t]{0.105\linewidth}
            \centering
            \includegraphics[width=0.95\linewidth]{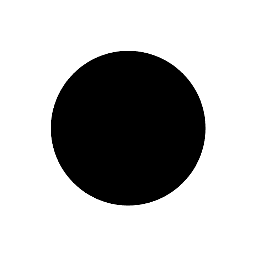}
        \end{minipage}
    }
    \hspace{-5mm}
    \subfigure{
        \begin{minipage}[t]{0.105\linewidth}
            \centering
            \includegraphics[width=0.95\linewidth]{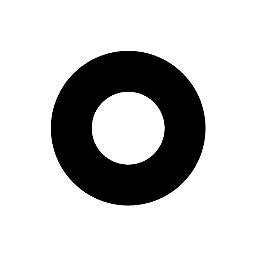}
        \end{minipage}
    }
    \hspace{-5mm}
    \subfigure{
        \begin{minipage}[t]{0.105\linewidth}
            \centering
            \includegraphics[width=0.95\linewidth]{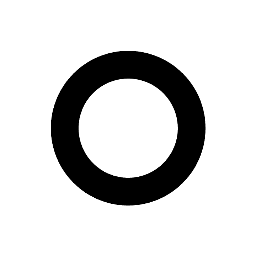}
        \end{minipage}
    }
    \hspace{-5mm}
    \subfigure{
        \begin{minipage}[t]{0.105\linewidth}
            \centering
            \includegraphics[width=0.95\linewidth]{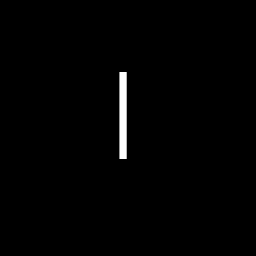}
        \end{minipage}
    }
    \hspace{-5mm}
    \subfigure{
        \begin{minipage}[t]{0.105\linewidth}
            \centering
            \includegraphics[width=0.95\linewidth]{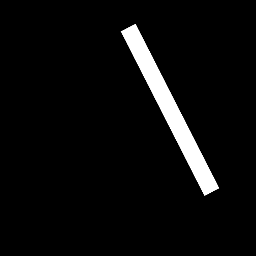}
        \end{minipage}
    }
    \hspace{-5mm}
    \subfigure{
        \begin{minipage}[t]{0.105\linewidth}
            \centering
            \includegraphics[width=0.95\linewidth]{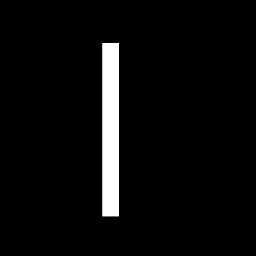}
        \end{minipage}
    }
        
    \caption{Visual reconstructions of simple geometric shapes.}
    \label{Fig-Generalization-line}
\end{figure}

In addition, the reconstruction results of the model on simple geometric shapes are shown in Fig. \ref{Fig-Generalization-line}. These results demonstrate the model's generalization capability across varied NLOS scenarios.

\begin{figure*}[!t]
    \centering
    \subfigure{
        \rotatebox{90}{\tiny{\ \ \quad \quad \quad \quad INPUT}} 
        \begin{minipage}[t]{0.11\linewidth} 
            \centering
            \includegraphics[width=1.08\linewidth]{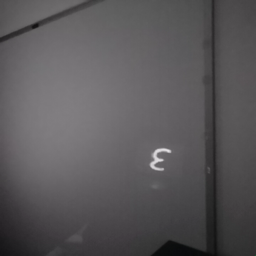}
        \end{minipage}
    }
    \hspace{-2mm}
    \subfigure{
        \begin{minipage}[t]{0.11\linewidth}
            \centering
            \includegraphics[width=1.08\linewidth]{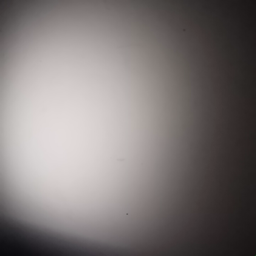}
        \end{minipage}
        \hspace{-0.5mm}
        \begin{minipage}[t]{0.11\linewidth}
            \centering
            \includegraphics[width=1.08\linewidth]{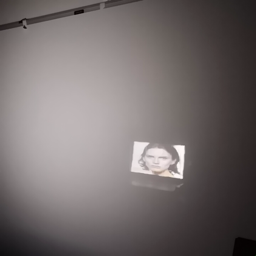}
        \end{minipage}
        \hspace{-0.5mm}
        \begin{minipage}[t]{0.11\linewidth}
            \centering
            \includegraphics[width=1.08\linewidth]{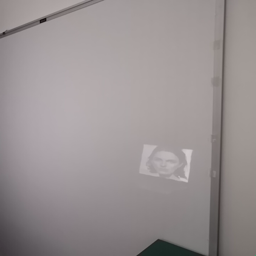}
        \end{minipage}

    }
    \hspace{-2mm}
    \subfigure{
        \begin{minipage}[t]{0.11\linewidth}
            \centering
            \includegraphics[width=1.08\linewidth]{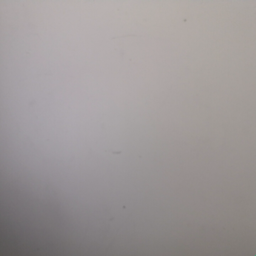}
        \end{minipage}
        \hspace{-0.5mm}
        \begin{minipage}[t]{0.11\linewidth}
            \centering
            \includegraphics[width=1.08\linewidth]{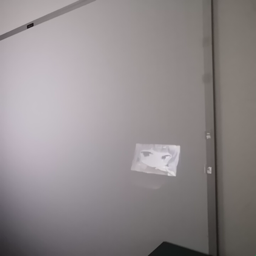}
        \end{minipage}
        \hspace{-0.5mm}
        \begin{minipage}[t]{0.11\linewidth}
            \centering
            \includegraphics[width=1.08\linewidth]
            {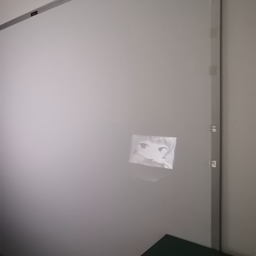}
        \end{minipage}
    }
    
    \vspace{-3mm}
    \setcounter{subfigure}{0}

    \subfigure{
        \rotatebox{90}{\tiny{\ \ \quad \quad \quad \ \ KBNet}}
        \begin{minipage}[t]{0.11\linewidth} 
            \centering
            \includegraphics[width=1.08\linewidth]{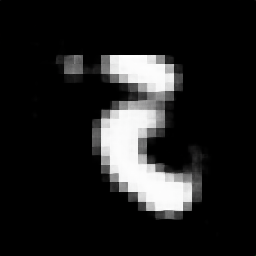}
        \end{minipage}
    }
    \hspace{-2mm}
    \subfigure{
        \begin{minipage}[t]{0.11\linewidth}
            \centering
            \includegraphics[width=1.08\linewidth]{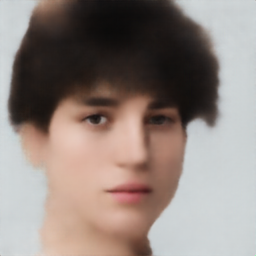}
        \end{minipage}
        \hspace{-0.5mm}
        \begin{minipage}[t]{0.11\linewidth}
            \centering
            \includegraphics[width=1.08\linewidth]{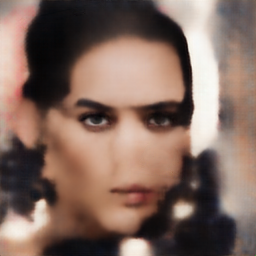}
        \end{minipage}
        \hspace{-0.5mm}
        \begin{minipage}[t]{0.11\linewidth}
            \centering
            \includegraphics[width=1.08\linewidth]{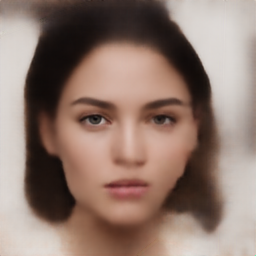}
        \end{minipage}
    }
    \hspace{-2mm}
    \subfigure{
        \begin{minipage}[t]{0.11\linewidth}
            \centering
            \includegraphics[width=1.08\linewidth]{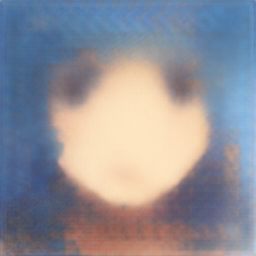}
        \end{minipage}
    \hspace{-0.5mm}
        \begin{minipage}[t]{0.11\linewidth}
            \centering
            \includegraphics[width=1.08\linewidth]{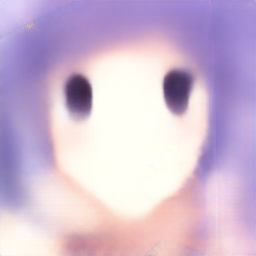}
        \end{minipage}
    \hspace{-0.5mm}
        \begin{minipage}[t]{0.11\linewidth}
            \centering
            \includegraphics[width=1.08\linewidth]{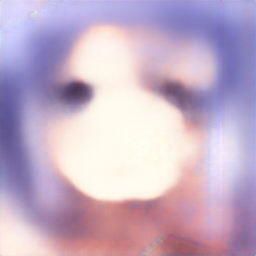}
        \end{minipage}
    }
    
    \vspace{-3mm}
    \setcounter{subfigure}{0}

    \subfigure{
        \rotatebox{90}{\tiny{\ \quad \quad \quad \ \ \ NAFNet}}
        \begin{minipage}[t]{0.11\linewidth} 
            \centering
            \includegraphics[width=1.08\linewidth]{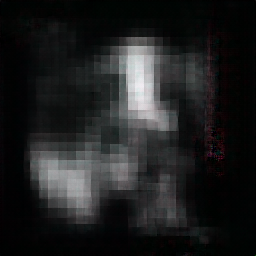}
        \end{minipage}
    }
    \hspace{-2mm}
    \subfigure{
        \begin{minipage}[t]{0.11\linewidth}
            \centering
            \includegraphics[width=1.08\linewidth]{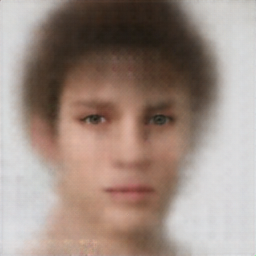}
        \end{minipage}
        \hspace{-0.5mm}
        \begin{minipage}[t]{0.11\linewidth}
            \centering
            \includegraphics[width=1.08\linewidth]{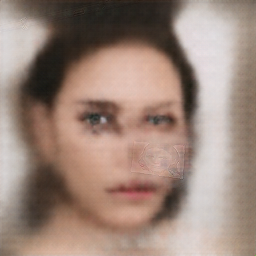}
        \end{minipage}
        \hspace{-0.5mm}
        \begin{minipage}[t]{0.11\linewidth}
            \centering
            \includegraphics[width=1.08\linewidth]{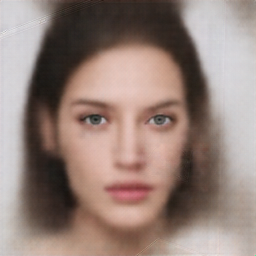}
        \end{minipage}
    }
    \hspace{-2mm}
    \subfigure{
        \begin{minipage}[t]{0.11\linewidth}
            \centering
            \includegraphics[width=1.08\linewidth]{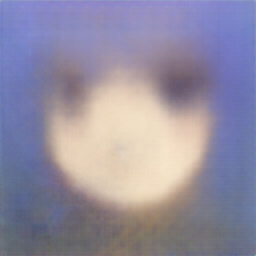}
        \end{minipage}
    \hspace{-0.5mm}
        \begin{minipage}[t]{0.11\linewidth}
            \centering
            \includegraphics[width=1.08\linewidth]{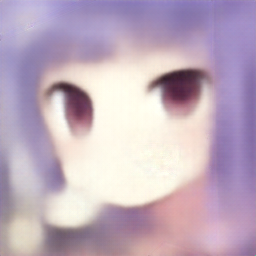}
        \end{minipage}
    \hspace{-0.5mm}
        \begin{minipage}[t]{0.11\linewidth}
            \centering
            \includegraphics[width=1.08\linewidth]{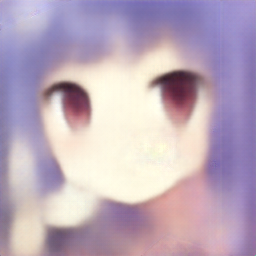}
        \end{minipage}
    }
    
    \vspace{-3mm}
    \setcounter{subfigure}{0}

    \subfigure{
        \rotatebox{90}{\tiny{\ \ \ \quad \quad \quad \ MSDINet}}
        \begin{minipage}[t]{0.11\linewidth} 
            \centering
            \includegraphics[width=1.08\linewidth]{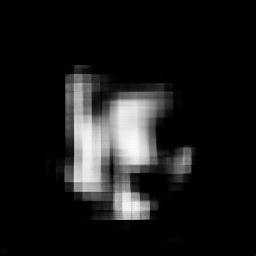}
        \end{minipage}
    }
    \hspace{-2mm}
    \subfigure{
        \begin{minipage}[t]{0.11\linewidth}
            \centering
            \includegraphics[width=1.08\linewidth]{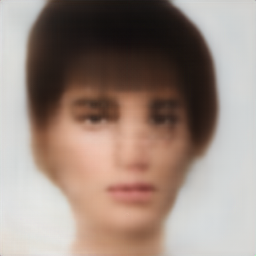}
        \end{minipage}
        \hspace{-0.5mm}
        \begin{minipage}[t]{0.11\linewidth}
            \centering
            \includegraphics[width=1.08\linewidth]{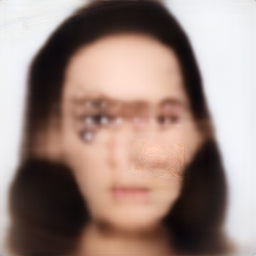}
        \end{minipage}
        \hspace{-0.5mm}
        \begin{minipage}[t]{0.11\linewidth}
            \centering
            \includegraphics[width=1.08\linewidth]{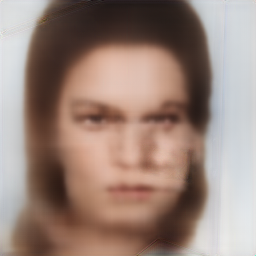}
        \end{minipage}
    }
    \hspace{-2mm}
    \subfigure{
        \begin{minipage}[t]{0.11\linewidth}
            \centering
            \includegraphics[width=1.08\linewidth]{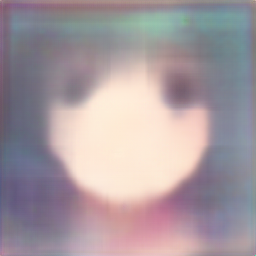}
        \end{minipage}
    \hspace{-0.5mm}
        \begin{minipage}[t]{0.11\linewidth}
            \centering
            \includegraphics[width=1.08\linewidth]{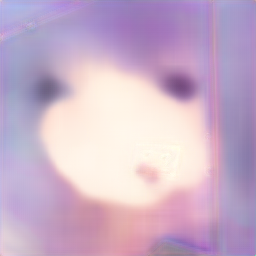}
        \end{minipage}
    \hspace{-0.5mm}
        \begin{minipage}[t]{0.11\linewidth}
            \centering
            \includegraphics[width=1.08\linewidth]{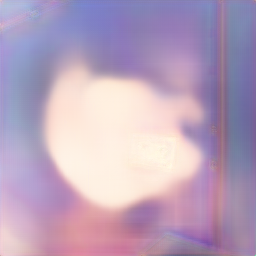}
        \end{minipage}
    }
    
    \vspace{-3mm}
    \setcounter{subfigure}{0}

    \subfigure{
        \rotatebox{90}{\tiny{\ \ \quad \quad \quad \ \ Uformer}}
        \begin{minipage}[t]{0.11\linewidth} 
            \centering
            \includegraphics[width=1.08\linewidth]{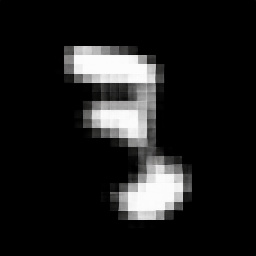}
        \end{minipage}
    }
    \hspace{-2mm}
    \subfigure{
        \begin{minipage}[t]{0.11\linewidth}
            \centering
            \includegraphics[width=1.08\linewidth]{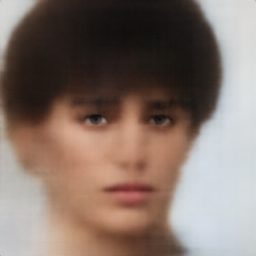}
        \end{minipage}
        \hspace{-0.5mm}
        \begin{minipage}[t]{0.11\linewidth}
            \centering
            \includegraphics[width=1.08\linewidth]{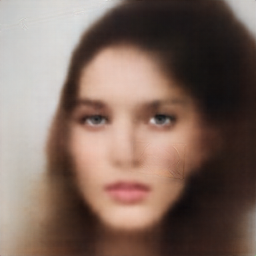}
        \end{minipage}
        \hspace{-0.5mm}
        \begin{minipage}[t]{0.11\linewidth}
            \centering
            \includegraphics[width=1.08\linewidth]{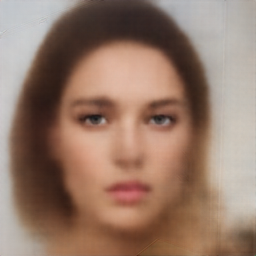}
        \end{minipage}
    }
    \hspace{-2mm}
    \subfigure{
        \begin{minipage}[t]{0.11\linewidth}
            \centering
            \includegraphics[width=1.08\linewidth]{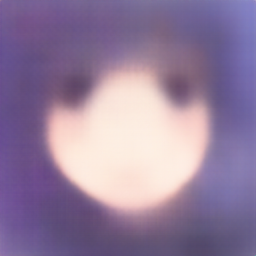}
        \end{minipage}
    \hspace{-0.5mm}
        \begin{minipage}[t]{0.11\linewidth}
            \centering
            \includegraphics[width=1.08\linewidth]{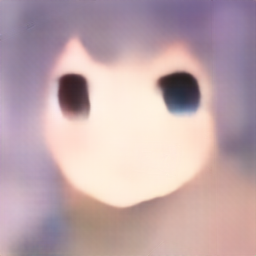}
        \end{minipage}
    \hspace{-0.5mm}
        \begin{minipage}[t]{0.11\linewidth}
            \centering
            \includegraphics[width=1.08\linewidth]{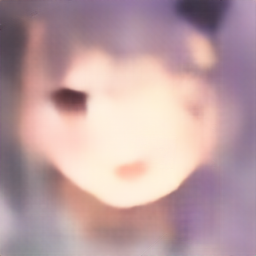}
        \end{minipage}
    }
    
    \vspace{-3mm}
    \setcounter{subfigure}{0}
    
    \subfigure{
        \rotatebox{90}{\tiny{\quad \quad \quad \quad SwinIR}}
        \begin{minipage}[t]{0.11\linewidth} 
            \centering
            \includegraphics[width=1.08\linewidth]{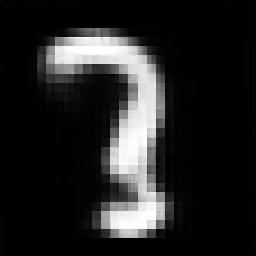}
        \end{minipage}
    }
    \hspace{-2mm}
    \subfigure{
        \begin{minipage}[t]{0.11\linewidth}
            \centering
            \includegraphics[width=1.08\linewidth]{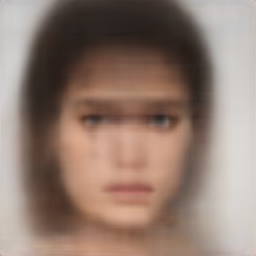}
        \end{minipage}
        \hspace{-0.5mm}
        \begin{minipage}[t]{0.11\linewidth}
            \centering
            \includegraphics[width=1.08\linewidth]{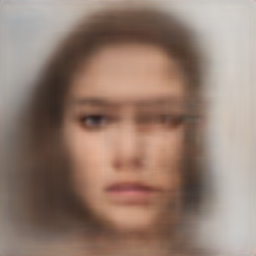}
        \end{minipage}
        \hspace{-0.5mm}
        \begin{minipage}[t]{0.11\linewidth}
            \centering
            \includegraphics[width=1.08\linewidth]{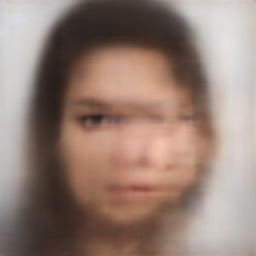}
        \end{minipage}
    }
    \hspace{-2mm}
    \subfigure{
        \begin{minipage}[t]{0.11\linewidth}
            \centering
            \includegraphics[width=1.08\linewidth]{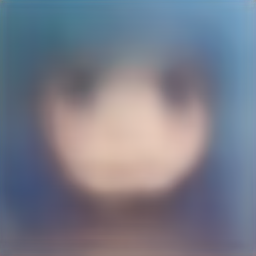}
        \end{minipage}
    \hspace{-0.5mm}
        \begin{minipage}[t]{0.11\linewidth}
            \centering
            \includegraphics[width=1.08\linewidth]{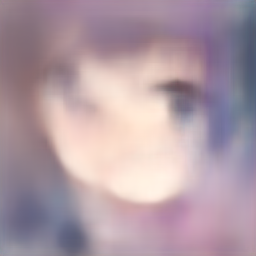}
        \end{minipage}
    \hspace{-0.5mm}
        \begin{minipage}[t]{0.11\linewidth}
            \centering
            \includegraphics[width=1.08\linewidth]{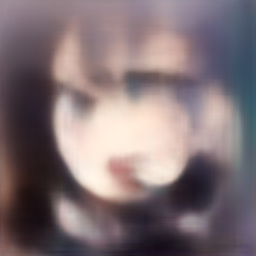}
        \end{minipage}
    }
    
    \vspace{-3mm}
    \setcounter{subfigure}{0}
    
    \subfigure{
        \rotatebox{90}{\tiny{\ \ \quad \quad \quad \ NLOS-OT}}
        \begin{minipage}[t]{0.11\linewidth} 
            \centering
            \includegraphics[width=1.08\linewidth]{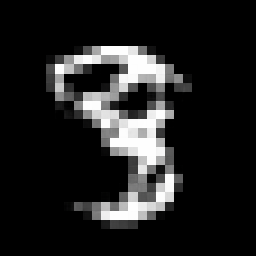}
        \end{minipage}
    }
    \hspace{-2mm}
    \subfigure{
        \begin{minipage}[t]{0.11\linewidth}
            \centering
            \includegraphics[width=1.08\linewidth]{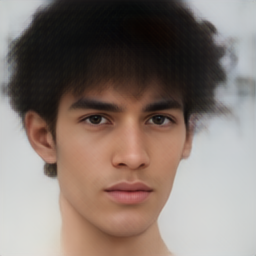}
        \end{minipage}
        \hspace{-0.5mm}
        \begin{minipage}[t]{0.11\linewidth}
            \centering
            \includegraphics[width=1.08\linewidth]{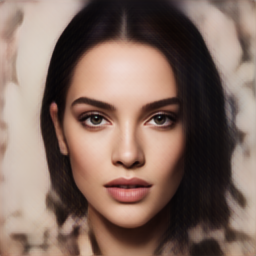}
        \end{minipage}
        \hspace{-0.5mm}
        \begin{minipage}[t]{0.11\linewidth}
            \centering
            \includegraphics[width=1.08\linewidth]{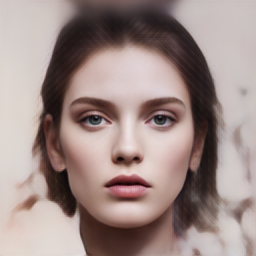}
        \end{minipage}
    }
    \hspace{-2mm}
    \subfigure{
        \begin{minipage}[t]{0.11\linewidth}
            \centering
            \includegraphics[width=1.08\linewidth]{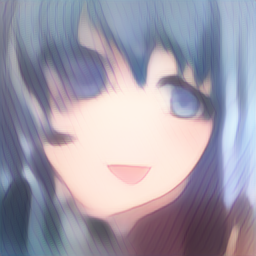}
        \end{minipage}
    \hspace{-0.5mm}
        \begin{minipage}[t]{0.11\linewidth}
            \centering
            \includegraphics[width=1.08\linewidth]{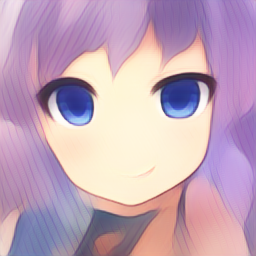}
        \end{minipage}
    \hspace{-0.5mm}
        \begin{minipage}[t]{0.11\linewidth}
            \centering
            \includegraphics[width=1.08\linewidth]{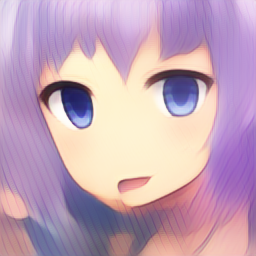}
        \end{minipage}
    }
    
    \vspace{-3mm}
    \setcounter{subfigure}{0}    
    
    \subfigure{
        \rotatebox{90}{\tiny{\ \quad \quad \quad \textbf{NLOS-LTM}}}
        \begin{minipage}[t]{0.11\linewidth} 
            \centering
            \includegraphics[width=1.08\linewidth]{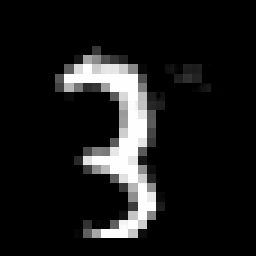}
        \end{minipage}
    }
    \hspace{-2mm}
    \subfigure{
        \begin{minipage}[t]{0.11\linewidth}
            \centering
            \includegraphics[width=1.08\linewidth]{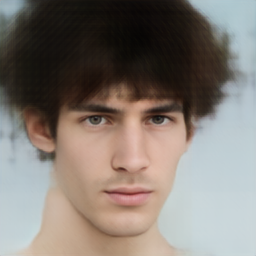}
        \end{minipage}
        \hspace{-0.5mm}
        \begin{minipage}[t]{0.11\linewidth}
            \centering
            \includegraphics[width=1.08\linewidth]{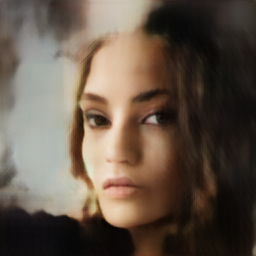}
        \end{minipage}
        \hspace{-0.5mm}
        \begin{minipage}[t]{0.11\linewidth}
            \centering
            \includegraphics[width=1.08\linewidth]{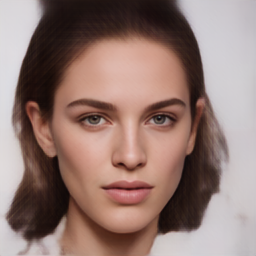}
        \end{minipage}
    }
    \hspace{-2mm}
    \subfigure{
        \begin{minipage}[t]{0.11\linewidth}
            \centering
            \includegraphics[width=1.08\linewidth]{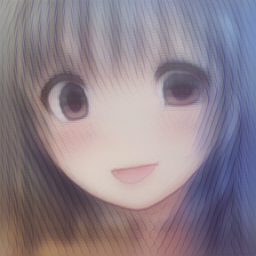}
        \end{minipage}
    \hspace{-0.5mm}
        \begin{minipage}[t]{0.11\linewidth}
            \centering
            \includegraphics[width=1.08\linewidth]{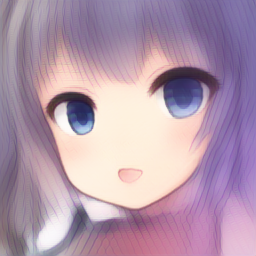}
        \end{minipage}
    \hspace{-0.5mm}
        \begin{minipage}[t]{0.11\linewidth}
            \centering
            \includegraphics[width=1.08\linewidth]{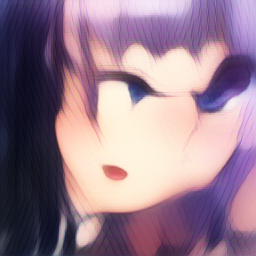}
        \end{minipage}
    }
    
    \vspace{-3mm}
    \setcounter{subfigure}{0}    
    
    \subfigure[MNIST]{
        \rotatebox{90}{\tiny{\quad \quad \quad \quad \quad GT}}
        \begin{minipage}[t]{0.11\linewidth} 
            \centering
            \includegraphics[width=1.08\linewidth]{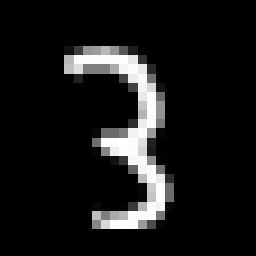}
            \centerline{\scriptsize{100;2;L;Wb}}
            \vspace{-2mm}
        \end{minipage}
    }
    \hspace{-2mm}
    \subfigure[Supermodel]{
        \begin{minipage}[t]{0.11\linewidth}
            \centering
            \includegraphics[width=1.08\linewidth]{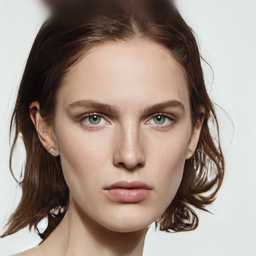}
            \centerline{\scriptsize{70;1;A;Wall}}
            \vspace{-2mm}
        \end{minipage}
        \hspace{-0.5mm}
        \begin{minipage}[t]{0.11\linewidth}
            \centering
            \includegraphics[width=1.08\linewidth]{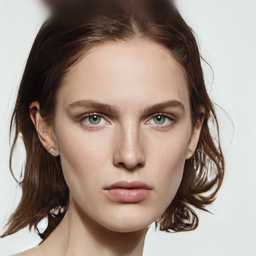}
            \centerline{\scriptsize{100;1;A;Wb}}
            \vspace{-2mm}
        \end{minipage}
        \hspace{-0.5mm}
        \begin{minipage}[t]{0.11\linewidth}
            \centering
            \includegraphics[width=1.08\linewidth]{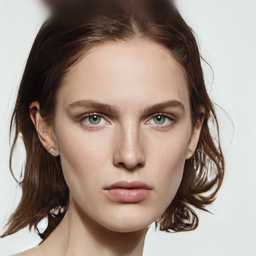}
            \centerline{\scriptsize{100;2;L;Wb}}
            \vspace{-2mm}
        \end{minipage}
    }
    \hspace{-2mm}
    \subfigure[Anime]{
        \begin{minipage}[t]{0.11\linewidth}
            \centering
            \includegraphics[width=1.08\linewidth]{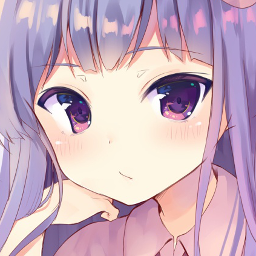}
            \centerline{\scriptsize{100;1;L;Wall}}
            \vspace{-2mm}
        \end{minipage}
    \hspace{-0.5mm}
        \begin{minipage}[t]{0.11\linewidth}
            \centering
            \includegraphics[width=1.08\linewidth]{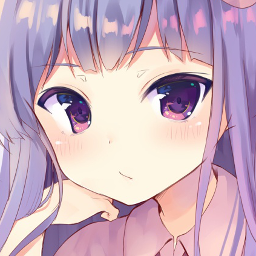}
            \centerline{\scriptsize{100;1;L;Wb}}
            \vspace{-2mm}
        \end{minipage}
    \hspace{-0.5mm}
        \begin{minipage}[t]{0.11\linewidth}
            \centering
            \includegraphics[width=1.08\linewidth]{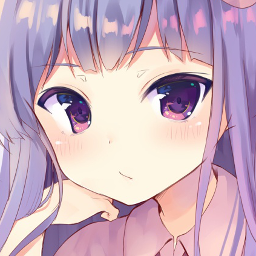}
            \centerline{\scriptsize{100;2;L;Wb}}
            \vspace{-2mm}
        \end{minipage}
    }  
    
    \caption{Visual comparison of the generalization ability across different light transport conditions. The model is trained under the light transport conditions listed in Table I and tested under other unseen light transport conditions.}
    \label{Fig-CrossConditions-Generalization}
\end{figure*}

\textit{2) Cross-Conditions Generalization Ability:}
The proposed method performs well under the light transport conditions included in the training set. It is also important to evaluate its ability to generalize to conditions outside the training set. In Fig. \ref{Fig-CrossConditions-Generalization}, we show some reconstruction results under this setting. For each dataset, the listed conditions are unseen during its model training. We can see that NLOS-LTM has better cross-condition generalization ability than other baseline methods. But in general, its reconstruction ability declines when encountering light transport conditions not represented in the training set as expected. This, from another point of view, demonstrates how heavily the NLOS projection and reconstruction processes are dependent on the light transport condition and the value of explicitly modeling the light transport condition as we did in NLOS-LTM. 
\begin{table*}[!t]
    \fontsize{7pt}{7pt}\selectfont
    \caption{Cross-datasets generalization ability comparison. The best and the second best scores of the multi-condition methods are \textbf{highlighted} and \underline{underlined} respectively.}
\begin{tabular}{ccccccccccccccccc}
\toprule
\textbf{STL-10}      & \multicolumn{2}{c}{\textbf{\begin{tabular}[c]{@{}c@{}}70;1;\\ A;Wall\end{tabular}}} & \multicolumn{2}{c}{\textbf{\begin{tabular}[c]{@{}c@{}}100;1;\\ A;Wall\end{tabular}}} & \multicolumn{2}{c}{\textbf{\begin{tabular}[c]{@{}c@{}}70;2;\\ A;Wall\end{tabular}}} & \multicolumn{2}{c}{\textbf{\begin{tabular}[c]{@{}c@{}}100;2;\\ A;Wall\end{tabular}}} & \multicolumn{2}{c}{\textbf{\begin{tabular}[c]{@{}c@{}}70;1;\\ A;Wb\end{tabular}}} & \multicolumn{2}{c}{\textbf{\begin{tabular}[c]{@{}c@{}}100;1;\\ A;Wb\end{tabular}}} & \multicolumn{2}{c}{\textbf{\begin{tabular}[c]{@{}c@{}}70;2;\\ A;Wb\end{tabular}}} & \multicolumn{2}{c}{\textbf{\begin{tabular}[c]{@{}c@{}}100;2;\\ A;Wb\end{tabular}}} \\
\textbf{Method}     & PSNR                                     & SSIM                                     & PSNR                                      & SSIM                                     & PSNR                                     & SSIM                                     & PSNR                                      & SSIM                                     & PSNR                                    & SSIM                                    & PSNR                                     & SSIM                                    & PSNR                                    & SSIM                                    & PSNR                                     & SSIM                                    \\ \midrule
KBNet               & \textbf{16.37}                           & \textbf{0.505}                           & 16.97                                     & \textbf{0.522}                           & 16.75                                    & \textbf{0.523}                           & {\ul 16.62}                               & \textbf{0.519}                           & {\ul 18.84}                             & {\ul 0.585}                             & {\ul 18.38}                              & {\ul 0.572}                             & {\ul 18.45}                             & {\ul 0.561}                             & {\ul 18.65}                              & {\ul 0.571}                             \\
NAFNet              & 16.30                                    & 0.465                                    & {\ul 17.14}                               & 0.485                                    & \textbf{16.96}                           & 0.483                                    & \textbf{16.90}                            & 0.489                                    & 18.57                                   & 0.543                                   & 18.16                                    & 0.530                                   & 18.17                                   & 0.522                                   & 18.38                                    & 0.528                                   \\
MSDINet             & 15.00                                    & 0.439                                    & 15.58                                     & 0.457                                    & 15.41                                    & 0.456                                    & 15.33                                     & 0.457                                    & 16.83                                   & 0.508                                   & 16.43                                    & 0.497                                   & 16.45                                   & 0.477                                   & 16.70                                    & 0.498                                   \\
Uformer             & 15.65                                    & 0.467                                    & 16.80                                     & 0.491                                    & 16.17                                    & 0.478                                    & 15.92                                     & 0.466                                    & 17.76                                   & 0.514                                   & 17.65                                    & 0.511                                   & 16.85                                   & 0.485                                   & 17.37                                    & 0.505                                   \\
SwinIR              & 15.99                                    & 0.484                                    & 16.73                                     & 0.504                                    & 16.38                                    & 0.496                                    & 16.21                                     & 0.489                                    & 17.04                                   & 0.512                                   & 16.65                                    & 0.504                                   & 16.61                                   & 0.494                                   & 16.79                                    & 0.553                                   \\
NLOS-OT             & {\ul 16.33}                              & 0.478                                    & 17.07                                     & 0.498                                    & 16.84                                    & 0.494                                    & 16.55                                     & 0.483                                    & 18.22                                   & 0.529                                   & 17.72                                    & 0.522                                   & 17.68                                   & 0.509                                   & 17.87                                    & 0.520                                   \\
\textbf{NLOS-LTM}   & \textbf{16.37}                           & {\ul 0.486}                              & \textbf{17.31}                            & {\ul 0.514}                              & {\ul 16.93}                              & {\ul 0.507}                              & 16.53                                     & {\ul 0.499}                              & \textbf{19.74}                          & \textbf{0.608}                          & \textbf{19.10}                           & \textbf{0.588}                          & \textbf{18.94}                          & \textbf{0.578}                          & \textbf{19.23}                           & \textbf{0.588}                          \\ \hline \hline
\textbf{MNIST}      & \multicolumn{2}{c}{\textbf{\begin{tabular}[c]{@{}c@{}}70;1;\\ A;Wall\end{tabular}}} & \multicolumn{2}{c}{\textbf{\begin{tabular}[c]{@{}c@{}}100;1;\\ A;Wall\end{tabular}}} & \multicolumn{2}{c}{\textbf{\begin{tabular}[c]{@{}c@{}}70;2;\\ A;Wall\end{tabular}}} & \multicolumn{2}{c}{\textbf{\begin{tabular}[c]{@{}c@{}}100;2;\\ A;Wall\end{tabular}}} & \multicolumn{2}{c}{\textbf{\begin{tabular}[c]{@{}c@{}}70;1;\\ A;Wb\end{tabular}}} & \multicolumn{2}{c}{\textbf{\begin{tabular}[c]{@{}c@{}}100;1;\\ A;Wb\end{tabular}}} & \multicolumn{2}{c}{\textbf{\begin{tabular}[c]{@{}c@{}}70;2;\\ A;Wb\end{tabular}}} & \multicolumn{2}{c}{\textbf{\begin{tabular}[c]{@{}c@{}}100;2;\\ A;Wb\end{tabular}}} \\
\textbf{Method}     & PSNR                                     & SSIM                                     & PSNR                                      & SSIM                                     & PSNR                                     & SSIM                                     & PSNR                                      & SSIM                                     & PSNR                                    & SSIM                                    & PSNR                                     & SSIM                                    & PSNR                                    & SSIM                                    & PSNR                                     & SSIM                                    \\ \midrule
KBNet               & 13.11                                    & \textbf{0.333}                           & 13.11                                     & {\ul 0.289}                              & 13.12                                    & \textbf{0.330}                           & 12.77                                     & \textbf{0.314}                           & {\ul 16.16}                             & \textbf{0.294}                          & 15.60                                    & \textbf{0.276}                          & 15.45                                   & \textbf{0.371}                          & {\ul 16.54}                              & \textbf{0.288}                          \\
NAFNet              & 13.37                                    & 0.095                                    & 13.97                                     & 0.082                                    & 13.56                                    & 0.076                                    & {\ul 13.55}                               & 0.072                                    & 14.91                                   & 0.091                                   & 15.03                                    & 0.102                                   & 15.04                                   & 0.100                                   & 15.50                                    & 0.088                                   \\
MSDINet             & 12.10                                    & {\ul 0.288}                              & 12.18                                     & \textbf{0.294}                           & 12.13                                    & {\ul 0.288}                              & 12.25                                     & {\ul 0.273}                              & 12.46                                   & {\ul 0.217}                             & 12.40                                    & {\ul 0.229}                             & 12.70                                   & {\ul 0.202}                             & 12.76                                    & 0.160                                   \\
Uformer             & 13.58                                    & 0.145                                    & {\ul 14.20}                               & 0.213                                    & {\ul 13.57}                              & 0.180                                    & 13.28                                     & 0.128                                    & 15.82                                   & 0.115                                   & {\ul 15.65}                              & 0.139                                   & {\ul 16.36}                             & 0.049                                   & 15.87                                    & 0.072                                   \\
SwinIR              & 12.49                                    & 0.107                                    & 12.68                                     & 0.102                                    & 12.51                                    & 0.109                                    & 12.51                                     & 0.084                                    & 13.21                                   & 0.117                                   & 12.99                                    & 0.102                                   & 13.06                                   & 0.091                                   & 13.04                                    & 0.093                                   \\
NLOS-OT             & {\ul 13.73}                              & 0.129                                    & 13.79                                     & 0.170                                    & 13.76                                    & 0.160                                    & 13.12                                     & 0.123                                    & 14.04                                   & 0.148                                   & 14.48                                    & 0.170                                   & 14.36                                   & 0.148                                   & 14.24                                    & 0.138                                   \\
\textbf{NLOS-LTM}   & \textbf{14.63}                           & 0.146                                    & \textbf{14.75}                            & 0.151                                    & \textbf{14.63}                           & 0.151                                    & \textbf{13.74}                            & 0.130                                    & \textbf{17.18}                          & 0.160                                   & \textbf{17.23}                           & 0.182                                   & \textbf{16.50}                          & 0.168                                   & \textbf{17.46}                           & {\ul 0.164}                             \\ \hline \hline
\textbf{Supermodel} & \multicolumn{2}{c}{\textbf{\begin{tabular}[c]{@{}c@{}}70;1;\\ A;Wall\end{tabular}}} & \multicolumn{2}{c}{\textbf{\begin{tabular}[c]{@{}c@{}}100;1;\\ A;Wall\end{tabular}}} & \multicolumn{2}{c}{\textbf{\begin{tabular}[c]{@{}c@{}}70;2;\\ A;Wall\end{tabular}}} & \multicolumn{2}{c}{\textbf{\begin{tabular}[c]{@{}c@{}}100;2;\\ A;Wall\end{tabular}}} & \multicolumn{2}{c}{\textbf{\begin{tabular}[c]{@{}c@{}}70;1;\\ A;Wb\end{tabular}}} & \multicolumn{2}{c}{\textbf{\begin{tabular}[c]{@{}c@{}}100;1;\\ A;Wb\end{tabular}}} & \multicolumn{2}{c}{\textbf{\begin{tabular}[c]{@{}c@{}}70;2;\\ A;Wb\end{tabular}}} & \multicolumn{2}{c}{\textbf{\begin{tabular}[c]{@{}c@{}}100;2;\\ A;Wb\end{tabular}}} \\
\textbf{Method}     & PSNR                                     & SSIM                                     & PSNR                                      & SSIM                                     & PSNR                                     & SSIM                                     & PSNR                                      & SSIM                                     & PSNR                                    & SSIM                                    & PSNR                                     & SSIM                                    & PSNR                                    & SSIM                                    & PSNR                                     & SSIM                                    \\ \midrule
KBNet               & 14.93                                    & 0.552                                    & 15.48                                     & 0.565                                    & 15.31                                    & {\ul 0.587}                              & 15.01                                     & 0.568                                    & {\ul 17.39}                             & {\ul 0.619}                             & 17.62                                    & {\ul 0.630}                             & {\ul 17.89}                             & {\ul 0.639}                             & {\ul 18.11}                              & {\ul 0.639}                             \\
NAFNet              & 15.20                                    & {\ul 0.562}                              & 16.41                                     & {\ul 0.576}                              & 15.99                                    & 0.576                                    & 15.70                                     & {\ul 0.578}                              & 17.37                                   & 0.591                                   & {\ul 17.73}                              & 0.593                                   & 17.82                                   & 0.610                                   & 17.88                                    & 0.622                                   \\
MSDINet             & 13.32                                    & 0.469                                    & 13.57                                     & 0.450                                    & 13.43                                    & 0.485                                    & 13.21                                     & 0.433                                    & 15.02                                   & 0.534                                   & 14.83                                    & 0.520                                   & 15.13                                   & 0.503                                   & 15.35                                    & 0.518                                   \\
Uformer             & 14.71                                    & 0.503                                    & 16.05                                     & 0.525                                    & 15.43                                    & 0.524                                    & 14.69                                     & 0.517                                    & 16.51                                   & 0.544                                   & 17.27                                    & 0.557                                   & 16.21                                   & 0.536                                   & 16.28                                    & 0.544                                   \\
SwinIR              & 14.32                                    & 0.502                                    & 15.03                                     & 0.509                                    & 14.60                                    & 0.506                                    & 14.34                                     & 0.503                                    & 14.85                                   & 0.512                                   & 14.99                                    & 0.509                                   & 14.94                                   & 0.512                                   & 15.02                                    & 0.512                                   \\
NLOS-OT             & {\ul 15.65}                              & 0.545                                    & {\ul 16.46}                               & 0.554                                    & {\ul 16.10}                              & 0.554                                    & {\ul 15.84}                               & 0.545                                    & 17.35                                   & 0.592                                   & 17.47                                    & 0.588                                   & 17.50                                   & 0.585                                   & 17.57                                    & 0.591                                   \\
\textbf{NLOS-LTM}   & \textbf{15.72}                           & \textbf{0.596}                           & \textbf{16.91}                            & \textbf{0.619}                           & \textbf{16.51}                           & \textbf{0.630}                           & \textbf{15.95}                            & \textbf{0.605}                           & \textbf{19.10}                          & \textbf{0.702}                          & \textbf{19.38}                                    & \textbf{0.710}                          & \textbf{19.33}                          & \textbf{0.699}                          & \textbf{19.48}                           & \textbf{0.690}                          \\ \hline \hline
\textbf{Real}       & \multicolumn{2}{c}{\textbf{\begin{tabular}[c]{@{}c@{}}70;1;\\ A;Wall\end{tabular}}} & \multicolumn{2}{c}{\textbf{\begin{tabular}[c]{@{}c@{}}100;1;\\ A;Wall\end{tabular}}} & \multicolumn{2}{c}{\textbf{\begin{tabular}[c]{@{}c@{}}70;2;\\ A;Wall\end{tabular}}} & \multicolumn{2}{c}{\textbf{\begin{tabular}[c]{@{}c@{}}100;2;\\ A;Wall\end{tabular}}} & \multicolumn{2}{c}{\textbf{\begin{tabular}[c]{@{}c@{}}70;1;\\ A;Wb\end{tabular}}} & \multicolumn{2}{c}{\textbf{\begin{tabular}[c]{@{}c@{}}100;1;\\ A;Wb\end{tabular}}} & \multicolumn{2}{c}{\textbf{\begin{tabular}[c]{@{}c@{}}70;2;\\ A;Wb\end{tabular}}} & \multicolumn{2}{c}{\textbf{\begin{tabular}[c]{@{}c@{}}100;2;\\ A;Wb\end{tabular}}} \\
\textbf{Method}     & PSNR                                     & SSIM                                     & PSNR                                      & SSIM                                     & PSNR                                     & SSIM                                     & PSNR                                      & SSIM                                     & PSNR                                    & SSIM                                    & PSNR                                     & SSIM                                    & PSNR                                    & SSIM                                    & PSNR                                     & SSIM                                    \\ \midrule
KBNet               & 12.51                                    & 0.439                                    & 12.87                                     & 0.438                                    & 12.80                                    & 0.461                                    & 12.74                                     & 0.445                                    & {\ul 13.17}                             & {\ul 0.483}                             & 13.97                                    & 0.490                                   & 14.32                                   & {\ul 0.496}                             & 14.10                                    & {\ul 0.504}                             \\
NAFNet              & 12.64                                    & 0.419                                    & 13.19                                     & 0.431                                    & 13.05                                    & 0.425                                    & {\ul 13.27}                               & 0.439                                    & 12.99                                   & 0.430                                   & 13.79                                    & 0.441                                   & 14.26                                   & 0.463                                   & {\ul 14.21}                              & 0.465                                   \\
MSDINet             & 11.47                                    & 0.407                                    & 11.72                                     & 0.398                                    & 11.73                                    & 0.397                                    & 11.56                                     & 0.374                                    & 11.66                                   & 0.428                                   & 12.40                                    & 0.436                                   & 12.67                                   & 0.430                                   & 12.43                                    & 0.431                                   \\
Uformer             & 12.35                                    & 0.338                                    & 13.03                                     & 0.338                                    & 12.57                                    & 0.338                                    & 12.38                                     & 0.356                                    & 12.92                                   & 0.368                                   & 13.92                                    & 0.359                                   & 13.60                                   & 0.369                                   & 13.46                                    & 0.376                                   \\
SwinIR              & 12.16                                    & 0.430                                    & 12.43                                     & 0.432                                    & 12.28                                    & 0.430                                    & 12.33                                     & 0.433                                    & 11.88                                   & 0.424                                   & 12.50                                    & 0.433                                   & 12.65                                   & 0.432                                   & 12.39                                    & 0.430                                   \\
NLOS-OT             & \textbf{13.26}                           & \textbf{0.483}                           & {\ul 13.58}                               & {\ul 0.476}                              & \textbf{13.51}                           & {\ul 0.475}                              & \textbf{13.40}                            & {\ul 0.473}                              & 12.57                                   & 0.474                                   & {\ul 14.07}                              & {\ul 0.508}                             & {\ul 14.38}                             & 0.507                                   & 13.99                                    & 0.498                                   \\
\textbf{NLOS-LTM}   & {\ul 13.17}                              & {\ul 0.474}                              & \textbf{13.70}                            & \textbf{0.489}                           & {\ul 13.28}                              & \textbf{0.486}                           & 13.15                                     & \textbf{0.490}                           & \textbf{14.39}                          & \textbf{0.545}                          & \textbf{14.96}                           & \textbf{0.542}                          & \textbf{15.33}                          & \textbf{0.554}                          & \textbf{14.96}                           & \textbf{0.549}                          \\ \bottomrule
\end{tabular}
\label{Table-Generalization-8}
\end{table*}
\subsection{Effectiveness of the Reprojection Network} 
\begin{figure}[!t]
    \centering
    \subfigure{
        \rotatebox{90}{\tiny{\quad \ INPUT}}
        \begin{minipage}[t]{0.11\linewidth} 
            \centering
            \includegraphics[width=1\linewidth]{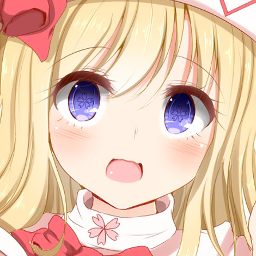}
        \end{minipage}
    }
    \hspace{-4mm}
    \subfigure{
        \begin{minipage}[t]{0.11\linewidth}
            \centering
            \includegraphics[width=1\linewidth]{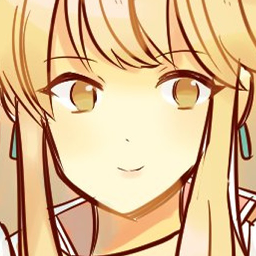}
        \end{minipage}
    }
    \hspace{-4mm}
    \subfigure{
        \begin{minipage}[t]{0.11\linewidth}
            \centering
            \includegraphics[width=1\linewidth]{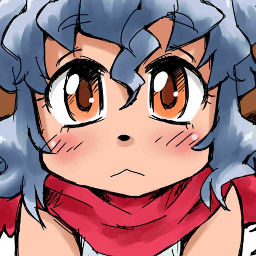}
        \end{minipage}
    }
    \hspace{-4mm}
    \subfigure{
        \begin{minipage}[t]{0.11\linewidth}
            \centering
            \includegraphics[width=1\linewidth]{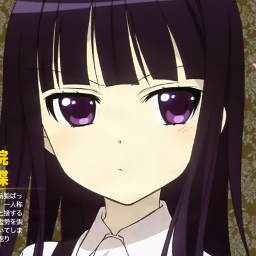}
        \end{minipage}
    }
    \hspace{-4mm}
    \subfigure{
        \begin{minipage}[t]{0.11\linewidth}
            \centering
            \includegraphics[width=1\linewidth]{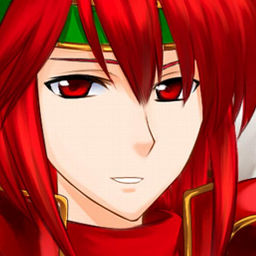}
        \end{minipage}
    }
    \hspace{-4mm}
    \subfigure{
        \begin{minipage}[t]{0.11\linewidth}
            \centering
            \includegraphics[width=1\linewidth]{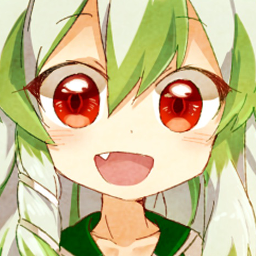}
        \end{minipage}
    }
    \hspace{-4mm}
    \subfigure{
        \begin{minipage}[t]{0.11\linewidth}
            \centering
            \includegraphics[width=1\linewidth]{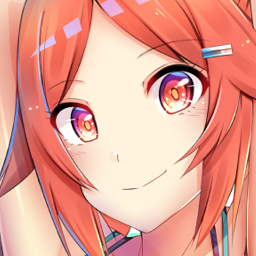}
        \end{minipage}
    }
    \hspace{-4mm}
    \subfigure{
        \begin{minipage}[t]{0.11\linewidth}
            \centering
            \includegraphics[width=1\linewidth]{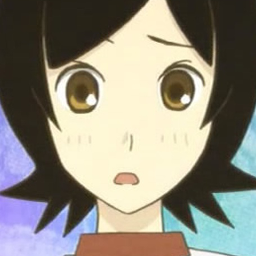}
        \end{minipage}
    }
    
    \vspace{-3.4mm}
    \setcounter{subfigure}{0}
    
    \subfigure{
        \rotatebox{90}{\tiny{\ \ Projection}}
        \begin{minipage}[t]{0.11\linewidth} 
            \centering
            \includegraphics[width=1\linewidth]{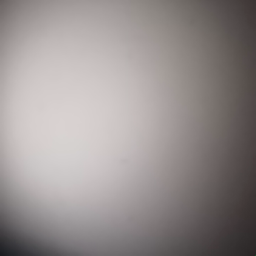}
        \end{minipage}
    }
    \hspace{-4mm}
    \subfigure{
        \begin{minipage}[t]{0.11\linewidth}
            \centering
            \includegraphics[width=1\linewidth]{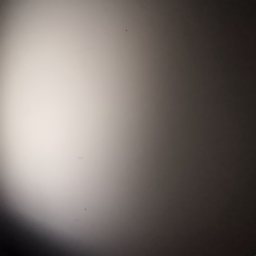}
        \end{minipage}
    }
    \hspace{-4mm}
    \subfigure{
        \begin{minipage}[t]{0.11\linewidth}
            \centering
            \includegraphics[width=1\linewidth]{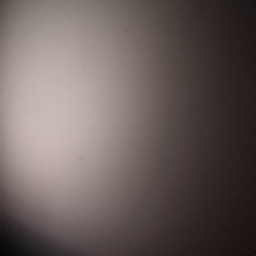}
        \end{minipage}
    }
    \hspace{-4mm}
    \subfigure{
        \begin{minipage}[t]{0.11\linewidth}
            \centering
            \includegraphics[width=1\linewidth]{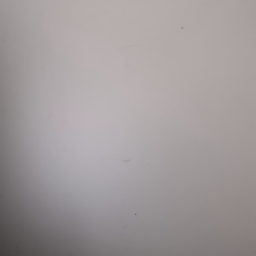}
        \end{minipage}
    }
    \hspace{-4mm}
    \subfigure{
        \begin{minipage}[t]{0.11\linewidth}
            \centering
            \includegraphics[width=1\linewidth]{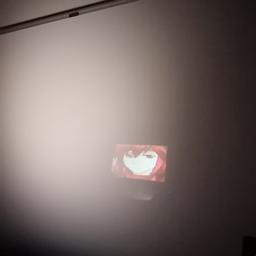}
        \end{minipage}
    }
    \hspace{-4mm}
    \subfigure{
        \begin{minipage}[t]{0.11\linewidth}
            \centering
            \includegraphics[width=1\linewidth]{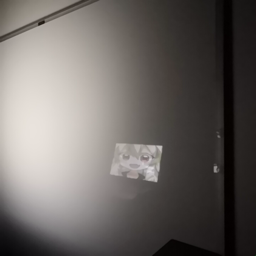}
        \end{minipage}
    }
    \hspace{-4mm}
    \subfigure{
        \begin{minipage}[t]{0.11\linewidth}
            \centering
            \includegraphics[width=1\linewidth]{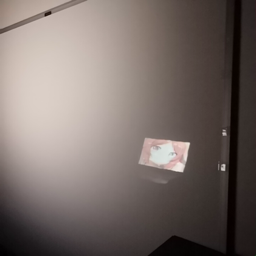}
        \end{minipage}
    }
    \hspace{-4mm}
    \subfigure{
        \begin{minipage}[t]{0.11\linewidth}
            \centering
            \includegraphics[width=1\linewidth]{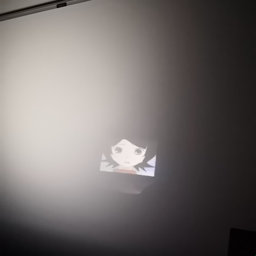}
        \end{minipage}
    }
    
    \vspace{-3.4mm}
    \setcounter{subfigure}{0}
    
    \subfigure{
        \rotatebox{90}{\tiny{Reprojection}}
        \begin{minipage}[t]{0.11\linewidth} 
            \centering
            \includegraphics[width=1\linewidth]{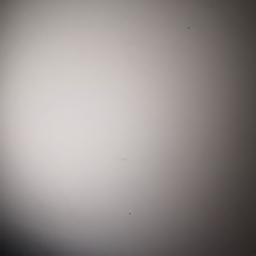}
            \centerline{\tiny{100;1;L;Wall}}
        \end{minipage}
    }
    \hspace{-4mm}
    \subfigure{
        \begin{minipage}[t]{0.11\linewidth}
            \centering
            \includegraphics[width=1\linewidth]{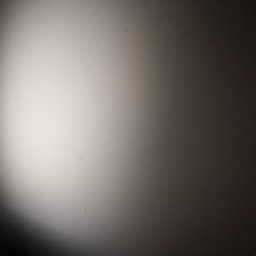}
            \centerline{\tiny{70;2;A;Wall}}
        \end{minipage}
    }
    \hspace{-4mm}
    \subfigure{
        \begin{minipage}[t]{0.11\linewidth}
            \centering
            \includegraphics[width=1\linewidth]{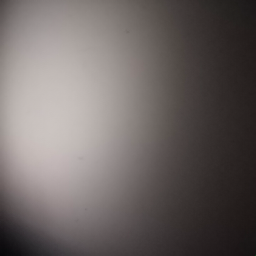}
            \centerline{\tiny{100;2;A;Wall}}
        \end{minipage}
    }
    \hspace{-4mm}
    \subfigure{
        \begin{minipage}[t]{0.11\linewidth}
            \centering
            \includegraphics[width=1\linewidth]{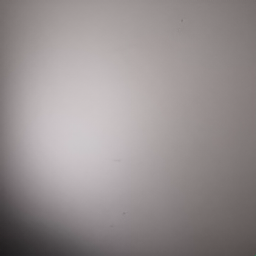}
            \centerline{\tiny{70;2;L;Wall}}
        \end{minipage}
    }
    \hspace{-4mm}
    \subfigure{
        \begin{minipage}[t]{0.11\linewidth}
            \centering
            \includegraphics[width=1\linewidth]{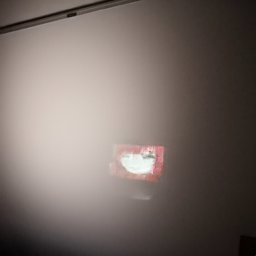}
            \centerline{\tiny{70;1;A;Wb}}
        \end{minipage}
    }
    \hspace{-4mm}
    \subfigure{
        \begin{minipage}[t]{0.11\linewidth}
            \centering
            \includegraphics[width=1\linewidth]{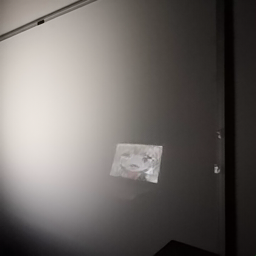}
            \centerline{\tiny{70;2;A;Wb}}
        \end{minipage}
    }
    \hspace{-4mm}
    \subfigure{
        \begin{minipage}[t]{0.11\linewidth}
            \centering
            \includegraphics[width=1\linewidth]{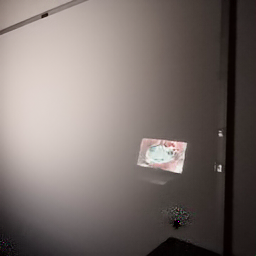}
            \centerline{\tiny{100;1;A;Wb}}
        \end{minipage}
    }
    \hspace{-4mm}
    \subfigure{
        \begin{minipage}[t]{0.11\linewidth}
            \centering
            \includegraphics[width=1\linewidth]{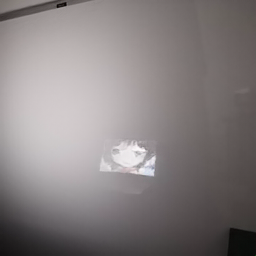}
            \centerline{\tiny{70;1;L;Wb}}
        \end{minipage}
    }
    \caption{Examples of projection images generated by the reprojection network $G_p$ under different light transport conditions.}
    \label{Fig-Reprojection}
\end{figure}

As a byproduct of the joint training framework, we obtain a reprojection network $G_p$ that can map a hidden image to the projection image with the guide of a light transport representation. Here, we qualitatively evaluate the performance of this reprojection network. Fig. \ref{Fig-Reprojection} shows some examples of projection images generated by $G_p$ under different light transport conditions. 
This demonstrates the effectiveness of $G_p$ and the learned light transport representations. Although $G_p$ is not used during testing, it enables the synthesis of projection images, which is valuable for studying the NLOS imaging problem.
\begin{table}[t]
    \footnotesize
    \caption{Ablation study for NLOS-LTM on Four-Anime dataset.}
\begin{tabular}{ccccc}
\toprule
                           & \textbf{\begin{tabular}[c]{@{}c@{}}100;1;\\ A;Wall\end{tabular}} & \textbf{\begin{tabular}[c]{@{}c@{}}70;2;\\ A;Wall\end{tabular}} & \textbf{\begin{tabular}[c]{@{}c@{}}70;1;\\ A;Wb\end{tabular}} & \textbf{\begin{tabular}[c]{@{}c@{}}70;2;\\ A;Wb\end{tabular}} \\
\textbf{Model}             & PSNR                                                            & PSNR                                                           & PSNR                                                         & PSNR                                                         \\ \midrule
w/o OT                     & 13.42                                                           & 13.87                                                          & 16.28                                                        & 16.24                                                        \\
w/o joint learning         & 11.67                                                           & 13.30                                                          & 14.05                                                        & 14.14                                                        \\
modulation w/o multi-scale & 13.17                                                           & 13.77                                                          & 16.74                                                        & 16.78                                                        \\
w/ concat modulation       & 13.40                                                           & 13.85                                                          & 17.12                                                        & 16.96                                                        \\
w/o vector quantization    & 13.37                                                           & 13.85                                                          & 17.01                                                        & 17.01                                                        \\ \midrule
\textbf{NLOS-LTM}          & \textbf{13.43}                                                  & \textbf{13.88}                                                 & \textbf{17.21}                                               & \textbf{17.24}                                               \\ \bottomrule
\end{tabular}
\label{Table-ablation-NLOS-LTM}
\end{table}
\subsection{Ablation Study}
To demonstrate the effectiveness of each component in NLOS-LTM, we conduct ablation studies on the Four-Anime dataset, and the results are shown in Table \ref{Table-ablation-NLOS-LTM}. We first omit the pretraining of the encoder $E_h$ and the decoder $D_h$ and remove the OT loss in Eq. \eqref{L_r}, so that the encoder and decoder of the reconstruction network $G_r$ are simultaneously learned (denoted as ``w/o OT''). Then, we remove the reprojection network $G_p$ but reserve $E_c$ and $ Q_c$ to extract the light transport representation from the projection images, which is used to aid image reconstruction (denoted as ``w/o joint learning''). Next, we remove the multi-scale modulation and only use one LTM block to modulate the feature maps (denoted as ``modulation w/o multi-scale''). Subsequently, we remove the LTM blocks and directly concatenate light transport representation with feature maps from the encoder (denoted as ``w/ concat modulation''). To ensure fair comparisons of computational cost, we add a residual block after the feature concatenation. Finally, we consider the case where vector quantization is not used and the output of $E_c$ is directly taken as the light transport representation for the following feature modulation (denoted as ``w/o vector quantization''). Our ablation studies indicate that each component of our method contributes to improving the performance. These results further demonstrate the effectiveness of the light transport representation learning and multi-scale light transport modulation for passive NLOS imaging. 
\section{Discussion}
\subsection{Necessity of a Complex Nonlinear Network}
Given that the forward model is linear, it naturally raises the question of why use a complex nonlinear network rather than a simple linear model. This is because the inverse problem of reconstructing $\mathbf{x}$ from $\mathbf{y}$ is highly ill-posed due to the high condition number of $\mathbf{A}$. This ill-posedness leads to a solution space that is highly sensitive to noise and small perturbations in $\mathbf{y}$. A complex nonlinear network is required for the following reasons:

\textit{1) Addressing Ill-Posedness:} The inverse problem of reconstructing hidden images from projections is highly sensitive to noise. Nonlinear networks can handle this instability better than linear models.

\textit{2) Learning Priors and Regularization:} Deep learning models can learn complex patterns and regularize the reconstruction, improving image quality.

\textit{3) Capturing Nonlinearities in Real-World Data:} Real-world scenarios often introduce nonlinear effects. Nonlinear networks can capture these subtleties, leading to more accurate reconstructions.

\textit{4) Adaptability and Generalization:} Deep networks can adapt to different conditions by learning shared representations, making them versatile for practical use.

\subsection{Relation to the Physical Light Transport Model}
Our approach explicitly integrates the principles of the physical light transport model into the network design, enhancing its interpretability and performance. The following aspects illustrate this integration:

\textit{1) Joint Learning Strategy:} The network learns to reconstruct hidden images and simulate the projection process, ensuring consistency with physical light transport principles.

\textit{2) Learning Light Transport Representations:} Using vector quantization and contrastive learning, the network separates the light transport information from the hidden image information, aligning with physical constraints.

\textit{3) Role of Light Transport Modulation Blocks:} These blocks dynamically adjust the reconstruction process based on inferred light transport conditions, enhancing the network's ability to handle various scenarios effectively.
\subsection{Limitation and Future Work}
Our work advances the field from training specific model for each condition to training a unified model for multiple conditions. As the first trial, the number of conditions we currently consider is limited. But it is possible to extend it to handle more light transport conditions. In the following, we propose several potential strategies for future work.

\textit{1) Expanding the Training Dataset:} The current dataset includes only a limited number of light transport conditions. Therefore, it is crucial to expand the dataset to encompass a broader range of conditions. With a more extensive and diverse dataset, the model can be designed and trained to handle a wider variety of light transport scenarios. Consequently, such a model will also exhibit better generalization capabilities. This step is necessary to ensure the model's robustness and versatility across different conditions.

\textit{2) Handling More Light Transport Conditions:} When the data set contains a large number of optical transport conditions, learning a code for each condition becomes inefficient or even infeasible. But we can still set to learn a dictionary that contains a set of codes. Then a light transport condition is represented as a weighted combination of the codes in the dictionary. Projection images captured under the same condition are required to have the same or similar combination weights. The model learned in this way can handle a more realistic setting with a huge more number of light transport conditions. Note that learning a set of discrete light transport representations which is studied in this work is also the basis for the new setting. We will explore this extension in our future work.

\textit{3) Incorporating Physical Models:} Integrating physical models of light transport into the training process can provide additional constraints and priors that help the model better understand the underlying physics. This method allows the network to learn and apply these physical laws, so as to achieve more accurate and reliable reconstruction under various optical transport conditions.

Additionally, while using a monitor to present the hidden scene is a setting widely adopted by existing works on passive NLOS imaging, it has certain limitations. In future work, it is valuable to investigate a broader range of scene settings, employing the following potential strategies.

\textit{4) Integrating Advanced Sensing Technology:} Regular cameras provide high-resolution 2D image data, while time-of-flight (ToF) cameras add depth information through precise distance measurements, and event cameras capture high-speed changes, allowing for real-time tracking of dynamic elements. By combining these complementary sources, the model gains access to richer auxiliary data that could improve passive NLOS reconstruction in more complex and diverse scenes.

\textit{5) Incorporating 3D Reconstruction Technology:} Another direction is to introduce 3D reconstruction technology to handle complex 3D geometries and non-Lambertian bidirectional reflectance distribution functions (BRDFs). One promising approach to achieve 3D reconstruction with NLOS-LTM is to use multiple relay surfaces and viewpoints to capture more comprehensive information about the hidden scene, thereby improving reconstruction ability.

\section{Conclusion}
In this paper, we propose NLOS-LTM, a deep learning-based method for passive NLOS imaging. With a single network, it can effectively restore hidden images from projection images that are captured under multiple light transport conditions. A strategy that jointly learns the reconstruction and the reprojection networks is employed during training. We use a light transport encoder followed by vector quantization to obtain a latent light transport representation from the projection image. The latent representation is then used to modulate both the reconstruction and the reprojection networks through a set of LTM blocks in a multi-scale way. The experimental results demonstrate that our method outperforms other methods in both quantitative and qualitative comparisons.
\ifCLASSOPTIONcaptionsoff
\newpage
\fi

\bibliography{document.bib} 

\bibliographystyle{IEEEtran} 
\end{document}